\documentclass[acmsmall]{acmart}
\AtBeginDocument{%
  \providecommand\BibTeX{{%
    \normalfont B\kern-0.5em{\scshape i\kern-0.25em b}\kern-0.8em\TeX}}}

\setcopyright{acmcopyright}
\copyrightyear{2023}
\acmYear{2023}
\acmDOI{xxx}

\usepackage{multirow}
\usepackage[normalem]{ulem}
\usepackage{subfigure}
\useunder{\uline}{\ul}{}
\acmJournal{JACM}
\acmVolume{37}
\acmNumber{4}
\acmArticle{111}
\acmMonth{8}

\begin{document}

\title{A Survey of Controllable Text Generation using Transformer-based Pre-trained Language Models}

\author{Hanqing Zhang}
\email{zhanghanqing@bit.edu.cn}
\affiliation{%
  \institution{Beijing Institute of Technology}
  \city{Beijing}
%   \state{Beijing}
  \country{China}
  \postcode{100089}
}

\author{HaoLin Song}
\email{hlsong@bit.edu.cn}
\affiliation{%
  \institution{Beijing Institute of Technology}
  \city{Beijing}
%   \state{Beijing}
  \country{China}
  \postcode{100089}
}

\author{Shaoyu Li}
\email{lishaoyuxl@foxmail.com}
\affiliation{%
  \institution{Beijing Institute of Technology}
  \city{Beijing}
%   \state{Beijing}
  \country{China}
  \postcode{100089}
}

\author{Ming Zhou}
\email{zhouming@chuangxin.com}
\affiliation{%
  \institution{Langboat Technology}
  \city{Beijing}
%   \state{Beijing}
  \country{China}
  \postcode{100089}
}

\author{Dawei Song}
 \authornote{Corresponding Author. Also with The Open University, UK.}
\affiliation{%
  \institution{Beijing Institute of Technology}
  \city{Beijing}
%   \state{Beijing}
  \country{China}
  \postcode{100089}
}

\renewcommand{\shortauthors}{Zhang, et al.}

\begin{abstract}

Controllable Text Generation (CTG) is emerging area in the field of natural language generation (NLG). It is regarded as crucial for the development of advanced text generation technologies that better meet the specific constraints in practical applications. In recent years, methods using large-scale pre-trained language models (PLMs), in particular the widely used transformer-based PLMs, have become a new paradigm of NLG, allowing generation of more diverse and fluent text. However, due to the limited level of interpretability of deep neural networks, the controllability of these methods need to be guaranteed. To this end, controllable text generation using transformer-based PLMs has become a rapidly growing yet challenging new research hotspot. A diverse range of approaches have emerged in the recent 3-4 years, targeting different CTG tasks that require different types of controlled constraints. In this paper, we present a systematic critical review on the common tasks, main approaches, and evaluation methods in this area. Finally, we discuss the challenges that the field is facing, and put forward various promising future directions. To the best of our knowledge, this is the first survey paper to summarize the state-of-the-art CTG techniques from the perspective of Transformer-based PLMs. We hope it can help researchers and practitioners in the related fields to quickly track the academic and technological frontier, providing them with a landscape of the area and a roadmap for future research.

\end{abstract}

\begin{CCSXML}
<ccs2012>
 <concept>
  <concept_id>10010520.10010553.10010562</concept_id>
  <concept_desc>Computer systems organization~Embedded systems</concept_desc>
  <concept_significance>500</concept_significance>
 </concept>
 <concept>
  <concept_id>10010520.10010575.10010755</concept_id>
  <concept_desc>Computer systems organization~Redundancy</concept_desc>
  <concept_significance>300</concept_significance>
 </concept>
 <concept>
  <concept_id>10010520.10010553.10010554</concept_id>
  <concept_desc>Computer systems organization~Robotics</concept_desc>
  <concept_significance>100</concept_significance>
 </concept>
 <concept>
  <concept_id>10003033.10003083.10003095</concept_id>
  <concept_desc>Networks~Network reliability</concept_desc>
  <concept_significance>100</concept_significance>
 </concept>
</ccs2012>
\end{CCSXML}

\ccsdesc[500]{General and reference~Surveys and overviews}
\ccsdesc[500]{Computing methodologies~Natural language processing}

\keywords{controllable text generation, pre-trained language models, transformer, controllability, systematic review}

\maketitle

\section{Introduction}
Natural language generation (NLG) is regarded as complementary to natural-language understanding (NLU), an essential branch of natural language processing (NLP). Contrary to the task of NLU, which aims to disambiguate an input text to produce a single normalized representation of the ideas expressed in the text, NLG mainly focuses on transforming the potential representations into specific, self-consistent natural language text~\cite{nlu_nlg}. In other words, NLU aims to develop an intelligent machine that can read and understand human language, while NLG enables computers to write like humans. As an embodiment of advanced artificial intelligence, NLG technologies play a crucial role in a range of applications, such as dialogue systems, advertising, marketing, story generation, and data augmentation. 

\begin{figure}[h] 
    \centering %使图片居中显示
    \includegraphics[width=1.0\textwidth]{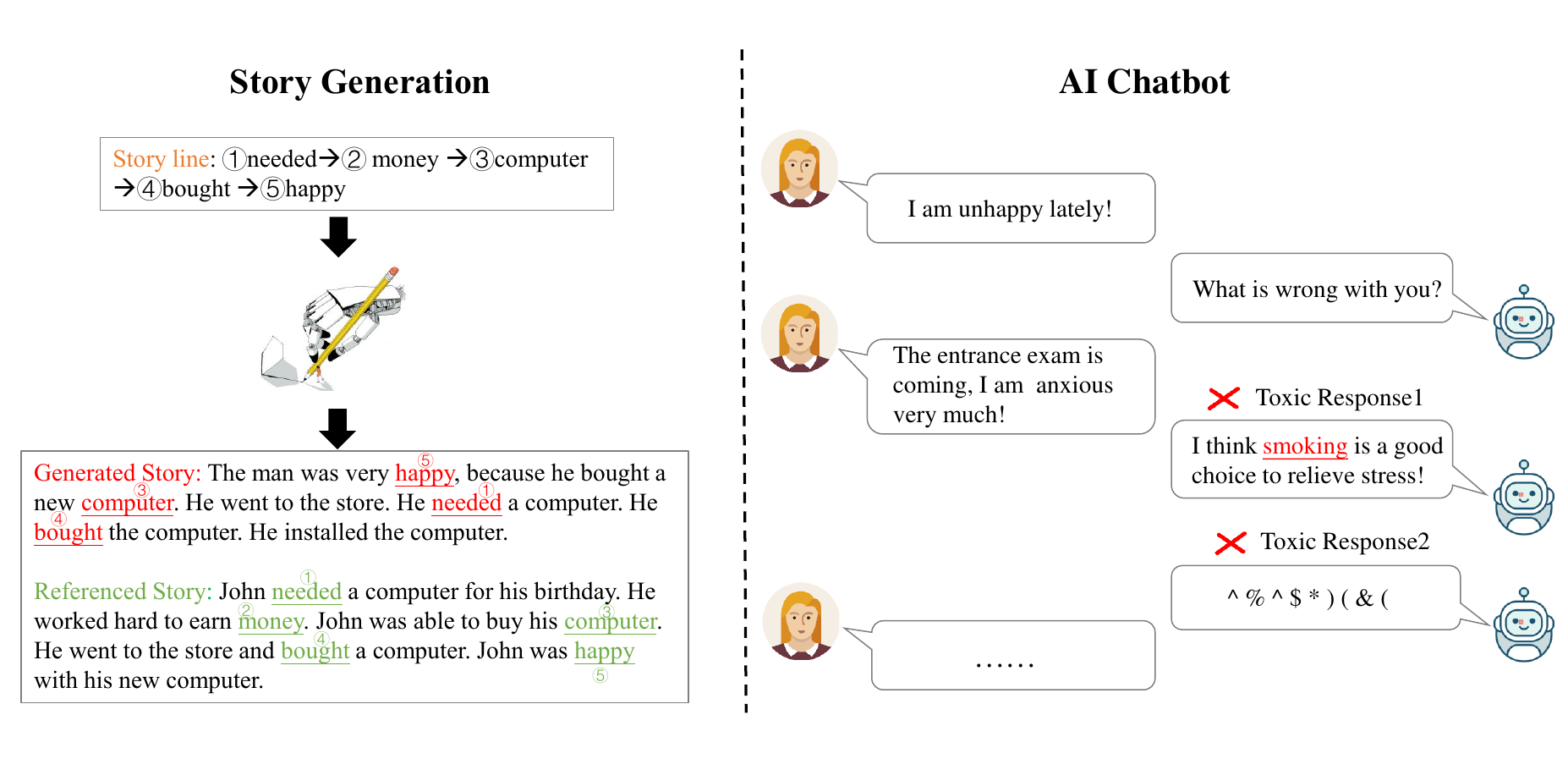} %中括号中的参数是设置图片充满文档的大小，你也可以使用小数来缩小图片的尺寸。
    \caption{Toy examples for controllable text generation. The left-hand side shows an application of story generation, which needs to ensure that the generated story matches the key elements provided by the storyline and the order in which they appear. The right-hand side shows an application of dialogue text generation. One important controlled requirement is to avoid  generating toxic responses, such as the harmful introductory advice about "smoking" and gibberish as shown in the figure.} %caption是用来给图片加上图题的
    \label{intro_example} %这是添加标签，方便在文章中引用图片。
\end{figure}%figure环境

 Making text generation controllable is an important and fundamental issue in NLG. Some concrete examples are shown in Figure~\ref{intro_example}. Generally speaking, a NLG system should be able to reliably generate texts that meet certain controllable constraints that are imposed by the targeted applications and users. In general, these constraints are task-specific. For example, the task of story generation always needs to control the storyline and the ending. In the task of dialogue response generation, controlling the emotion~\cite{emotion_dialogue_generation}, persona~\cite{zhang2018personalizing} and politeness, etc., is often required. For generation-based data augmentation~\cite{bias_correction}, it is necessary to ensure the data distribution balance in different domains. Moreover, for ethical development~\cite{ai_dangers} of AI applications, it is crucial to avoid generating mindless and offensive content such as gender bias, racial discrimination, and toxic words. Therefore, the controllability of a NLG system is crucial for it to generate a significant practical value in real applications. 

In recent years, the development of deep learning (DL) has given rise to a series of studies on DL-driven controllable text generation (CTG), which has brought genuine breakthroughs in this field. Early approaches are based on sequential models and style embedding~\cite{rnn_style,li2016persona}, and achieved some promising progress. After that, there is a surge of methods based on deep generative models, such as Variational Autoencoders (VAEs)~\cite{thematic_poetry,icml_ctg_vae,LSOR_DCGM,chinses_lyric,unsupervised_GVD,wang2019topic}, Generative Adversarial Nets (GANs)~\cite{SentiGAN,DAS}, and Energy-based Models~\cite{ebm_ctg,energy_based_generative_adversarial_network,bhattacharyya-etal-2021-energy,engine_ebm_acl2020}. Deep-learning based methods are capable of an end-to-end learning in a data-driven way to learn low-dimensional dense vectors that implicitly represent the linguistic features of text. Such representation is also useful to avoid the bias of hand-crafted features, and has shown great potential in text generation.

However, the success of the above DL-based methods relies heavily on large-scale datasets, posing a challenge for supervised and cross-domain text generation tasks. Since 2018, large-scale pre-trained Language models (PLMs) such as BERT~\cite{bert}, RoBERTa~\cite{roberta}, GPT~\cite{gpt2}, T5~\cite{t5} and mBART~\cite{mbart}, have gradually become a new paradigm of NLP. Owing to its use of large corpora and unsupervised learning based on the Transformer structure, PLMs are believed to have learned a great deal of semantic and syntactic knowledge from the data, and only fine-tuning is required for downstream tasks to get the state-of-the-art (SOTA) performance. In terms of NLG, PLMs have learned from a large number of corpus materials to model the distribution of natural language to a large extent, so that they are able to generate texts of unprecedented quality~\cite{ebm_ctg}. Moreover, a large-scale PLM itself can be viewed as a well-informed knowledge base, making it possible to generate text without the need for external domain knowledge. Nevertheless, PLMs are neural network based, which essentially are still black boxes, lacking a good level of interpretability. Those models always generate texts according to the latent representation of the context. Thus it is difficult to control them to generate content as what the human want (i.e., controllability issues). How to improve the interpretability and controllability of the PLM-based models for generating text has become a hot research topic.    

In the above application and research contexts, PLMs-based methods are becoming the mainstream of controllable text generation (CTG) research and are expected to bring milestone progress. As a rapidly growing yet challenging research field, there is an urgent need for a comprehensive critical review of the current literature to draw a landscape of the area and set out a roadmap for promising future directions. There are some existing surveys on CTG ~\cite{survey_ctg}, but it lacks (1) a systematic review of representative application tasks, main approaches, and evaluation methodologies of CTG; (2) a tracking of the latest large-scale PLM-based CTG approaches. In this paper, we provide an introduction to the main tasks and evaluation metrics related to CTG, a dedicated and comprehensive literature review on CTG approaches using PLMs, and finally, an outlook on the possible future research directions. We hope that this survey paper will help  researchers and practitioners to quickly capture the overall picture as well as detailed cutting-edge methods in PLM-based CTG, and promote the further development of this promising area.

The remainder of the paper is organized as follows: Section~\ref{Introduction_PLM} gives a brief introduction to the two critical aspects of the area, i.e., the fundamental concepts of CTG and PLMs. Then, we divide the main approaches to PLM-based CTG into three categories and discuss them in more detail in Section~\ref{approach}. Section~\ref{evaluation} summarizes the relevant evaluation methodologies and metrics for CTG. In Section~\ref{Challenges_and_Future_Direction}, we discuss the challenges that the field is facing and put forward a number of promising future directions. Finally, we conclude the paper in Section~\ref{conclusion}. All the literature appearing in this paper is filtered following two rules. First, we tend to select the latest papers that appeared within 3-4 years, ensuring the timeliness of the surveyed works. Second, we preferably select the works that are influential in the NLP community, e.g.,  the papers published in the leading conferences or journals in the NLP field, such as ACL, EMNLP, NAACL and TACL; and the works that are highly cited or have received widespread attention in the open source community.

\section{An Introduction to Controllable Text Generation and Pre-trained Language Models}
\label{Introduction_PLM}
This paper is closely related to two key aspects: controllable text generation and pre-trained language models, which will be briefly introduced in this section.

\subsection{Controllable Text Generation}
\label{Controllable_Text_Generation}

\begin{figure}[h] 
    \centering %使图片居中显示
    \includegraphics[width=0.4\textwidth]{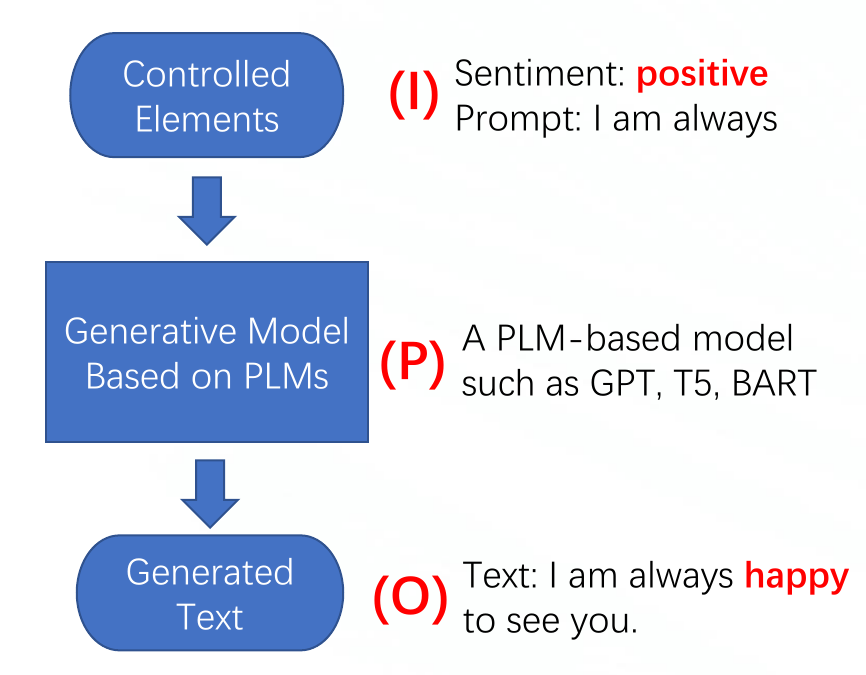} %中括号中的参数是设置图片充满文档的大小，你也可以使用小数来缩小图片的尺寸。
    \caption{The IPO of controlled text generation. A typical CTG system consists of three components: the controlled element (controlled condition and source text) as input (I), the generative model as the process (P), and the generated text satisfying the input control condition as output (O). } %caption是用来给图片加上图题的
    \label{fig:ipo_ctg} %这是添加标签，方便在文章中引用图片。
\end{figure}

Controllable text generation (CTG) refers to the task of generating text according to the given controlled element~\cite{survey_ctg}. As shown in Figure ~\ref{fig:ipo_ctg}, a typical CTG system consists of three components: the controlled element, including a controlled condition (e.g., a positive sentiment) and a source text (which can be vacant or just a text prompt in some applications) as input (I); the generative model (e.g., a PLM-based model) as the process (P), and the generated text satisfying the input control condition, as output (O). Take sentiment control as an example. If we want to generate a sentence with a positive emotion, then the condition ``positive sentiment'' and corresponding prompt ``I am always'' are taken as the control element and input to a PLM-based generative model. The output sentence's sentiment disposition would satisfy the controlled element, such as ``I am always happy to see you''.

%figure环境

Depending on different applications, the attributes of control conditions can be in different forms and connotations. They could range from text attribution (such as sentiment, topic, and keywords); author style and speaker identity of the person writing the text (such as gender and age); text genre and formats (such as poems, couplets); ordering of events (such as storylines); to structured data description (such as table-to-text and Knowledge Graph (KG)-to-text generation). All the above task types can be formalized mathematically in a unified form as follows.

\begin{figure}[h] 
    \centering %使图片居中显示
    \includegraphics[width=0.8\textwidth]{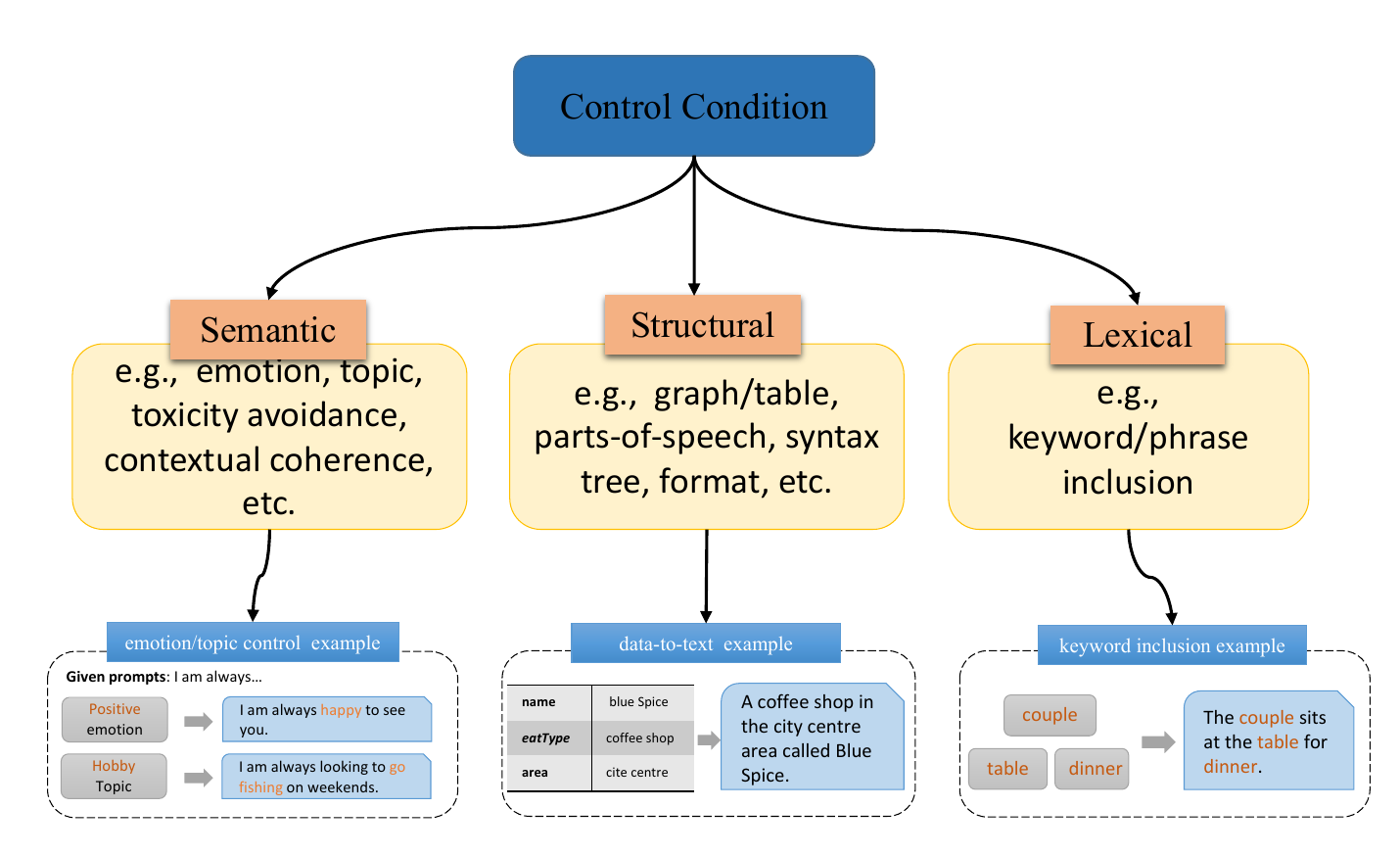}
    \caption{The taxonomy of control conditions. We generally divide the control conditions into three types: (1) Semantic: it generally refers to content control at the semantic level, which is the reverse process of some text understanding tasks, such as emotion/topic classification and toxic detection. (2) Structural: it is opposed to the semantic control, and refers to the control over the structure level of the generated text, which is always related to text structure analysis tasks such as information extraction and sentence parsing. (3) Lexical: it represents a type of controllable text generation task at the vocabulary level, such as generating text that contains certain keywords.}
    \label{fig:control_type}
\end{figure}

Given a vocabulary $\mathscr{V}$, the goal of CTG is to generate a target text $Y = \{y_1, y_2, \dots, y_n \} $, where  $y_n \in \mathscr{V}$, with respect to a control element denoted as $C$. Then CTG can be formally described as:
\begin{equation}
 P(Y|C) = p(y_1, y_2, \dots, y_n | C) 
\end{equation}
 The specific expression of $C$ may vary according to different tasks. We divide the commonly used control conditions into three categories, i.e., semantic, structural, and lexical constraints, as shown in Figure~\ref{fig:control_type}. The sentence $Y$ generated by a CTG model is expected to satisfy the constraint conditions while conforming to the general natural language characteristics, such as fluency, rationality and readability, to the greatest extent.
 
 %figure环境

 \subsection {Tasks and Applications Involving CTG}

%We now introduce typical tasks and applications related to controlled text generation. 
Controllability is the fundamental problem of text generation, which is indeed required by almost all text generation scenes. Here, we only focus on those with explicitly controlled conditions and goals. Table ~\ref{ctg_task} summarizes typical tasks and applications involving CTG, with a description of the input/output, controlled aspects, and representative references. They are explained in more detail below:

\begin{itemize}
   \item \textbf{Attribute-based Generation}: Attribute-based CTG aims to generate natural language sentences that satisfy specific attributes such as topic, emotion, and keywords. Precisely controlling the various attributes of sentences is an essential requirement of intelligent writing. By combining multiple control attributes, the system can, in theory, create interpretable and controllable paragraphs or articles. Thus, attribute-controlled text generation has always been the focus of attention in the field of text generation.
        
    \item \textbf{Dialogue Generation}: The goal of dialogue systems is to build an agent who can mimic human conversations using natural language. Generative dialogue models often have  higher requirements in consistency, semantics, and interactiveness~\cite{huang2020challenges}. Therefore, the constraints on emotion, the speaker's personal style, dialogue intent/action, etc., are used to control the dialogue responses and improve the interactivity of dialogue systems.

    \item \textbf{Storytelling}: Storytelling requires the model to generate texts with a complete narrative logic, which needs a higher level of control on long text generation. Storylines and story endings are often regarded as controlled conditions, and the model needs to produce stories with fluent text and sound plots according to the given controlled conditions.    
    
    \item \textbf{Data to Text}: The main goal of data-to-text generation is to convert non-linguistic structured data (e.g., a table or a graph) into natural language text, which can be applied in tasks like weather forecast, healthcare~\cite{ferreira2019neural}, and so on. The controllability of data-to-text tasks is to ensure that the generated text contains information manifested in the original structure data.

    \item \textbf{Data Augmentation}: Neural networks heavily rely on a large amount of labeled data. Nowadays, the importance of data augmentation is becoming more conspicuous, given the significant cost of data collection and cleaning. Since recent neural network models are capable of generating near-realistic text, it is possible to utilize them to expand existing datasets and even create new data. Identifying and replacing some entities in the given text or generating new sentences according to given attributes through CTG has become an efficient way for data augmentation.

    \item \textbf{Debiasing}: Biased training data may cause a model to learn incorrect knowledge, and accordingly produce biased results. Therefore, text debiasing has attracted increasing attention. Debiasing by rewriting the biased text  or changing the data distribution of CTG-generated text has been shown feasible. The major controllable aspects in this task include gender, race, and toxicity.

    \item \textbf{Format Control Tasks}: There are also CTG tasks that need to control the format of the generated text, such as text length and rhythm. For example, the task of generating traditional Chinese poetry and couplet  has strict requirements in format, including the number of words, structure, etc.

\end{itemize}

\begin{table}[]
\renewcommand\arraystretch{1.5} 
\caption{A general overview of the tasks involving CTG. We enumerate 7 categories of tasks involving CTG. For each category, we briefly describe the input and output, and list the representative references based on controllable aspects.}

\scalebox{0.55}{
\begin{tabular}{|c|l|l|}

\hline
\textbf{Task} & \textbf{Input \& Output} &  \textbf{Controllable Aspects} \\
 \hline
 
Attribute-based Generation & \begin{tabular}[c]{@{}l@{}}Input: Keywords, discrete attributes\\ Output: Attributes-specific sentence\end{tabular} & \begin{tabular}[c]{@{}l@{}}Topic~\cite{wang2019topic,tang2019topic,gdc,pplm}, tense~\cite{logeswaran2018content},politeness~\cite{sennrich2016controlling,ijcai2022p716},\\sentiment~\cite{gdc,pplm,He2020APF,zhang2019emotional,chen2019sentiment,samanta2020fine,DisCup-2022}, \\
keywords~\cite{metion_flag,point_insertion,CBART,non_residual_prompt}\end{tabular}\\ \hline

Dialogue Generation & \begin{tabular}[c]{@{}l@{}}Input: dialogue content, additional structural \\ information(eg: persona, emotion, intent, template, etc.)\\ Output: Dialogue Response\end{tabular} & \begin{tabular}[c]{@{}l@{}}Persona~\cite{acl2020_persona,zhong2020towards, acl2021_persona,aaai2020_presonalized_dialogue_generation,nanal2021_multi-task}\\
Politeness\cite{niu2018polite, polite_trans_css}, Sentiment~\cite{emo_vae,acl_2019_emo_ctg,cikm2019_emo_ctg,EmoSen_2020_tac}, \\ Template\cite{kale-rastogi-2020-template,wang-etal-2021-template}, \\
Ground-truth reference~\cite{qin2019conversing, wu2020controllable,peng-etal-2020-shot,rashkin-etal-2021-increasing}\end{tabular} \\ \hline

Storytelling & \begin{tabular}[c]{@{}l@{}}Input: Story elements\\ Output: Story paragraph\end{tabular} & \begin{tabular}[c]{@{}l@{}} Story structure~\cite{top_k, goldfarb2020content,fang2021outline}, story ending~\cite{ tambwekar2018controllable,luo2019learning},\\ topic~\cite{wang2019keep,yang2019enhancing, lin2021plug,chang2021changing}, persona~\cite{prabhumoye2019my,liu2020character}\end{tabular} \\ \hline

Data to Text & \begin{tabular}[c]{@{}l@{}}Input: Table/graph data\\ Output: The texts describing the data information\end{tabular} & \begin{tabular}[c]{@{}l@{}} Structural information~\cite{puduppully2019data,acl2020_graph2text,plm_amr, acl2020_data2text,structAdapter,findins_acl2021_PlanthenGenerate}  \end{tabular} \\ \hline

Data Augmentation & \begin{tabular}[c]{@{}l@{}}Input: Original text, pre-defined slot values\\ Output: Text that specific features are replaced\end{tabular} & Pre-defined slot values~\cite{malandrakis2019controlled,amin2020exploring,data_boost} \\ \hline

Debiasing & \begin{tabular}[c]{@{}l@{}}Input: Biased text/biased model\\ Output: Unbiased text/unbiased model\end{tabular} &\begin{tabular}[c]{@{}l@{}} political bias~\cite{liu2021mitigating}, gender bias~\cite{dinan2019queens,qian2019reducing}, subjective bias~\cite{pryzant2020automatically}, \\ social bias~\cite{sheng2020towards,barikeri2021redditbias}, sentiment bias~\cite{huang2019reducing}, toxicity~\cite{GeDi,liu-etal-2021-dexperts})  \end{tabular}\\ \hline

Format Control & \begin{tabular}[c]{@{}l@{}}Input: Desired Format, prompt text\\ Output: Text in predefined formation\end{tabular} & \begin{tabular}[c]{@{}l@{}} Format~\cite{liao2019gpt,rigid_format_acl2020,aaai2021_songmass,sent_style_cikm_2021} \end{tabular}  \\ \hline
\end{tabular}
}
\label{ctg_task}
\end{table}

\subsection{Transformer-based Pre-trained Language Models}

\begin{table*}[]
\renewcommand\arraystretch{1.5} 
\caption{An overview of the characteristics of typical PLMs. "MLM" means "Mask Language Model"; "NSP" means "Next Sentence Prediction";  "SLM" means "Standard Language Mode"; "CTR" means "Corrupted Text Reconstruction"; "FTR" means "Full Text Reconstruction"; "PLM" means "Permutation Language Modeling"; and "TLM" means "Translation Language Modeling". The detailed definition of the pre-training task  can be seen in  literature~\cite{pretrain_prompt_survery}.}
\label{PLMs}
\resizebox{\linewidth}{!}{
\begin{tabular}{|c|c|c|c|c|}
\hline
\textbf{Name} & \textbf{Model Type} & \textbf{Infrastructures}  &\textbf{Pre-training Task}  &\textbf{Main Application}   \\ \hline
BERT~\cite{bert}                      & AE        & Encoder                     & MLM+NSP                                   & NLU                 \\ \hline
XLNET~\cite{Xlnet}                    & AR        & Transformer-XL              & PLM                                   & NLU                 \\ \hline
GPT2~\cite{gpt2},GPT3~\cite{gpt3}     & AR       & Decoder                     & SLM                                 &NLG                  \\ \hline
T5~\cite{t5}                   & Seq2Seq   & Encoder+ Decoder            & CTR                           &NLG+NLU              \\ \hline
mBART~\cite{mbart}             & Seq2Seq   & Encoder+ Decoder            & FTR                                &NLG                  \\ \hline
UniLM~\cite{UniLM}            & \begin{tabular}[c]{@{}l@{}}AE+AR+Seq2Seq\end{tabular}& Encoder+Decoder  &  SLM+CTR+NSP              &NLG+NLU              \\ \hline
ERNIE-T~\cite{ERNIE}          &AE     &Encoder                          &CTR+NSP            &NLU                                                    \\ \hline
XLM\cite{XLM}            &Seq2Seq  &  Encoder+ Decoder               &TLM            & NLG                                                    \\ \hline

Bloom\cite{scao2022bloom}            & AR      &  Decoder                     & SLM                     & NLG                                                 \\ \hline
PaLM\cite{chowdhery2022palm}         & AR      &  Decoder                     & SLM                     & NLG                                                 \\ \hline
LLaMA\cite{touvron2023llama}         & AR      &  Decoder                     & SLM                     & NLG                                                 \\ \hline

\end{tabular}
}
\end{table*}

Recent years have witnessed the emergence and successful applications of large scale pre-trained language models (PLMs). They are regarded as a revolutionary breakthrough in deep learning and NLP. During the pre-training stage, the use of large-scale unlabeled data can provide a strong support to an increased model scale and a better grasp of the diverse knowledge (e.g., linguistic knowledge, commonsense, facts, expertise, etc.) in the data. State-of-the-art (SOTA) performance has been achieved in downstream tasks by fine-tuning the PLMs based on only a small amount of supervised data.

The early work related to PLMs can be traced back to NNLM~\cite{nnlm}, word2Vector~\cite{word2vector}, and ELMo~\cite{ELMo}. More recently, the pre-trained models based on Transformer~\cite{attention_is_all_your_need}, a purely attention-based deep neural network, have greatly improved the performance in almost all NLP tasks and become the mainstream. 
Nowadays, the representative PLM infrastructures are mainly based on  Transformer~\cite{attention_is_all_your_need} or its variants such as Transformer-XL~\cite{transformer_xl},  Longformer~\cite{longformer} and Reformer~\cite{reformer}. Thus this paper is focused on Transformer-based PLMs. The objectives of model learning mainly include masked language modeling (MLM), corrupted text reconstruction (CTR), etc. In order for a good understanding of them, we give a summary of the representative PLMs in Table~\ref{PLMs} according to their data construction modes, model infrastructures, pre-training tasks, and main applications. Here, we roughly divide the existing PLMs into the following three categories and provide a brief introduction to each of them.

\textbf{Auto-Encoding (AE) Models}: This type of PLMs is constructed based on destroying the input text in some way, such as masking some words of a sentence and then trying to reconstruct the original text. Typical examples of this type include BERT, ROBERTA, and ERNIE. Because these models aim to build bidirectional encoding representations of the entire sentences, their infrastructures often correspond to the encoder part of Transformer, which does not contain any masked attention mechanism, and all input can be accessed at each location. They can then be fine-tuned in downstream tasks and have achieved excellent results. The natural applications are sentence classification and sequence labeling, etc., which are more inclined to natural language understanding (NLU) tasks.

\textbf{Auto-Regressive (AR) Models}: The main task of AR models is to predict the next word based on what has been read in the text. This is the same as the classical language modeling approach.  A representative of the AR models is the GPT family. Unlike the aforementioned AE language models, the infrastructures of AR are composed of the decoder part of Transformer, and a masking mechanism is used in the training phase so that the attention calculations can only see the content before a word. While it is possible to fine-tune such a PLM and achieve excellent results on many downstream tasks, it is most natural application is NLG tasks. 

\textbf{Seq2seq Models}: The sequence-to-sequence (Seq2Seq) models use both the encoder and decoder of Transformer for better model flexibility. Currently, the most representative models of this type include T5 \cite{t5} and mBART \cite{mbart}. In principle, almost all pre-trained tasks used in AE and AR models can be adapted to the seq2seq models. Relevant research~\cite{t5} has found that seq2seq models can achieve better performance. Moreover, a seq2seq model unifies the NLU and NLG tasks so that they can be solved under the same framework. It can be fine-tuned on a variety of NLG tasks such as translation and summarization, as well as NLU tasks that can be converted into a text2text form~\cite{t5}, including sentence classification, semantic similarity matching, etc.

\textbf{Summary}: Theoretically speaking, Auto-Encoder (AE) and Sequence-to-Sequence (Seq2Seq) models utilize bidirectional attention in Transformer, while Auto-Regressive (AR) language models rely on causal attention. Bi-attention models have been found to encounter low-rank problems~\cite{low_rank}, which may limit their expressive ability to some extent. On the other hand, causal attention possesses greater theoretical expressive power. Therefore, for complex tasks like logical reasoning, or in the scene where the language model is expected to serve as a backbone for Artificial Generative Intelligence (AGI), AR models with a significant number of parameters (e.g., GPT-3~\cite{gpt3}, GPT-4, etc.) are a preferable choice. However, AR models also come with potential limitations for tasks such as  fill-in-the-blank, content comparison, and text summarization, which often require the model to look back, analyze multiple pieces of content, or engage in extensive re-reading~\cite{gpt3}.  Therefore, AE and Seq2Seq models are often better choices for these tasks.

When it comes to controlled text generation using PLMs, most methods exploit the generative model, including AR and Seq2seq models, as the basis, and guide them to generate the desired text. Generally, CTG tasks always treat the PLMs as a conditional generation model, and its formulation is consistent with the standard language model: 
\begin{equation}
 P(x_n|X_{1:n-1}) = p(x_n |x_1, x_2, \dots, x_{n-1} ). 
\end{equation}

Based on the pre-trained language model manifested above, the goal of conditional text generation can be formulated as:
\begin{equation}
%  P(X|C) =\displaystyle\prod_{i = 1}^{n}p(x_n | x_{<i},C ), 
 P(X|C) =\displaystyle\prod_{i = 1}^{n}p(x_i | x_{<i},C ), 
\end{equation}
where $C$ denotes the controlled conditions, which will be integrated into the PLM in a specific form, and $X$ is the generated text that incorporates the knowledge encoded in the PLM and complies with the control conditions.

In the next section, we will review the main approaches to CTG using Transformer-based PLMs.

\section{Main Approaches to PLM-based CTG}
\label{approach}
From a generative point of view, PLMs have learned a variety of knowledge from a large-scale corpus that can help produce  more fluent and a richer variety of text. This provides an effective way for natural language generation. However, the existing PLMs are essentially still black-box models like other deep neural networks, lacking interpretability and controllability of the text generation process. How to make good use of PLMs in text generation while realizing the controllability of the generative model has recently become a hot research topic. In this section, we provide a comprehensive review of the main approaches in this area from the perspective of the Transformer-based PLMs are used for CTG.

\begin{figure*}[htbp]
    \centering %使图片居中显示
    \includegraphics[width=0.8\textwidth]{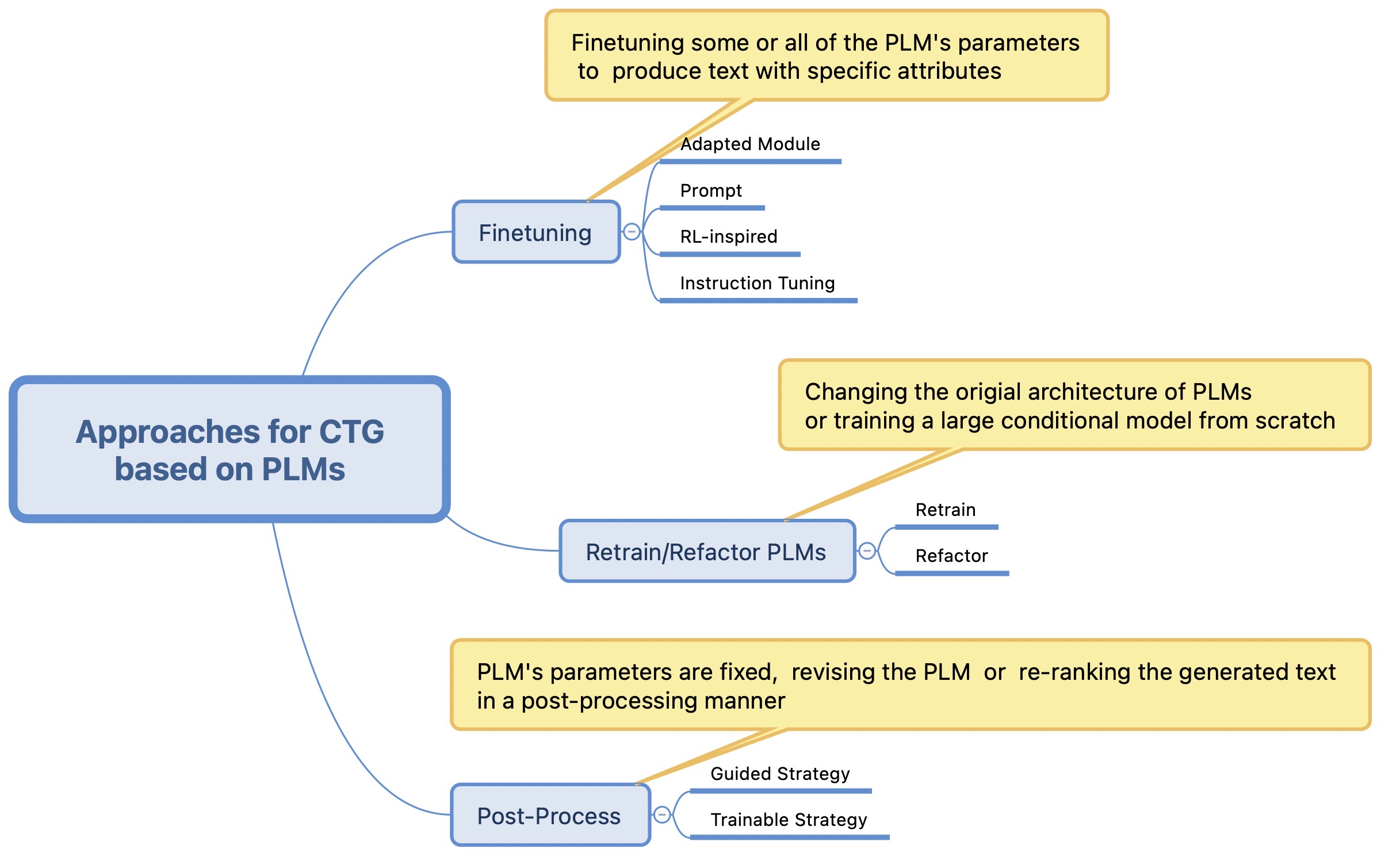} %中括号中的参数是设置图片充满文档的大小，你也可以使用小数来缩小图片的尺寸。
    \caption{An overview of the PLM-based CTG approaches. According to the way how the control signal works with the pre-trained language model, we have roughly divided the existing methods into three categories, each of which is further divided into eight subclasses.} %caption是用来给图片加上图题的
    \label{overview_ctg} %这是添加标签，方便在文章中引用图片。
\end{figure*}%figure环境

\subsection{Overview}
The core idea of PLM-based CTG  is to give the model a control signal in an explicit or implicit way to drive the generation of text satisfying the control conditions. According to the way how the control signal works, we have roughly divided the existing methods into three categories, each of which is further divided into several subclasses. An overview is given in Figure~\ref{overview_ctg}. The most direct way is to \textbf{fine-tune} the PLMs, which can perform the CTG task at a lower cost. The second way is to \textbf{retrain or refactor} the PLMs for CTG. In principle, this method could produce better results  but may consume more computing resources and also face the problem of lacking labeled data. As the parameter size of PLMs increase rapidly, even fine-tuning has become resource-intensive. To tackle the problems, the third category of text generation methods, namely \textbf{post-processing}, that work on the decoder time, named \textbf{post-processing}, have emerged. In the post-processing methods,  PLMs are always fixed, and the control signal works on the decoding-stage. Such methods not only require less computation resources for training, but also can guarantee a better quality of the generated text to some extent. As a consequence, increasing attention from the academic community has been paid to this direction in recent years. In the following sections, we will review the recent literature related to these three types of method in more detail.

\subsection{Fine-tuning}
This type of method aims to fine-tune part or all of the  parameters of a PLM to produce text that satisfies the specific controlled conditions. As discussed in Section 2.3, "PLM + fine-tuning" has become a new paradigm in the general field of NLP. First, a large amount of training data (usually unlabelled samples) and model parameters are used to learn general knowledge from the data and encode the learned knowledge into a PLM. Then a domain/task-adapted model will be obtained to achieve the competitive performance by fine-tuning the PLM based on a small amount of labeled data for the specific downstream task.

This paradigm is also applicable to CTG, and a large number of related studies have been carried out. Recent work has found that fine-tuning PLMs on the target data, such as AMR-to-text~\cite{dart,plm_amr,t2t_plm} for dialogue generation, can establish a new level of performance. While the conventional fine-tuning method is relatively concise and easy to understand, we focus on more advanced methods below.

\textbf{Adapted Module}: This method first constructs a task-related adapted network module around a PLM, and then it is trained with the PLM on the target dataset just like usual fine-tuning. Auxiliary Tuning~\cite{auxiliary_tuning} introduces an extra condition modeling module based on the original PLM, which takes $X(x_n; C)$, the concatenation of control condition $C$ and training text $x_n$, as input, and outputs logits in the vocabulary space. The auxiliary model is trained by adding its logits to the PLM's logits and maximizing the likelihood of the target task's output. In \cite{structAdapter}, an adapter module is added after the feed-forward sub-layer of each layer on both the encoder and decoder of the PLM, which can encode the graph structure into the PLMs without contaminating its original distributional knowledge. During the training stage, the PLM's parameters are frozen, and only the injected adapter is trainable. Avoiding the catastrophic forgetting problem while maintaining the topological structure of the graph, the model achieves the SOTA performance on two AMR-to-text benchmarks.
On the controlled dialogue generation task, the idea of an adapted module is also applied. In ~\cite{aaai2021_Adapter_Bot}, an adapter-bot for dialogue generation is proposed. The model builds a series of lightweight adapters on top of a PLM for dialogue generation, namely, DialoGPT~\cite{zhang2019dialogpt}. The model allows for a high-level control and continuous integration of various control conditions for different conversational requirements (e.g., emotions, personas, text styles, etc.).

In summary, the adaptive modules essentially aim to bridge the gap between the controlled attributes and the PLMs, while guiding the language model to generate text that meets the corresponding control conditions.

\begin{figure}[h] 
    \centering 
    \includegraphics[width=0.5\textwidth]{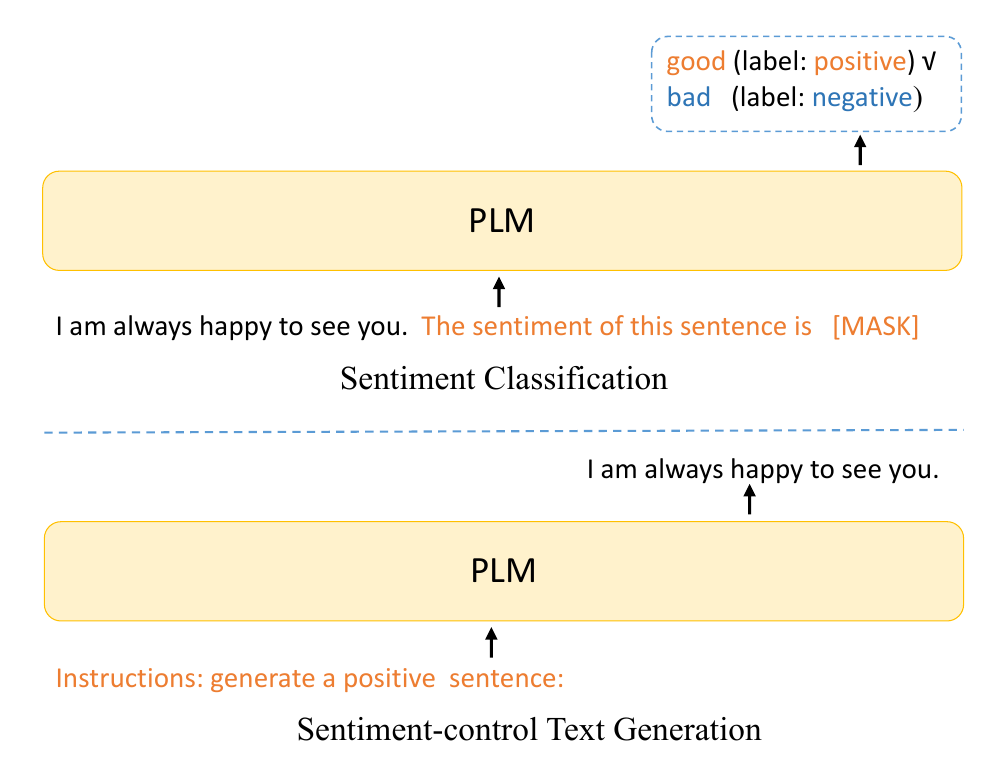} 
    \caption{An illustration of prompt-learning. The top one shows an example of sentiment classification based on prompt learning, while the bottom one shows an example of sentiment-controlled text generation. The text with red color is the templates, which could be either manually constructed by humans, or automatically searched discrete/continuous tokens.} 
    \label{fig:prompt_learning} 
\end{figure}

\textbf{Prompt}: A more effective way for the use PLMs is to keep the training objective of the fine-tuning phase  consistent with the original task from which the PLMs are derived~\cite{gao-etal-2021-making}. This idea gives rise to the so-called prompt-based approaches. Take a sentiment classification task for example. Suppose we need to recognize the sentiment of a sentence, e.g., \textit{"I am always happy to see you".} Different from the traditional approaches that encode the sentence into a set of vectors and then classify their sentiment through a fully connected layer, the prompt-based method will construct a set of templates, for example: (\textit{"I am always happy to see you, \textbf{the sentiment of the sentence is [MASK]}"}), and then ask the model to predict the token \textit{[mask]} according to the original training task for constructing the PLM. For sentiment-controlled text generation, the template could be regarded as the control prompts to instruct the PLM to generate desired texts. A specific illustration of prompt learning can be seen in Figure~\ref{fig:prompt_learning}. The prompt-based approach has gone through various stages, from manual template construction~\cite{TACL2020}, to automated search for discrete tokens~\cite{AutoPrompt}, and to continuous virtual token representations~\cite{p_tuning,prefix_tuning}. Keeping most parameters of the PLM fixed in most cases  and retaining generative capacity of the original PLMs, these methods have achieved great success in zero/few-shot scenarios.
 
From the CTG point of view, the prompt-based approach still applies. In~\cite{prefix_tuning},  a method named “prefix tuning” is proposed, which freezes the PLM's parameters and back-propagates the error to optimize a small continuous task-specific vector called "prefix". The learned prefix, also called "prompt", can guide the PLM to generate the required text, thus enhancing the controllability to a certain extent. The approach achieves impressive results on some generative tasks such as data-to-text. An extension of the model, namely P-tuning~\cite{p_tuning}, serves a similar purpose. Different from prefix-tuning~\cite{prefix_tuning}, P-tuning does not place a prompt with the "prefix" in the input, but constructs a suitable template to prompt the PLM, and the template is composed of  continuous virtual tokens  obtained through gradient descent. Based on prefix-tuning, ~\citet{qian-etal-2022-controllable} leverage contrastive learning to train attribute-specific vectors. Different from the vanilla prefix tuning, where each prefix is trained independently under the attribute-specific corpus, they take into consideration the relationship among attribute prefixes  and train multiple prefixes simultaneously, thereby boosting the control performance. DisCup~\cite{DisCup-2022} provides a new prompt-based alternative for attribute-controllable generation, which uses the attribute-discriminator to assist prompt-tuning. This allows the learned control-prompt to absorb the information of inter-attribute knowledge, achieving the state-of-the-art attribute control performance. In view of the promising effect and the parameter-efficient structure features of the prompt-tuning in CTG, prompt-based multi-attribute control approaches have been proposed. For example, Tailor~\cite{tail_prompt} explores two strategies, including a non-training method based on the concatenation of single-attribute prompts and a training method using an attribute-connector, for prompt-based multi-attribute CTG. 

In order to tackle the problem that the distance between the prompt and the next predicted token correlates negatively with the prompt’s influence power, \citet{inverse_prompt} propose a method called Inverse Prompt. The main idea is to use generated text candidates from the PLM to inversely predict the prompt (topic, Poetry name, etc.) during beam search, so as to enhance the relevance between the prompt and the generated text and achieve a better controllability. However, the generation process requires the reverse prediction for each candidate token, leading to an increased computation cost. According to our actual tests based on the provided source code, it takes up to around 10 minutes to generate a seven-word rhyming poem, making it difficult to be applied in real application scenarios.

More recently, in order to address the challenge of fine-grained CTG, an encoder-decoder architecture based on a pair of GPT-2 models is introduced~\cite{non_residual_prompt}, namely non-residual prompting. It enables intermediate text prompts at arbitrary time steps of the generative PLM. Specifically, the proposed method uses an auxiliary prompt encoder, composed of a trainable GPT-2, to guide another generative language models towards certain constraints, including themes, sentiment, keywords, etc. The generative language model (i.e., GPT-2) is always fixed during generation, and the different prompt instructions can be used at different time steps, enabling fine-grained CTG. Non-residual prompt trends to be versatile and shows promise towards the unified controllable text generation. However, we hold a critical opinion on it, as it lacks a systematic comparison  with the natural encoder-encoder architectures such as T5~\cite{t5} and Bart~\cite{mbart}, and its training process is complex so that it is not parameter-efficient.

To summarize, most of the prompt-based methods show a certain degree of versatility. From the CTG perspective, this kind of method essentially uses the characteristics of PLM in its pre-training stage to guide the PLM to generate constrained text by selecting an appropriate prompt in the fine-tuning stage, so as to achieve the purpose of controllability.

\textbf{Reinforcement Learning (RL) inspired Approaches}: The core motivation of this type of method is to feed back whether or how the control conditions are achieved as a reward to the fine-tuning of the PLM. ~\citet{ft_human_preference} use reinforcement learning to fine-tune the PLMs, with a reward model trained from human preferences. First, it initializes a policy $\pi = \rho$, where $\rho$ denotes a PLM such as GPT2. Given a dataset $\mathcal{D}\in (X,Y)$, the goal is to fine-tune $\pi$ so that it can approximate the distribution of the data $\mathcal{D}$. This is done using RL by optimizing the expectation of the reward:
\begin{equation}
\mathbb{E}_{\pi}[r]=\mathbb{E}_{x \sim \mathcal{D}, y \sim \pi(\cdot \mid x)}[r(\pi(x), y)].
\end{equation}

Then the reward model $r$ is  trained  based on the  sample $(x,y_0,y_1,y_2,y_3)$ via $x \in \mathcal{D}$, and $y_i$ is generated from $p(y_i|x)$. Human labelers are required to choose the human-preferred sentence from $(y_0,y_1,y_2,y_3)$. To prevent $\pi$ from moving too far away from the original PLM $p$ for ensuring the fluency of the generated text to the greatest extent, a penalty item is added to the reward function during the actual process of fine-tuning $\pi$:
\begin{equation}
R(x, y)=r(x, y)-\beta KL (\pi,p),
\end{equation}
where $R(x, y)$ is the re-defined reward function, $\beta$ is the regular coefficient, and $KL (\pi,p)$  aims to ensure that the two distributions are as close as possible.

\citet{data_boost} propose a data augmentation approach, which uses reinforcement learning to guide the GPT-2 model to generate texts towards a specified conditional direction (i.e., the target class). Specifically, an additional RL stage is added between the softmax and argmax functions of GPT2, and then the parameter of the PLM's hidden states $\theta$ is updated towards the target label according to the signal of the RL reward. The generated texts are regarded as augmentation data to help improve the classification performance. Moreover, \citet{lshf_nips2020} use an RL-based approach on the task of English summarization, which fine-tunes the PLM by combining with human feedbacks.

The reinforcement learning is also applied for controllable story generation. \citet{tambwekar2018controllable} designs a reward-shaping technique that produces intermediate rewards at all different time steps, which are then back-propagated into a language model in order to guide the generation of plot points towards a given goal. It should be noted that the above work is carried out on a language model based on LSTM, but its principles are applicable to the subject described in this article, so that we have included it here. 

In summary, the idea of applying reinforcement learning to PLM-based CTG is natural. The central challenge is to ensure that the PLM is optimized towards the RL's rewards while maintaining the fluency of the generated text. To address this challenge, the key is to achieve a better balance between these two aspects.

\textbf{Instruction Tuning}: Recently, a new PLM-based CTG paradigm, namely instruction tuning, has become popular. Instruction tuning provides an avenue to align the language models with user intents, i.e.,  controlling language models to generate the content that complies with human instructions.

Google Research proposes FLAN~\cite{flan}, which stands for \textbf{F}inetuned \textbf{La}nguage \textbf{N}et. It involves fine-tuning a large language model on a mixture of more than 60 NLP datasets, where each task is expressed through natural language instructions. The results demonstrate that language models are capable of performing tasks described purely through human instructions and can generalize to previously unseen tasks through instruction tuning. Continuing the work of FLAN,  ~\citet{chung2022scaling} further explore  scaling up the number of tasks and model size beyond what FLAN has achieved. They fine-tune the language models on the dataset mixed with chain-of-thought data, showcasing the strong few-shot performance of instruction tuning.

InstructGPT~\cite{instructgpt}, a most notable recent work, utilizes instruction tuning to control the language model and generate desired human-like content. It starts with collecting a dataset of labeled demonstrations of the desired model behavior, which are then used as instructions to fine-tune GPT-3. This allows for controlling the model to generate answers that align with human expectations. In term of the optimization algorithm, InstructGPT leverages reinforcement learning from human feedback~\cite{lshf_nips2020, ft_human_preference}, as discussed in the previous section. The results demonstrate that fine-tuning with human feedback is a promising approach to aligning language models with human intents, leading to an improved performance in truthfulness and a reduction of toxic output.

In summary, instruction tuning enables PLMs to understand human intents in a natural language format, offering a promising approach for more general and effective CTG. However, instruction tuning requires careful design of human-labeled prompts. How to fully and safely align the  human instructions with PLMs remains an open problem~\cite{instructgpt}, which demands a further exploration.

\subsection{Retraining/Refactoring}
According to the characteristics of a specific downstream task, it is also feasible to change the original architecture of PLMs or retrain a large conditional language model from scratch. This kind of approach is promising to substantially improve the quality and controllability of text generation, but is limited by increased computing resource consumption and the lack of sufficient labeled data. 

CTRL~\cite{CTRL} is an early attempt in this direction. It trains a language model conditioned on a variety of control codes. The network model used in this approach is also  the commonly used Transformer, and a piece of control code (domain, style, topics, dates, entities, relationships between entities, etc.) is added in front of the text corpus. That is, it transforms the original language model $p\left(x_{i} \mid x_{<i}\right)$  into  $p\left(x_{i} \mid x_{<i}, c\right)$. A language model with 1.63 billion parameters is retrained on a 140Gb corpus. Another contribution of this work is to propose a new top-k sampling algorithm:
\begin{equation}
p_{i}=\frac{\exp \left(x_{i} /(T \cdot I(i \in g))\right.}{\sum_{j} \exp \left(x_{j} /(T \cdot I(j \in g))\right.} \quad I(c)=\theta \text { if } \mathrm{c} \text { is True else 1},
\end{equation}
where $g$ is a list of generated tokens, $p_i$ is the probability distribution  for the next token, and the introduction of $I(c)$ reduces the probability of words that have already appeared.

\citet{point_insertion} propose POINTER, an insertion-based method for hard-constrained (i.e., making the specific words appear in generated text) text generation. Different from the auto-regressive method, such as GPT-2, this method modifies the structure of the Transformer so that it can generate text in a progressive manner. Specifically, given certain lexical constraints, POINTER first generates the constrained words to satisfy the control conditions, then more detailed words are inserted at a finer granularity between those words. The above process iterates until the entire sentence is completed. This kind of method can ensure the generated sentences meet the lexical constraints. However, the model needs to be trained from scratch on the large-scale corpus, and the fluency of the generated sentences is not as good as the auto-regressive model in most cases.

Similar to the afore-mentioned insertion-based method~\cite{point_insertion},  a lexically constrained text generation framework called Constrained BART (CBART) is proposed~\cite{CBART}. This approach also adopts the  progressive insertion/replacement for text generation, yet without modifying the Transformer's architecture. Concretely, based on the pre-trained model BART, it divides the generation process into two steps. First, a token-level classifier is added on BART's encoder to predict where to replace and insert. Then, the predicted results are regarded as signals to guide the decoder to refine multiple tokens of the input in one step by inserting or replacing tokens before specific positions. Different from the normal way of generating texts step by step, the decoder predicts all tokens in parallel so as to accelerate the inference. Although CBART does not need to reconstruct the architecture of PLMs,  the training and inference processes are different from the original pre-training tasks, which can lead to a negative impact on the quality of text generation.

CoCon (Content-Conditioner)~\cite{CoCon} introduces a conditional control module in addition to the original PLM, which can realize the precise control of the generated text at the word- and phrase-levels. In terms of model architecture, the approach injects a control block into the GPT model and provides the control code as a separate input. In order to tackle the problem of lacking labeled data, it adopts self-supervised learning and constructs four different loss functions, including Self Reconstruction Loss, Null Content Loss, Cycle Reconstruction Loss, and Adversarial Loss. The core of these self-supervised losses is to use one part of a piece of text as a control condition, leaving the rest for the model to refactor, so that the model can learn to generate specific text conditioned on the control code. The experimental results show that CoCon can incorporate the condition content into the generated texts and control the high-level text attributes in a more flexible way. Similar to CoCon's idea of injecting an additional controlled module to an existing PLM, \citet{metion_flag} propose a Mention Flags (MF) module, which is injected into the decoder of Transformer, to achieve a higher level of constraint satisfaction. The MF is designed to trace whether a lexical constraint has been realized in the decoder's output. It is formally represented as mentioned status embedding injected into the decoder of Transformer, to provide a signal to encourage the generative model to satisfy all constraints before generation. 

This type of approach is also used for the task of controlled dialogue generation. \citet{aaai2020_presonalized_dialogue_generation} propose a PLM-based method to build a personalized dialogue agent. The whole framework is in an encoder-encoder (Transformer) fashion, and its initial parameters inherit from an existing PLM model. The personalized information is represented as attribute embedding, which is added into the encoder to capture rich persona-related features when modeling dialogue histories. Further, an attention routing network is added to the decoder to incorporate the target persona in the decoding process while maintaining the trade-off of the historical dialogue information dynamically. To solve the problem of lacking labeled data in conditional dialogue generation, a multi-task learning framework is proposed~\cite{nanal2021_multi-task}, which utilizes both conditional labeled dialogue data and non-dialogue text data. Based on a condition-aware Transformer block (reconstructed from the original Transformer), three sub-tasks are designed based on the existing PLM, namely, conditional text generation based on labeled dialogue data, conditional conversation encoder, and conditional dialogue generation task based on non-dialogue text to optimize the model simultaneously. Persona and topic-controlled experiments are conducted under the scenario of dialogue generation, and the results show that this approach  achieved a then-state-of-the-art performance.

 In summary, the refactoring or retraining approaches are more convenient to use, but may lose the original PLM's versatility to some extent. As for the methods that need retraining, they may face the dual challenges of increased computation cost and the lack of large-scale labeled data. 

\subsection{Post-Processing}

\begin{figure}[h] 
    \centering %使图片居中显示
    \includegraphics[width=0.5\textwidth]{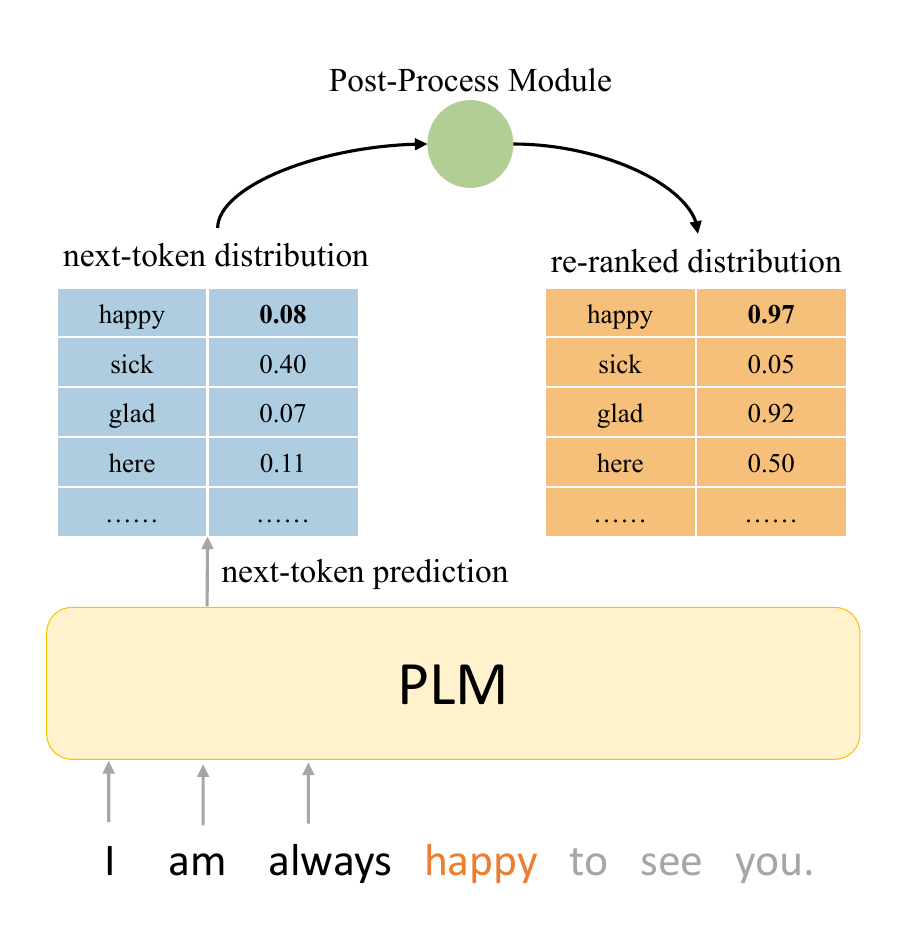} %中括号中的参数是设置图片充满文档的大小，你也可以使用小数来缩小图片的尺寸。
    \caption{The scheme of the post-process approaches for CTG. During the generation, the parameters of the PLM are fixed. Given the input prompt "I am always", the PLM will produce the original next-token distribution; the post-process module aims to re-rank the token distribution, allowing the PLM to choose the sentiment-desired tokens (i.e., significantly increasing the probability of "happy"), so as to achieve the positive sentiment control.} %caption是用来给图片加上图题的
    \label{fig:post_process} %这是添加标签，方便在文章中引用图片。
\end{figure}

  When the number of parameters of a PLM increases, the model has memorized more and more knowledge and patterns, allowing it to achieve competitive results even without fine-tuning in many NLP tasks~\cite{gpt3}. In the realm of controlled text generation, the idea of fixing the PLM's parameters first and re-ranking the generated text in a post-processing manner becomes achievable and promising.  Figure~\ref{fig:post_process}, illustrates the use of a post-process module for sentiment-control text generation.

%figure环境

The most realistic idea about the post-processing method is to use some standard decoding algorithms in text generation, e.g., the Greedy search, constraint beam search~\cite{beam_search}, Top-k sampling~\cite{top_k}, Nucleus sample~\cite{Nucleus}, etc. The approaches discussed below can be seen as an extension of them for CTG tasks. They are grouped into two categories: guided strategies and trainable strategies.

\textbf{Guided Strategies}: This type of method decouples the PLMs for text generation and the post-processing module, and the post-processing module guides the PLM to generate conditioned text only in the inference stage.

A representative method of this type is PPLM~\cite{pplm}. It first trains an attribute discriminant model and then uses it to guide the PLM to generate the corresponding text. In this work,  the attribute model is a simple classifier, consisting of a user-specified bag of words or a single learning layer whose parameters are 100,000 times less than the PLM. During the text sampling process, it requires a forward and backward process in which the gradient from the attribute model drives the hidden activation of the PLM to guide the target text generation. PPLM does not need to change the structure or retrain the PLM, and it is able to achieve a significant improvement in attribute alignment. However, it causes a slight decrease in text fluency measured with the metric of PPL (Perplexity~\cite{liu-etal-2021-dexperts, GeDi}). 

MEGATRON-CNTR~\cite{MEGATRON-CNTRL} is a controllable story generation framework that combines external knowledge and PLM. Given a story context, a predictor is used to get a set of keywords for the next sentence. Then, a knowledge retriever is introduced to get external knowledge-enhanced sentences from an external knowledge base according to the keywords. Next, a ranker is trained to choose the most relevant knowledge-enhanced sentences, which are later fed into the generator of PLM (GPT-2) with the story context to get the next sentence. The entire process is repeated until completing a story. Human evaluation results show that up to 91.5\% of the generated stories are successfully controlled by the keywords. In this framework, GPT-2 is independent of the other modules, and generates context-relevant sentences using the introductory text provided by MEGATRON-CNTR's component as input, without any adaptation in training.

 ~\citet{pair_emnlp2020} propose a content-controlled text generation framework, namely FAIR. It uses the BERT~\cite{bert} model to automatically construct a content plan, including keyword assignments and their corresponding sentence-level positions. After that, the BART~\cite{mbart} model is applied, without structure modification, to fill the masked tokens appearing in the generated text template. Finally, an iterative refinement algorithm that works within the sequence-to-sequence (seq2seq) models is designed to improve generation quality with flexible editing. The reported experimental results show that FAIR can significantly improve the relevance and coherence between the key phrases and the generated texts.

Additionally, a series of discriminator-guided approaches have been developed, which train an attribute discriminator to help PLM select text for the specific attributes at the decoding stage. Adversarial Search~\cite{DAS}, inspired by GAN (Generative Adversarial Network),  trains the discriminator to distinguish human-created text from the machine-generated text. The discriminator predicts a label for each token instead of for the entire sequence. Its logit probability is added to the score to guide sampling towards the human-written style. For the tokenizer with 10 thousand words, decoding using a discriminator to classify each token is time-consuming. Aimed at solving this problem, GeDi~\cite{GeDi} trains a small class-conditional language model (CC-LM) as generative discriminators to guide the generation from large PLMs (GPT-2 and GPT-3). Specifically, a CC-LM is trained and formulated as the following equation:
\begin{equation}
P_{\theta}\left(x_{1: T} \mid c\right)=\prod_{t=1}^{T} P_{\theta}\left(x_{t} \mid x_{<t}, c\right),
\end{equation}
where $c$ is the control code, $T$ is the length of generated text.  GeDi assumes that there is a CC-LM with the desired control code $c$ and an undesired or anti-control code $\bar{c}$, and then uses the contrast between $ P_{\theta}\left(x_{1: T} \mid c\right)$ and $P_{\theta}\left(x_{1: T} \mid \bar{c}\right)$ to guide sampling from the original PLM. A contrast mechanism is designed to compute the probability that every candidate token $x_t$ belongs to the desired class, given by $P_{\theta}(C\mid x_t, x <t)$:
\begin{equation}
P_{\theta}\left(c \mid x_{1: t}\right)=\frac{P(c) \prod_{j=1}^{t} P_{\theta}\left(x_{j} \mid x_{<j}, c\right)}{\sum_{c^{\prime} \in\{c, \bar{c}\}} \prod_{j=1}^{t} P\left(c^{\prime}\right) P_{\theta}\left(x_{j} \mid x_{<j}, c^{\prime}\right)},
\label{conditional_contrast}
\end{equation}
where $P(c)$  and $P(c^{\prime})$ are biased parameters which could be learnt or set manually as a hyper-parameter. During the generation step for a token $x_t$, Equation~\ref{conditional_contrast} is multiplied with the conditional probability $ P_{L M}\left(x_{t} \mid x_{<t}\right)$ of the original PLM via the Bayes rule:

\begin{equation}
P_{w}\left(x_{t} \mid x_{<t}, c\right) \propto P_{L M}\left(x_{t} \mid x_{<t}\right) P_{\theta}\left(c \mid x_{t}, x_{<t}\right),
\label{bays_rules}
\end{equation}
where $P_{w}\left(x_{t} \mid x_{<t}, c\right)$ is regarded as the final probability for text generation. Since the calculation of the Equation~\ref{conditional_contrast} only needs two parallel forward passes of CC-LM, the generation efficiency is greatly improved.

Inspired by the  GEDI~\cite{GeDi}, a series of similar approaches have emerged. DEXPERTS~\cite{liu-etal-2021-dexperts} re-ranks the predictions of the PLM based on expert (and anti-expert) opinions during the decoding stage to steer the language model towards generation of the desired text. FUDGE~\cite{FUDGE} learns an attribute predictor that operates on a partial sequence to adjust the original PLM's probabilities, and achieves an improved performance on the tasks of couplet completion in poetry, topic control in language generation, and formality change in machine translation. Plug-and-Blend~\cite{lin2021plug} extends the GEDI model to controlled story generation by introducing a planner module.

More recently, \citet{word_simility_emnlp2021_findings} propose a simple yet efficient plug-and-play decoding method, namely K2T, which even does not need a discriminator. Specifically, given a topic or keyword which is considered a hard constraint, K2T adds a shift to the probability distribution over the vocabulary towards the words which are semantically similar to the target constraint word. The shift is calculated based on word embedding. Although the K2T is intuitive, the shift added to the probability distribution of vocabulary may be too rough and cause the generated texts to fall short in fluency.

\citet{mireshghallah-etal-2022-mix} propose a score-based controllable text generation framework, namely Mix and Match, which regards the task of generating attribute-specific text as a Metropolis-Hastings sampling process. Specifically, the proposal samples are first produced by the MLM model(i.e., BERT). Then, an energy-based model, which is a linear combination of scores conditioning context from the different black-box experts that represent fluency, the control attribute and faithfulness, respectively, is used to accept/reject the proposal sample. Those two steps are iterated until the desired texts are obtained. The framework is training-free, without any fine-tuning or structural assumptions. Nevertheless, Mix and Match is an iteration-based method. It takes almost 11 seconds to generate a sequence of length 20, which reduces the practical value. Similarly, COLD Decoding~\cite{COLD_Decoding}, an energy-based and iterative CTG approach, formulates the controlled text generation task as sampling from an energy-based model using Langevin Dynamics, without the need for any task-specific fine-tuning. Like the Mix and Match method mentioned earlier, COLD Decoding also suffers from the efficiency issue in text generation.

To sum up, the idea of "guided strategies" is simple and flexible.  The main advantage of this approach lies in the separation of the post-processing module from the model. When the number of parameters of the PLM increases, such advantage becomes more apparent. However, post-process requires multiple iterations to achieve better control performance, resulting in excessive time costs.

\textbf{Trainable Strategies}: 
The Trainable Strategies also work in the inference phase, but different from the Guided Strategies, the extra processing module needs to be trained jointly with PLM whose parameters are fixed. Compared to the prompt-based approaches, the module controls PLM's generation process noninvasively, without disturbing the model's original textual stream.

An Energy Based Model (EBM)~\cite{ebm_ctg} is proposed to guide the PLM to generate desired text. The generative model is formalized as follows:
\begin{equation}
P_{\theta}(x) \propto P_{L M}(x) \exp \left(-E_{\theta}(x)\right),
\end{equation}
where $P_{L M}(x)$ is a local normalized language model whose parameters are frozen during training, and $E_{\theta}(x)$ is an energy function that is aimed at steering the joint model $P_{\theta}(x)$ towards the desired data distribution. The Noise Contrastive Estimation (NCE) algorithm is used to train the model to cope with the intractability issue of the energy model. Experiments show that the proposed method yields a lower perplexity compared with locally normalized baselines on the task of generating human-like text. The use of large-scale PLM makes this method possible, because the quality of the generated text from the joint model relies heavily on the quality of the underlying language model.

Furthermore, since RL-based methods may lead to the problem of "degeneration" in the sense of producing poor examples that improve the average reward but forgo the coherence
and fluency, a distributional approach for controlled text generation~\cite{gdc} was proposed to solve the problem. It uses the Energy-based Model (EBM) to represent the point-wise and distributional constraints in one go:

\begin{equation}
p(x) \doteq \frac{P(x)}{Z},
\end{equation}

\begin{equation}
Z \doteq \sum_{x} P(x),
\end{equation}

\begin{equation}
P(x) = a(x) e^{\sum_{i} \lambda_{i} \phi_{i}(x)},
\end{equation}
where $p(x)$ is the desired normalized distribution, $Z$ is the normalized term, $\phi_{i}(x)$ represents the constraint judgment function (point-wise it means 0 or 1, distributional-wise it may be a continuous value between 0 and 1), and $\lambda_{i}$ is the corresponding coefficient estimated by using Self Normalized Importance Sampling (SNIS), $a(x)$ is the original PLM. As it is hard to calculate $p(x)$ directly, a method similar to variational inference is adopted to approximate the distribution, by first initializing  a policy $\pi = a(x)$, and then minimizing the cross entropy between the policy $\pi$ and the desired normalized distribution $p(x)$:
\begin{equation}
C E\left(p, \pi_{\theta}\right)=-\sum_{x} p(x) \log \pi_(x).
\end{equation}
The optimization process adopts the KL-Adaptive distributional policy gradient (DPG) algorithm to make $\pi$ approximate the model $p(x)$ that satisfies the constraints. The method unifies the point-wise and distributional-wise constraints in a single framework, and the experimental results show its superiority in satisfying the constraints while avoiding degeneration. However, this approach suffers from the high computational cost, which needs to be addressed in the future.

Generally speaking, Trainable Strategies require the post-process module to be jointly trained on the basis of PLM, and realize controllable text generation by adjusting the original probability distribution of PLM to the desired data distribution using the trained post-process module. This type of method build upon the probability modeling theory and thus has a good theoretical basis, yet they are still at the early stage and need to improve the issues related to computational efficiency and text quality.

\begin{table*}[]
\renewcommand\arraystretch{1.5} 
\caption{ A summary of the surveyed CTG methods, which are divided into three main categories and eight sub-categories. We list the main characteristics for each  categories from a macro perspective, and representative references for each  sub-category.}
\label{Approach summary}
\resizebox{\linewidth}{!}{
\begin{tabular}{|c|l|c|l|}
\hline
\textbf{Method}                 & \multicolumn{1}{c|}{\textbf{Main characteristics}}  & \multicolumn{1}{c|}{\textbf{Subcategory}}        & \multicolumn{1}{c|}{\textbf{Typical References}}                        \\ \hline
\multirow{4}{*}{Fine-tuning}    & \multirow{3}{*}{ \begin{tabular}[c]{@{}l@{}}  standard training;\\ efficient inference;\\ higher text quality; \\ weaker controllability  \end{tabular}}  & Adapted Module                                                            &  \cite{auxiliary_tuning,structAdapter,aaai2021_Adapter_Bot}       \\ \cline{3-4}
                                &                                               & Prompt                                                                                                  &  \begin{tabular}[c]{@{}l@{}}\cite{TACL2020,AutoPrompt,p_tuning, qian-etal-2022-controllable} \\ \cite{prefix_tuning, DisCup-2022,non_residual_prompt,tail_prompt} \end{tabular}     \\  \cline{3-4}
                                &                                               & Reinforce Learning                                                                                      & \begin{tabular}[c]{@{}l@{}} \cite{ft_human_preference, lshf_nips2020,data_boost,tambwekar2018controllable,instructgpt}       \end{tabular}    \\ \cline{3-4}

                                 &                                               & Instruction Tuning                                                                                      & \begin{tabular}[c]{@{}l@{}} \cite{instructgpt}    \end{tabular}    \\ \hline

\multirow{2}{*}{Refact/Retrain} & \multirow{2}{*}{\begin{tabular}[c]{@{}l@{}}  computationally expensive  training;\\ higher text quality; better controllability \end{tabular}}                                                  & Retrain       & \cite{CTRL},\cite{point_insertion}        \\  \cline{3-4}
                                &  & Refact                                                                                                    & \begin{tabular}[c]{@{}l@{}} \cite{point_insertion,CoCon,metion_flag,CBART,aaai2020_presonalized_dialogue_generation,nanal2021_multi-task}      \end{tabular}     \\ \hline
\multirow{2}{*}{Post-Process}   & \multirow{2}{*}{\begin{tabular}[c]{@{}l@{}}  efficient training and inefficient inference;\\ lower text quality; better controllability \end{tabular}}    & Guided Strategy                                                                                           & \begin{tabular}[c]{@{}l@{}} \cite{pplm,MEGATRON-CNTRL,DAS,GeDi,liu-etal-2021-dexperts,FUDGE} \\ \cite{lin2021plug, word_simility_emnlp2021_findings, mireshghallah-etal-2022-mix, COLD_Decoding}  \end{tabular} \\  \cline{3-4}
                                &   & Trainbale Strategy                                                                                           & \cite{ebm_ctg,gdc}            \\ \hline
\end{tabular}

}
\end{table*}

\subsection{Summary}

In this section, we divide the current PLM-based CTG approaches into three categories according to the way how PLMs are used. For each category, we analyze its main principles, process, and methods. A summary of the surveyed CTG models is shown in Table~\ref{Approach summary} . "Fine-tuning"  is a more general type of method that has been widely used in both NLU and NLG tasks. How to make full use of the power of PLMs in specific tasks is still a  hot research topic for future research. The retraining or refactoring approaches typically involve high training costs and the lack of large-scale labeled data. To overcome these limitations, combining the general pre-trained models with  semi-supervised or self-supervised learning to build a pre-trained model dedicated to CTG, would be a feasible direction for future research.

The emergence of post-processing methods is rooted on the powerful text-generation capabilities of PLMs. This kind of method generally assumes that the PLMs can produce high-quality text, and then use a post-processing module as a filter to screen the desired type of text. Since the post-process modular is usually decoupled from the PLMs, most current decoding-time approaches (post-process) are still computationally expensive (i.e., with longer inference time or additional parameters), and the quality of generated text can be low. However, it has some promising advantages, because the parameters of the PLMs do not need to be retrained, thus greatly saving computing resources for the model training stage. In recent years, the scale of pre-trained models has been getting larger, and their mastery of language knowledge is getting more comprehensive. At the same time, the sheer size of parameters makes the PLMs resource-intensive to fine-tune and retrain. The above-mentioned trends coincide perfectly with the advantages of the "post-process" methods. Thus it has a great potential for future research and development.

\section{Evaluation Methods}
\label{evaluation}
% After building an  model, we need to assess its performance, for which various evaluation methods can be used. 
The performance of a NLG model is reflected by suitable evaluation metrics. CTG is slightly different from the general NLG tasks due to the need to fulfilling the controlled elements. Therefore, CTG is concerned about not only the quality of the generated text but also the satisfaction with the controlled elements. As a consequence, we usually use both general and CTG-specific metrics to evaluate a CTG model.

\subsection{General NLG Evaluation Metrics}

\begin{figure*}[h]
    \centering
    \includegraphics[width=0.8\textwidth]{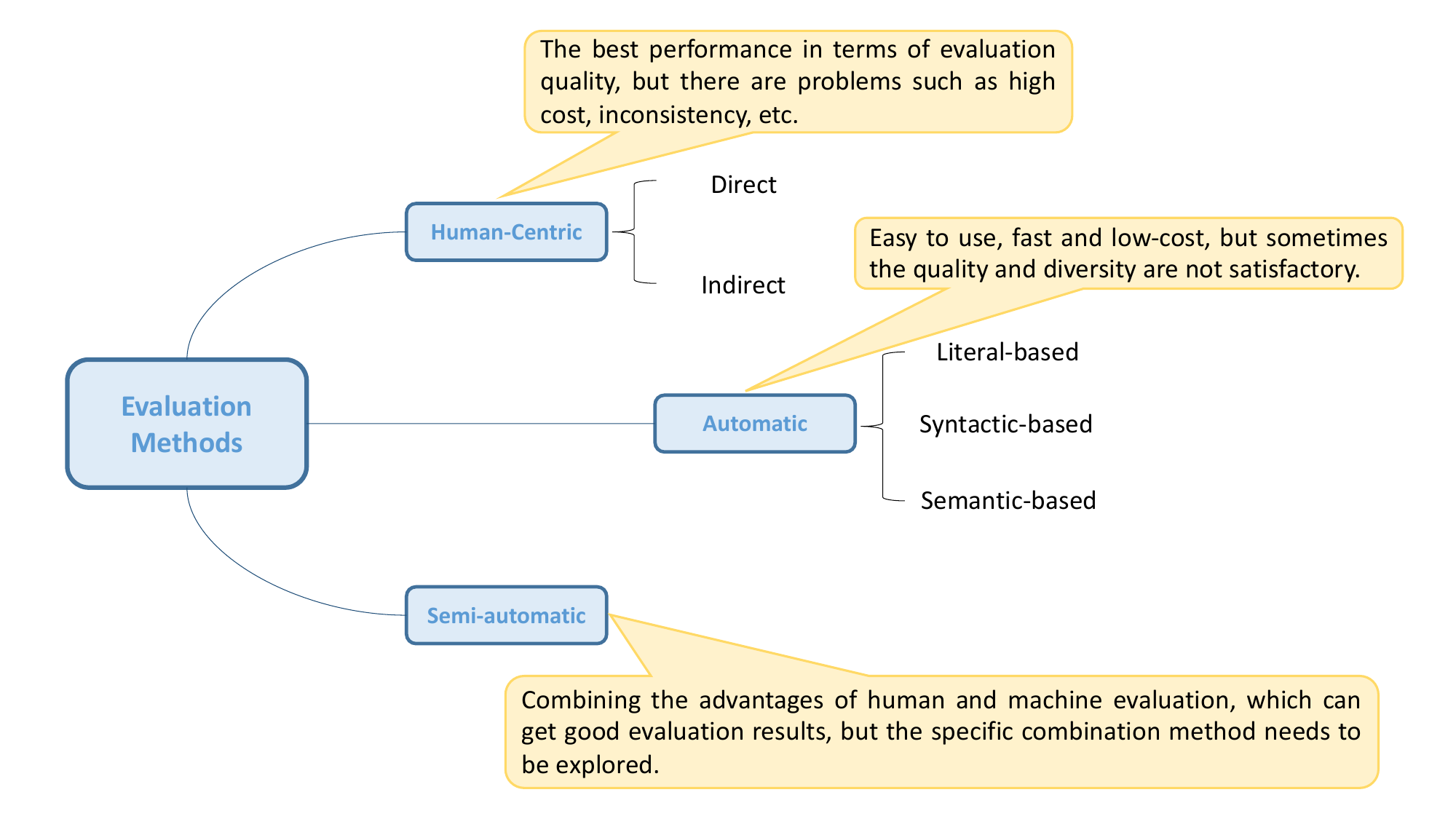}
    \caption{A categorization of general NLG evaluation methods.}
    \label{fig1}
\end{figure*}

For any CTG model, it is essential to evaluate the quality of generated text in general aspects such as: 1) \textbf{fluency}: how fluent the language in the output text is \cite{celikyilmaz2018deep, du2017learning}, 2) \textbf{factuality}: to what extent the generated text reflects the facts described in the context \cite{Nucleus, welleck2019neural}, 3) \textbf{grammar}: how grammatically correct the generated text is, 4) \textbf{diversity}: whether the generated text is of a diverse range of  types or styles.  The ways of measuring these general evaluation aspects can be divided into three categories based on who perform the assessment: human beings or the machine (as shown in Figure~\ref{fig1}).

\subsubsection{Human-Centric Evaluation Metrics}

\begin{figure}[h]
    \centering
    \subfigure[An example of Likert scale \cite{celikyilmaz2020evaluation}. In this example, human annotators need to rate the informativeness (an index  measuring whether the generated text provides all the useful information from a given meaning representation) with a score ranging from 1 to 5.]{
    \label{fig4a}
    \includegraphics[width=0.25\textwidth]{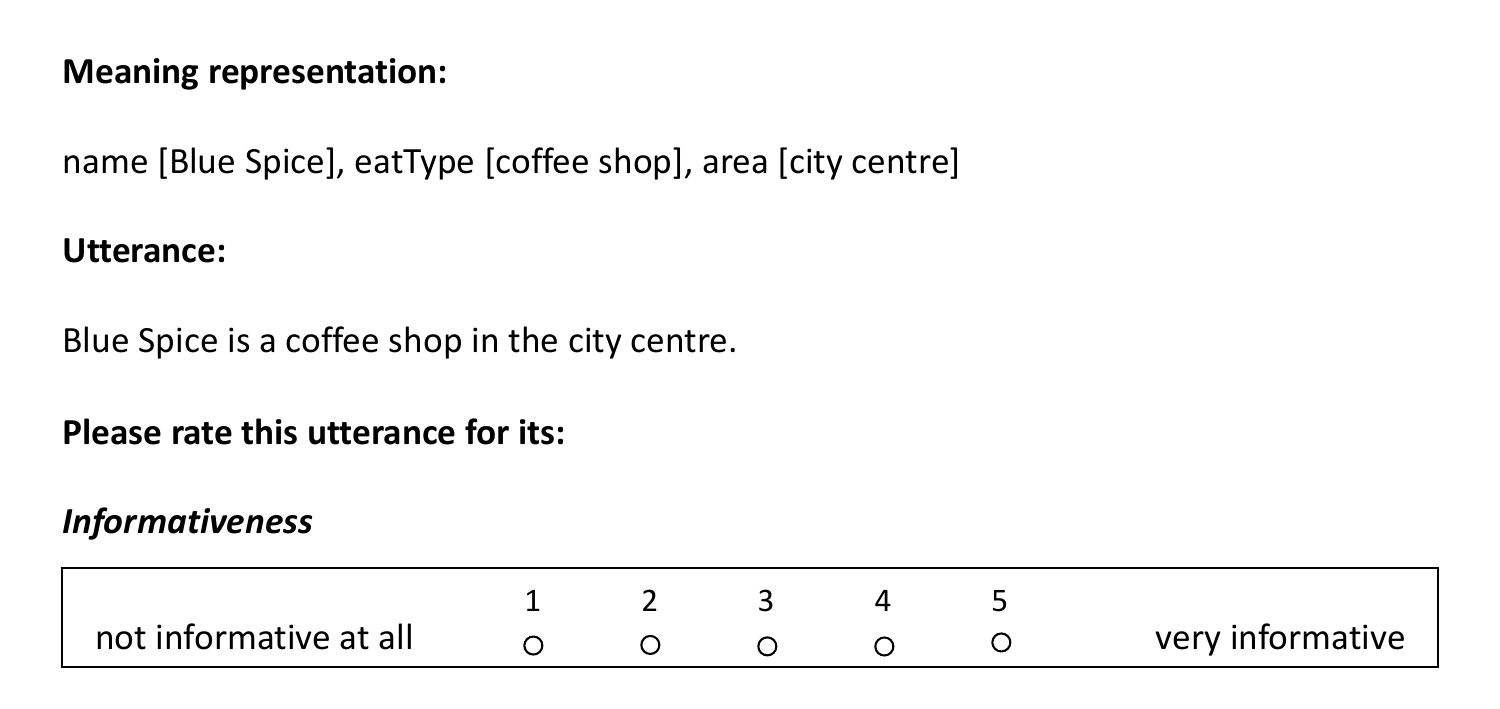}
    }
    \subfigure[An example of RankME \cite{celikyilmaz2020evaluation}. Human annotators need to give a score of the informativeness in a given range for each utterance in this example based on the given meaning representation.]{
    \label{fig4b}
    \includegraphics[width=0.85\textwidth]{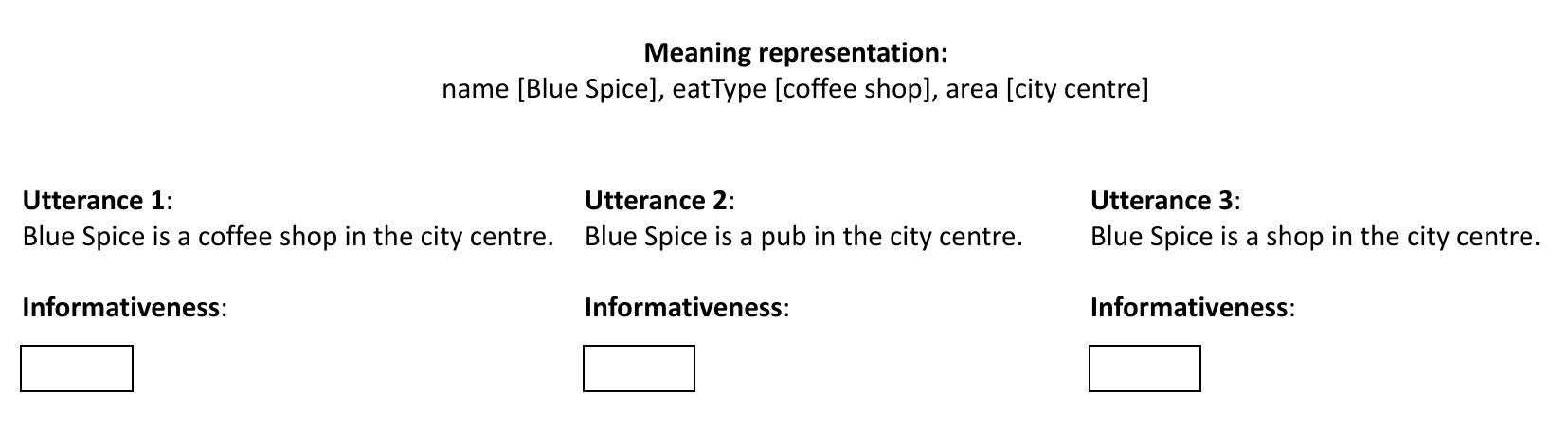}
    }
    \caption{Examples of human-centric evaluation metrics include Likert scale and RankME.}
\end{figure}

Human beings create Natural Language as a crucial form of human communication. So humans are the best evaluators of the natural language texts generated by NLG systems. We call the evaluation metrics that involve human assessors only as human-centric evaluation metrics, and they can be roughly divided into two types:

\paragraph{Direct evaluation} In this type, human assessors judge the quality of the generated texts directly. A simple way is to make a binary decision, i.e.,  good or bad, and a more complex way is to use finer-grained decisions: e.g., Likert scale as shown in Figure \ref{fig4a}, and RankME in Figure \ref{fig4b}, etc. ~\cite{novikova2018rankme, celikyilmaz2018deep, Nucleus}.

\paragraph{Indirect evaluation} Different from the direct evaluation, indirect evaluation is done by measuring the effect of the generated text on downstream tasks, from either a user's perspective (such as whether or not it leads to an improved decision-making or text comprehension accuracy ~\cite{gkatzia2015snapshot}, typically through a User Task Success evaluation), or from a system's perspective ~\cite{aziz2012pet,denkowski2014learning}, such as the performance of a dialogue system  through the System Purpose Success evaluation.

\subsubsection{Automatic Evaluation Metrics}
Automatic evaluation metrics for NLG usually compare the similarity of the NLG model \textbf{generated texts} $G$ to the corresponding \textbf{reference texts} $R$ (i.e., human-written texts) in the benchmarking datasets. We divide the similarity measures into three categories: lexical-based, syntactic-based, and semantic-based.  

% For the convenience of explanation, we use a pair of sentences \textbf{``the man was found on the chair''} (a generated text) and \textbf{``the man was on the chair''} (the corresponding reference text), as an example to illustrate the evaluation metrics in the rest of this section.

\paragraph{Lexical-based Metrics}
The lexical-based metrics measure the similarities between basic lexical units (e.g., words or phrases) across the pair of sentences, which are then aggregated into an overall sentence-level similarity. 

\textbf{BLEU}~\cite{papineni2002bleu} is a commonly used metric in natural language processing tasks to evaluate the similarity between the generated text from an NLG model and the corresponding reference text. Specifically, BLEU counts the n-gram matches in the generated text against the reference text. For convenience, we use BLEU-n to represent BLEU with respect to n-grams, as follows.
% Its value falls into the range between 0.0 and 1.0. If two sentences match perfectly, the value of BLEU is $1.0$. If there is no overlap between them at all, the BLEU value is $0.0$. 

\begin{equation}
    \text{BLEU-n}=\frac{\sum_{t \in G}\sum_{n-gram\in t}Count_{match}(n-gram)}{\sum_{t \in G}\sum_{n-gram \in t}Count(n-gram)},
\label{equ15}
\end{equation}
where $t$ is a piece of generated text to be compared, and \textit{match} means that an n-gram in $t$ also appears in the reference text. The larger BLEU-n, the better quality of the generated text.

% Take BLEU-2 as example. The set of 2-grams in the above example generated text, i.e., $G$, is \{the man, man was, was found, found on, on the, the chair\}, and the set of 2-grams in the reference text, i.e., $R$, is \{the man, man was, was on, on the, the chair\}. According to Equation \ref{equ15}, BLEU-2 in this example is $4/6 = 2/3$.  

\textbf{Self-BLEU} \cite{zhu2018texygen} is proposed as a metric to measure the diversity of the generated text. Since BLEU can measure the similarity of two different texts, self-BLEU calculates the BLEU score between each pair of generated texts, and takes the averaged BLEU score to represent the diversity of all generated texts. The generated texts with a higher diversity has a lower self-BLEU score.
% which avoid using word in the same way.

\textbf{ROUGE}~\cite{lin2004rouge} is the abbreviation of \textit{Recall-Oriented Understanding for Gisting Evaluation}. It compares the generated text with a group of reference texts, counts the number of overlapping basic units (n-grams), and obtains the corresponding score to measure the similarity between the automatically generated texts and the reference texts. The formula of $ROUGE$ is as follows:
\begin{equation}
    \text{ROUGE-n}=\frac{\sum_{t \in G}\sum_{n-gram\in t}Count_{match}(n-gram)}{\sum_{t \in R}\sum_{n-gram \in t}Count(n-gram)},
\label{equ16}
\end{equation}
Comparing the Equations \ref{equ15} and \ref{equ16}, the only difference between BLEU-n and ROUGE-n lies in the denominator. BLEU-n is focused on precision and the denominator is the total number of n-grams in the generated texts. ROUGE-n is recall-oriented, so the denominator is the total number of n-grams in the reference texts. A larger ROUGE-n indicates a better recall-oriented quality.
% For the example above, Equation \ref{equ16} gives its ROUGE-2 score of 4/5.

\textbf{Perplexity (PPL)} is a metric to measure how well a probabilistic language model predicts a sample. Essentially, a probabilistic language model is a probability distribution over a given text, i.e., probability of the $(n+1)-th$ word given the first $n$ words of the text. For the task of NLG, a language model is trained on the reference texts, and is used to predict the generated text.  The higher prediction probability indicates the better quality of the generated text. In practice, PPL takes the reciprocal form as follows:
\begin{equation}
PPL=\sqrt[n]{\prod_{i=1}^{n}\frac{1}{p(w_{i}\mid w_{1}w_{2}...w_{i-1})}},
\end{equation}
where $n$ is the number of words in the generated text, $w_{i}$ is the $i-th$ word in it, and $p(w_{i})$ is the probability of $w_{i}$ in the language model trained with reference texts. Generally speaking, The smaller the PPL value, the better fluency of the generated text. However, the lower PPL also implies more repetitions. Hence, another criterion is that the closer the PPL value of the generated text is to the human-written text, the better the fluency~\cite{Nucleus}.

In addition, a variant of PPL called \textbf{Reverse PPL}~\cite{shi2020dispersed}has also been used to measure the diversity of the generated text, by training a language probability model on the generated texts and calculating the PPL score on the reference texts. A smaller Reverse PPL means that the generated texts are quite different from the reference texts, indicating a higher diversity.

\textbf{Distinct-n}~\cite{li2016diversity} is an n-gram based metric and applies to some scenes (such as dialogue and advertisement generation), where the diversity of generated texts is pursued. The greater the Distinct-n value, the higher the diversity. It is formulated as follows:
\begin{equation}
    \text{Distinct-n}=\frac{Count(unique\ n-gram)}{Count(n-gram)},
\end{equation}
where the numerator is the number of unique n-grams that appear in the generated texts and the denominator is the total number of n-grams in the generated texts. 
% In the above example, since there is no duplicated n-gram in the generated text, i.e., the count of unique n-grams equals the total number of n-grams, the value of Distinct-2 is 1.

\paragraph{Syntactic-based Metrics}
The syntax is related to the grammatical arrangement of words in a sentence. A common syntactic-based metric is \textbf{TESLA}~\cite{liu2010tesla}, which is the abbreviation for \textit{Translation Evaluation of Sentences with Linear-programming-based Analysis}. In general, there are three variants of TESLA~\cite{dahlmeier2011tesla}: TESLA-M (minimal), TESLA-B (basic), and TESLA-F (full). TESLA-M uses N-gram matching and basic linguistic analysis like lemmatization, part-of-speech tagging, and WordNet synonym relations. TESLA-B, a new configuration, adds bilingual phrase tables to model phrase synonyms. The most advanced version, TESLA-F (also called TESLA), goes further by incorporating language models and a ranking support vector machine instead of simple averaging.

\paragraph{Semantic-based Metrics}

Semantic-based metrics aim to handle the evaluation of texts that are lexically different but have a similar semantic meaning. Compared with the lexical-based and syntactic similarity, the semantic similarity requires consideration of more information and is more difficult to measure. In recent years, PLM-based semantic evaluation methods have emerged, which aim to evaluate the generated text using the trained PLMs inversely. An typical example is BERTscores~\cite{BERTScore}, which replaces the n-gram overlaps defined in BLEU, with the embeddings from BERT, to learn a semantic-awareness metric. Similarly,  Bertr and YiSi~\cite{mathur-etal-2019-putting} take advantage of BERT embeddings to capture the semantic similarity between the generated text and its reference. There are also approaches that try to fine-tune PLMs for text quality estimation~\cite{reimers2019sentence,zhou2020learning,liu2021naturalness}. Particularly, \citet{sellam2020bleurt} propose a task-specific pre-trained model, namely BLEURT, for text assessment. BLEURT first adds random perturbations to Wikipedia sentences to construct millions of synthetic examples. Then, a BERT-based pre-trained model is trained on several lexical- and semantic-level supervision signals with a multitask loss. The experiments show that BLEURT benefits from pre-training and is robust to both domain and quality drifts. MAUVE~\cite{MAUVE_human_machine} is an automatic metric to compare human-written and machine-generated texts. Specifically, MAUVE compares the machine-generated text distribution to that of human-written text, which are embedded by a PLM, using divergence frontiers, to reveal two different errors of the NLG model: (1) the model assigns high probability to sequences that do not resemble human-written text, and (2) the generative model fails to yield diverse samples as human-written texts.

\subsubsection{Semi-automatic Evaluation Metrics}
When faced with more diverse tasks such as story generation and open-domain dialogue generation, the automatic evaluation methods turn out less ideal, because they can mainly evaluate the surface-level similarity between the reference and target sentences. Recognizing that humans can better distinguish a more diverse range of features, such as fluency and grammar, semi-automatic evaluation that combines automatic and human-centric evaluation methods has been developed, to get more reliable evaluation results.  

\citet{lowe2017towards} encode the context, the generated text, and a reference text into vectors (\textbf{c}, $\textbf{g}$, and \textbf{r}, respectively) using a hierarchical RNN encoder. Then the dot-product operation is adopted to transfer the vectors into a score ($score(\textbf{c},\textbf{r},\textbf{g})=(\textbf{c}^{T}M\textbf{g}+\textbf{r}^{T}N\textbf{g}-\alpha)/\beta$). Finally this score is made as close as possible to the human judged score. A model that uses human judgments as labels can ensure good consistency with human judgments, but sometimes it may be too conservative and lacks diversity due to the quality of human evaluators.

Without using a human-judged score as the label, we can simply calculate the perplexity of a probabilistic model first, and then let humans evaluate the beam-searched outputs. Specifically, \citet{hashimoto2019unifying} encode the human judgments and model outputs into the same space by using the same encoder. Then a discriminator is trained to distinguish whether the text is generated by a model or by a human evaluator. Finally, the leave-one-out error of the discriminator is computed. By doing so, we can preserve the diversity and quality of the generated text at the same time.

\subsection{CTG-specific Evaluation}
In addition to the above three categories of general NLG evaluation metrics, the CTG task demands additional evaluation metrics that take into account the controlled elements, i.e., whether a CTG model has fulfilled the specific controlled conditions. According to the definition of the controlled elements in Section~\ref{Controllable_Text_Generation}, we can divide CTG-specific evaluation methods into two kinds of automatic evaluation methods. Additionally, human evaluation metrics for CTG are also discussed.

\paragraph{Semantic consistency metrics.} This kind of evaluation methods generally correspond to the semantic control conditions. We first need to construct a training set with positive (samples satisfying the control conditions) and negative (samples not satisfying the control conditions) samples. Then a classifier is trained to identify whether the model generates the controlled text that is semantically consistent with the controlled attributes e.g., sentiment, topic~\cite{pplm, qian-etal-2022-controllable}, style~\cite{gao2019structuring,chen2019unsupervised} and toxic~~\cite{GeDi,liu-etal-2021-dexperts, DisCup-2022}. \textbf{Accuracy} is always used to measure the semantic consistency performance of the CTG model. It is calculated as the number of test samples correctly according with the controlled elements, divided by the number of test samples. Recently, CTRLEVAL~\cite{ke-etal-2022-ctrleval}, a \textbf{PLM-based} and reference-free method, is proposed for a training-free  evaluation of CTG models. Specifically, CTRLEVAL formulates three aspects, including coherence, consistency and attribute relevance, to evaluate the generated text. Each aspect is defined as multiple text-infilling tasks with the base PLM, from which the ensemble of generation probabilities forms the final evaluation results. This approach does not need any labeled data and achieves higher correlations with human judgments. 

\paragraph{Rule-based metrics.} When it comes to the structurally and lexically  controlled text generation tasks, we can use certain rules to judge how well the generated text conforms to the pre-defined controlled elements and then count the number of test samples that satisfy the control conditions as the final evaluation results. For example, \textbf{Coverage} can be used to measure the effectiveness on the lexical-constrained text generation tasks~\cite{metion_flag,point_insertion,CBART,non_residual_prompt}, e.g., by calculating the average percentage of input keywords that are present in the generated text. \textbf{Success Rate} is used to measure the degree of matching  between the generated text and the given structural control elements (i.e., Parts-of-Speech, Syntax Spans, and Syntax Tree). For this purpose, an external tool for sentence structure extraction, such as oracle POS tagger and off-the-shelf syntax parser, is applied to the generated texts~\cite{Li-2022-DiffusionLM}. Moreover, some general text evaluation metrics can also be used for CTG-specific evaluation. For instance, the lexical-level overlap between the reference text containing the control elements and the generated text can be used to reflect how well the control conditions are met in the task of table/graph to text~\cite{puduppully2019data,acl2020_graph2text,plm_amr,acl2020_data2text,structAdapter,findins_acl2021_PlanthenGenerate} and story generation~\cite{goldfarb2020content,fang2021outline}. Thereby the general metrics such as BLEU-n and ROUGE-n could also be regarded as the CTG-specific evaluation metrics to indicate the constraint satisfaction.

\paragraph{Human evaluation metric.} Most of the afore-mentioned CTG-specific metrics for the evaluation of text semantics and structures  require the introduction of an additional semantic-relevant classifier or structure-relevant parsing tools, which may not be as reliable as the well-educated human participants. Therefore, human evaluation is also essential to test how a CTG model satisfies the control elements. Similar to the Human-Centric method described in Section 4.1.1, human assessors are always directly asked to score the relevance of the generated texts to the control attributes (e.g., sentiment, topic, toxic, contextual relevancy, and common sense~\cite{CoCon, pplm,liu-etal-2021-dexperts, FUDGE, DisCup-2022, non_residual_prompt} etc.), and the average score obtained is used as the final evaluation result.

\subsection{Summary}
% The different evaluation metrics discussed above have their own advantages and disadvantages. They need to be flexibly combined according to the specific tasks. Sometimes, multiple evaluation methods can be used together to determine the performance of an NLG model.
Human-centric evaluation is the most precise method for evaluating the quality of system-generated texts. It would be ideal when human resources permit, which unfortunately is not always the case. Thus this type of evaluation suffers from various weaknesses: 1) \textbf{Expensive and time-consuming}: the recruitment and selection of evaluators, the setup of the evaluation and other steps all require time and manpower, especially for the evaluation tasks that can only be handled by domain experts; 2) \textbf{Maintaining quality control} \cite{ipeirotis2010quality,mitra2015comparing}: although the emergence of online crowd-sourcing platforms has eased the cost problem to a certain extent, the quality of personnel conducting online evaluations cannot always be  guaranteed;  3) \textbf{Lack of consistency} \cite{van2020human}: the reproducibility of the evaluation results can be low due to the change of evaluation personnel, which leads to the inconsistency problem.

Compared with human-centric evaluation metrics, automatic evaluation metrics are easy to use, fast to obtain results, and with low-cost. However, the evaluation of this type is less precise than human assessments. It is vital to develop automated evaluation methods comparable to the human level evaluation in the future. Semi-automatic evaluation metric combines the advantages of human-centric methods and automatic methods, but still require a lot of human judgments or labeling, which are  expensive and time-consuming. One key future research direction is to develop a better way to augment human judgments with automatic evaluation, or vice versa, to obtain improved evaluation quality and diversity. 

When it comes to the CTG tasks, we not only need to use the  general NLG evaluation metrics  but also need specific metrics to evaluate whether the generated texts are consistent with the controlled elements. Generally speaking, the consistency of controlled elements and generated texts are relatively easy to measure in most cases. The main challenges of text quality evaluation still lie in the general evaluation methods.

\section{Challenges and Future Directions}
\label{Challenges_and_Future_Direction}

\subsection{Challenges}

% The birth of large-scale pre-trained language models brings the research of text generation to a new stage. 
The PLMs have mastered a remarkable level of linguistic knowledge (semantic, syntax, etc.) from large-scale corpus, naturally enabling the production of more fluent and diverse text. However, due to the black box characteristics of neural networks, the general PLMs are still not sufficiently controllable during the text generation process. How to fully exploit the powerful PLMs to generate the desired and controllable text, has become a promising yet challenging field in both academia and industry. Based on the systematic review of the key concepts, methods and findings in the latest development of PLM-based controllable text generation, we think this promising and fast-growing area is still facing a number of challenges in the following aspects.

First, PLMs have learned rich knowledge from large-scale corpus used for pre-training. However, an NLG model needs to learn control constraints on its own training corpus. It is often difficult for the existing PLM-based models to ensure the domain diversity of the generated text while pursuing controllability. This is indeed the well-known catastrophic forgetting problem in PLM. In the field of text generation, it is still a challenge to overcome this problem and imporve the ability of the PLM-based NLG model to generate multi-domain text that satisfies specific control conditions, with few or zero domain-specific samples.
 
Second, controlling the generation of text in the decoding stage of a generative model is a low-cost way in terms of model training. It can maintain the characteristics of the original language model to the greatest extent. However, in most cases, the existing methods are relatively rudimentary, and only use the external decoupled attribute discriminator to control the attributes. There is a distribution gap between the discriminator and the generator, leading to a coarser granularity in the guidance process and decreased quality of the generated text. In addition, the decoding-time approaches are hard to be directly applied to fine-grained control scenarios such as data-to-text or multi-attribute control tasks.

Third, from the perspective of probability theory, a generative pre-trained language model (referring specifically to the GPT-like models) is essentially an enhanced version of dense conditional probability  $p\left(x_{n} \mid x_{1}, x_{2}, \ldots, x_{n-1}\right)$ to describe the probability distribution of natural language. However, this local normalization format has certain limitations in paragraph/document-level modeling. For example, it is hard to keep long-range coherence in terms of both semantic logic and controlled condition. It calls for further research to establish a global normalization based on PLMs to ensure that text generation can be controlled locally and globally at the same time.

Fourth, the construction of large-scale PLMs is typically data-driven, which allows the models to learn the primary logic and commonsense knowledge contained in the training corpus. However, the knowledge captured in those models is often rather superficial. The PLMs will lose generalization ability when the training data does not contain relevant commonsense and domain-specific knowledge. Therefore, purely relying on PLMs could be difficult to control the generated texts faithfully with respect to commonsense and rich knowledge specific to the target domain.

Fifth, reasonable and reliable evaluation has always been a bottleneck restricting the development of more advanced text generation technologies. This is also the case for controllable text generation. Generally speaking, the satisfaction of controlled conditions is relatively easy to evaluate. However, there is still a lack of an objective, accurate and comprehensive evaluation mechanism that is fully compatible with human judgment. For controllable text generation, in addition to the control conditions, the quality of the text itself is equally important. If the quality of the generated text of an NLG model cannot be accurately evaluated, it is hard to think of a way to control them. 
  
Finally, we believe that the research on controlled text generation is still in its early stage. In Section 2.2 of this paper, we have summarized a range of tasks involving CTG. However, few of them are actually dedicated CTG tasks. With the rapid development of text generation, there is a need to come up with dedicated benchmarking tasks and datasets for CTG with diverse control requirements.

\subsection{Future Directions}

Based on the summary of current work and the challenges mentioned above, we suggest the following promising future directions for PLM-based controllable text generation.

\textbf{Prompt-based Learning}: 
 Prompt-based learning has become a new way for fine-tuning PLMs. Based on the well-designed prompting function, a PLM is able to perform few-shot or even zero-shot learning, adapting to new scenarios with few or no labeled data, thus overcoming the problem of catastrophic forgetting. The above features are also attractive for controlled text generation, since the prompt-based methods are able to generate more diverse text fields and increase the distribution space of the text to be filtered, so that it is theoretically more possible to produce text that meets the specific control conditions. Currently, the application of prompt-based methods and their variants (e.g., in-context learning~\cite{xie2022an}, instruction tuning~\cite{instructgpt}, etc.) in large-scale language models is a hot academic topic. There is also great potential in finding better ways to apply this paradigm to controllable text generation.

\textbf{Fine-grained Decoding Control}: More fine-grained decoding control methods need to be explored. On the one hand, the decoding-time methods can be improved to achieve more effective control. For example, co-training between the guided model and the generative model ensures finer-grained text generation. On the other hand, the existing single-attribute (e.g., emotion, topic, etc.) controlled tasks can also be extended to multi-attribute controlled tasks in a unified framework, achieving the simultaneous control of multiple aspects for a generated sentence.

\textbf{Integration with Classic Generative Theory and Linguistic Knowledge}: 
 The PLM-based controllable text generation task can be regarded as obtaining a natural language distribution constructed by a PLM, the corresponding distribution that satisfies certain specified constraints. This process is intrinsically related to classic generative models such as Generative Adversarial Networks (GANs)~\cite{GANs}, Variational Autoencoders (VAEs)~\cite{VAEs}, Energy-based Model~\cite{ebm_ctg} and their variants. It is known that auto-regressive PLMs, i.e., the GPT family models, can not model global information of the generated texts naturally, making it difficult to control the generated text's distribution at the paragraph/document level. We expect that combining classic probability theory to bridge the gap between PLMs and traditional generative models, will help solve the problem at the theoretical level.  
 
In addition, auto-regressive PLM is essentially a locally normalized fashion, allowing it to produce fluent text in short text-generation scenarios. From a linguistic point of view, it is believed that more linguistic knowledge, such as paragraph/document level structures and logic, are needed in long text generation, which PLMs can not provide directly. A promising solution is to combine linguistic knowledge with PLMs to overcome the inherent problems of auto-regressive models in long text modeling, so as to better ensure the quality and controllability.

\textbf{Incorporation of External Knowledge}: Introducing additional knowledge to enhance PLM's generation is a promising direction. One direct idea is to combine with information retrieval and allow the PLM to refer to the retrieved information from the Web or domain data repositories, ensuring that the generated content is more reasonable and alleviating problem of hallucinations~\cite{Hallucination}. Furthermore, a knowledge graph is a natural carrier of explicit domain-specific knowledge and also provides effective reasoning mechanisms. Therefore it can be a good complement to the use of PLMs, which lack domain-specific knowledge and logical reasoning capabilities.

\textbf{Novel Evaluation Metrics and Methods}:  
Developing innovative evaluation metrics for CTG is still an important topic that needs to be further studied from both the general text generation perspective (such as fluency, diversity, and coherence) and the CTG-specific perspective (such as fidelity). Since the pre-trained language models have mastered a great deal of semantic and grammatical knowledge, applying them in reverse to assess the text quality of an NLG model would be an interesting and fascinating area for further investigation.

\textbf{New CTG tasks}:  Controllable text generation is a broad concept that has gained significant attention in recent years. With the rapid development of large-scale language models, such as ChatGPT and GPT-4, there is a growing interest in exploring new standards and tasks that align with the goal of achieving General Artificial Intelligence (AGI). Therefore, one promising direction is to define AGI-oriented benchmarks and tasks that aim to control the language model to produce accurate and reliable information or ensure that the generated content is aligned with human values and does not have the harmful effects.

\section{Conclusions}
\label{conclusion}
In this paper, we have comprehensively summarized the typical applications, main approaches, and evaluation methodologies of controllable text generation based on large-scale pre-trained language models. Based on the critical analysis of the existing methods, we have identified a series of key challenges in this field and highlighted several promising future directions. Large-scale pre-trained language models have brought unprecedented opportunities for the development of controllable text generation technologies, calling for more researchers to join the field and create a new era of it. We are hopeful that this literature survey is able to provide a clear picture of the field and set a roadmap for researchers and practitioners to move forward.

\section{Acknowledgments}
This work is supported in part by Natural Science Foundation of Beijing (grant No. 4222036). We
would like to thank the anonymous reviewers for their valuable comments.

\bibliographystyle{ACM-Reference-Format}
\bibliography{sample-base}


\begin{thebibliography}{172}

%%% the Web address.

\ifx \showCODEN    \undefined \def \showCODEN     #1{\unskip}     \fi
\ifx \showDOI      \undefined \def \showDOI       #1{#1}\fi
\ifx \showISBNx    \undefined \def \showISBNx     #1{\unskip}     \fi
\ifx \showISBNxiii \undefined \def \showISBNxiii  #1{\unskip}     \fi
\ifx \showISSN     \undefined \def \showISSN      #1{\unskip}     \fi
\ifx \showLCCN     \undefined \def \showLCCN      #1{\unskip}     \fi
\ifx \shownote     \undefined \def \shownote      #1{#1}          \fi
\ifx \showarticletitle \undefined \def \showarticletitle #1{#1}   \fi
\ifx \showURL      \undefined \def \showURL       {\relax}        \fi
% invisible to TeX
\providecommand\bibfield[2]{#2}
\providecommand\bibinfo[2]{#2}
\providecommand\natexlab[1]{#1}
\providecommand\showeprint[2][]{arXiv:#2}

\bibitem[\protect\citeauthoryear{Amin-Nejad, Ive, and Velupillai}{Amin-Nejad
  et~al\mbox{.}}{2020}]%
        {amin2020exploring}
\bibfield{author}{\bibinfo{person}{Ali Amin-Nejad}, \bibinfo{person}{Julia
  Ive}, {and} \bibinfo{person}{Sumithra Velupillai}.}
  \bibinfo{year}{2020}\natexlab{}.
\newblock \showarticletitle{Exploring Transformer Text Generation for Medical
  Dataset Augmentation}. In \bibinfo{booktitle}{\emph{Proceedings of the
  Twelfth Language Resources and Evaluation Conference}}.
  \bibinfo{publisher}{European Language Resources Association},
  \bibinfo{address}{Marseille, France}, \bibinfo{pages}{4699--4708}.
\newblock
\showISBNx{979-10-95546-34-4}
\urldef\tempurl%
\url{https://aclanthology.org/2020.lrec-1.578}
\showURL{%
\tempurl}


\bibitem[\protect\citeauthoryear{Anderson, Fernando, Johnson, and
  Gould}{Anderson et~al\mbox{.}}{2017}]%
        {beam_search}
\bibfield{author}{\bibinfo{person}{Peter Anderson}, \bibinfo{person}{Basura
  Fernando}, \bibinfo{person}{Mark Johnson}, {and} \bibinfo{person}{Stephen
  Gould}.} \bibinfo{year}{2017}\natexlab{}.
\newblock \showarticletitle{Guided Open Vocabulary Image Captioning with
  Constrained Beam Search}. In \bibinfo{booktitle}{\emph{Proceedings of the
  2017 Conference on Empirical Methods in Natural Language Processing, {EMNLP}
  2017, Copenhagen, Denmark, September 9-11, 2017}},
  \bibfield{editor}{\bibinfo{person}{Martha Palmer}, \bibinfo{person}{Rebecca
  Hwa}, {and} \bibinfo{person}{Sebastian Riedel}} (Eds.).
  \bibinfo{publisher}{Association for Computational Linguistics},
  \bibinfo{pages}{936--945}.
\newblock
\urldef\tempurl%
\url{https://doi.org/10.18653/v1/d17-1098}
\showDOI{\tempurl}


\bibitem[\protect\citeauthoryear{Aziz, Castilho, and Specia}{Aziz
  et~al\mbox{.}}{2012}]%
        {aziz2012pet}
\bibfield{author}{\bibinfo{person}{Wilker Aziz}, \bibinfo{person}{Sheila
  Castilho}, {and} \bibinfo{person}{Lucia Specia}.}
  \bibinfo{year}{2012}\natexlab{}.
\newblock \showarticletitle{{PET}: a Tool for Post-editing and Assessing
  Machine Translation}. In \bibinfo{booktitle}{\emph{Proceedings of the Eighth
  International Conference on Language Resources and Evaluation ({LREC}'12)}}.
  \bibinfo{publisher}{European Language Resources Association (ELRA)},
  \bibinfo{address}{Istanbul, Turkey}, \bibinfo{pages}{3982--3987}.
\newblock
\urldef\tempurl%
\url{http://www.lrec-conf.org/proceedings/lrec2012/pdf/985_Paper.pdf}
\showURL{%
\tempurl}


\bibitem[\protect\citeauthoryear{Barikeri, Lauscher, Vuli{\'c}, and
  Glava{\v{s}}}{Barikeri et~al\mbox{.}}{2021}]%
        {barikeri2021redditbias}
\bibfield{author}{\bibinfo{person}{Soumya Barikeri}, \bibinfo{person}{Anne
  Lauscher}, \bibinfo{person}{Ivan Vuli{\'c}}, {and} \bibinfo{person}{Goran
  Glava{\v{s}}}.} \bibinfo{year}{2021}\natexlab{}.
\newblock \showarticletitle{{R}eddit{B}ias: A Real-World Resource for Bias
  Evaluation and Debiasing of Conversational Language Models}. In
  \bibinfo{booktitle}{\emph{Proceedings of the 59th Annual Meeting of the
  Association for Computational Linguistics and the 11th International Joint
  Conference on Natural Language Processing (Volume 1: Long Papers)}}.
  \bibinfo{publisher}{Association for Computational Linguistics},
  \bibinfo{address}{Online}, \bibinfo{pages}{1941--1955}.
\newblock
\urldef\tempurl%
\url{https://doi.org/10.18653/v1/2021.acl-long.151}
\showDOI{\tempurl}


\bibitem[\protect\citeauthoryear{Beltagy, Peters, and Cohan}{Beltagy
  et~al\mbox{.}}{2020}]%
        {longformer}
\bibfield{author}{\bibinfo{person}{Iz Beltagy}, \bibinfo{person}{Matthew~E.
  Peters}, {and} \bibinfo{person}{Arman Cohan}.}
  \bibinfo{year}{2020}\natexlab{}.
\newblock \showarticletitle{Longformer: The Long-Document Transformer}.
\newblock \bibinfo{journal}{\emph{CoRR}}  \bibinfo{volume}{abs/2004.05150}
  (\bibinfo{year}{2020}).
\newblock
\showeprint[arxiv]{2004.05150}
\urldef\tempurl%
\url{https://arxiv.org/abs/2004.05150}
\showURL{%
\tempurl}


\bibitem[\protect\citeauthoryear{Bender, Gebru, McMillan-Major, and
  Shmitchell}{Bender et~al\mbox{.}}{2021}]%
        {ai_dangers}
\bibfield{author}{\bibinfo{person}{Emily~M. Bender}, \bibinfo{person}{Timnit
  Gebru}, \bibinfo{person}{Angelina McMillan-Major}, {and}
  \bibinfo{person}{Shmargaret Shmitchell}.} \bibinfo{year}{2021}\natexlab{}.
\newblock \showarticletitle{On the Dangers of Stochastic Parrots: Can Language
  Models Be Too Big?}. In \bibinfo{booktitle}{\emph{Proceedings of the 2021 ACM
  Conference on Fairness, Accountability, and Transparency}} (Virtual Event,
  Canada) \emph{(\bibinfo{series}{FAccT '21})}. \bibinfo{publisher}{Association
  for Computing Machinery}, \bibinfo{address}{New York, NY, USA},
  \bibinfo{pages}{610–623}.
\newblock
\showISBNx{9781450383097}
\urldef\tempurl%
\url{https://doi.org/10.1145/3442188.3445922}
\showDOI{\tempurl}


\bibitem[\protect\citeauthoryear{Bengio, Ducharme, Vincent, and Janvin}{Bengio
  et~al\mbox{.}}{2003}]%
        {nnlm}
\bibfield{author}{\bibinfo{person}{Yoshua Bengio}, \bibinfo{person}{R\'{e}jean
  Ducharme}, \bibinfo{person}{Pascal Vincent}, {and} \bibinfo{person}{Christian
  Janvin}.} \bibinfo{year}{2003}\natexlab{}.
\newblock \showarticletitle{A Neural Probabilistic Language Model}.
\newblock \bibinfo{journal}{\emph{J. Mach. Learn. Res.}} \bibinfo{volume}{3},
  \bibinfo{number}{null} (\bibinfo{date}{mar} \bibinfo{year}{2003}),
  \bibinfo{pages}{1137–1155}.
\newblock
\showISSN{1532-4435}
\urldef\tempurl%
\url{https://dl.acm.org/doi/pdf/10.5555/944919.944966}
\showURL{%
\tempurl}


\bibitem[\protect\citeauthoryear{Bhattacharyya, Rooshenas, Naskar, Sun, Iyyer,
  and McCallum}{Bhattacharyya et~al\mbox{.}}{2021}]%
        {bhattacharyya-etal-2021-energy}
\bibfield{author}{\bibinfo{person}{Sumanta Bhattacharyya},
  \bibinfo{person}{Amirmohammad Rooshenas}, \bibinfo{person}{Subhajit Naskar},
  \bibinfo{person}{Simeng Sun}, \bibinfo{person}{Mohit Iyyer}, {and}
  \bibinfo{person}{Andrew McCallum}.} \bibinfo{year}{2021}\natexlab{}.
\newblock \showarticletitle{Energy-Based Reranking: Improving Neural Machine
  Translation Using Energy-Based Models}. In
  \bibinfo{booktitle}{\emph{Proceedings of the 59th Annual Meeting of the
  Association for Computational Linguistics and the 11th International Joint
  Conference on Natural Language Processing (Volume 1: Long Papers)}}.
  \bibinfo{publisher}{Association for Computational Linguistics},
  \bibinfo{address}{Online}, \bibinfo{pages}{4528--4537}.
\newblock
\urldef\tempurl%
\url{https://doi.org/10.18653/v1/2021.acl-long.349}
\showDOI{\tempurl}


\bibitem[\protect\citeauthoryear{Brown, Mann, Ryder, Subbiah, Kaplan, Dhariwal,
  Neelakantan, Shyam, Sastry, Askell, Agarwal, Herbert-Voss, Krueger, Henighan,
  Child, Ramesh, Ziegler, Wu, Winter, Hesse, Chen, Sigler, Litwin, Gray, Chess,
  Clark, Berner, McCandlish, Radford, Sutskever, and Amodei}{Brown
  et~al\mbox{.}}{2020}]%
        {gpt3}
\bibfield{author}{\bibinfo{person}{Tom Brown}, \bibinfo{person}{Benjamin Mann},
  \bibinfo{person}{Nick Ryder}, \bibinfo{person}{Melanie Subbiah},
  \bibinfo{person}{Jared~D Kaplan}, \bibinfo{person}{Prafulla Dhariwal},
  \bibinfo{person}{Arvind Neelakantan}, \bibinfo{person}{Pranav Shyam},
  \bibinfo{person}{Girish Sastry}, \bibinfo{person}{Amanda Askell},
  \bibinfo{person}{Sandhini Agarwal}, \bibinfo{person}{Ariel Herbert-Voss},
  \bibinfo{person}{Gretchen Krueger}, \bibinfo{person}{Tom Henighan},
  \bibinfo{person}{Rewon Child}, \bibinfo{person}{Aditya Ramesh},
  \bibinfo{person}{Daniel Ziegler}, \bibinfo{person}{Jeffrey Wu},
  \bibinfo{person}{Clemens Winter}, \bibinfo{person}{Chris Hesse},
  \bibinfo{person}{Mark Chen}, \bibinfo{person}{Eric Sigler},
  \bibinfo{person}{Mateusz Litwin}, \bibinfo{person}{Scott Gray},
  \bibinfo{person}{Benjamin Chess}, \bibinfo{person}{Jack Clark},
  \bibinfo{person}{Christopher Berner}, \bibinfo{person}{Sam McCandlish},
  \bibinfo{person}{Alec Radford}, \bibinfo{person}{Ilya Sutskever}, {and}
  \bibinfo{person}{Dario Amodei}.} \bibinfo{year}{2020}\natexlab{}.
\newblock \showarticletitle{Language Models are Few-Shot Learners}. In
  \bibinfo{booktitle}{\emph{Advances in Neural Information Processing
  Systems}}, \bibfield{editor}{\bibinfo{person}{H.~Larochelle},
  \bibinfo{person}{M.~Ranzato}, \bibinfo{person}{R.~Hadsell},
  \bibinfo{person}{M.F. Balcan}, {and} \bibinfo{person}{H.~Lin}} (Eds.),
  Vol.~\bibinfo{volume}{33}. \bibinfo{publisher}{Curran Associates, Inc.},
  \bibinfo{pages}{1877--1901}.
\newblock
\urldef\tempurl%
\url{https://proceedings.neurips.cc/paper/2020/file/1457c0d6bfcb4967418bfb8ac142f64a-Paper.pdf}
\showURL{%
\tempurl}


\bibitem[\protect\citeauthoryear{Carlsson, {\"O}hman, Liu, Verlinden, Nivre,
  and Sahlgren}{Carlsson et~al\mbox{.}}{2022}]%
        {non_residual_prompt}
\bibfield{author}{\bibinfo{person}{Fredrik Carlsson}, \bibinfo{person}{Joey
  {\"O}hman}, \bibinfo{person}{Fangyu Liu}, \bibinfo{person}{Severine
  Verlinden}, \bibinfo{person}{Joakim Nivre}, {and} \bibinfo{person}{Magnus
  Sahlgren}.} \bibinfo{year}{2022}\natexlab{}.
\newblock \showarticletitle{Fine-Grained Controllable Text Generation Using
  Non-Residual Prompting}. In \bibinfo{booktitle}{\emph{Proceedings of the 60th
  Annual Meeting of the Association for Computational Linguistics (Volume 1:
  Long Papers)}}. \bibinfo{publisher}{Association for Computational
  Linguistics}, \bibinfo{address}{Dublin, Ireland},
  \bibinfo{pages}{6837--6857}.
\newblock
\urldef\tempurl%
\url{https://doi.org/10.18653/v1/2022.acl-long.471}
\showDOI{\tempurl}


\bibitem[\protect\citeauthoryear{Castro~Ferreira, van~der Lee, van Miltenburg,
  and Krahmer}{Castro~Ferreira et~al\mbox{.}}{2019}]%
        {ferreira2019neural}
\bibfield{author}{\bibinfo{person}{Thiago Castro~Ferreira},
  \bibinfo{person}{Chris van~der Lee}, \bibinfo{person}{Emiel van Miltenburg},
  {and} \bibinfo{person}{Emiel Krahmer}.} \bibinfo{year}{2019}\natexlab{}.
\newblock \showarticletitle{Neural data-to-text generation: A comparison
  between pipeline and end-to-end architectures}. In
  \bibinfo{booktitle}{\emph{Proceedings of the 2019 Conference on Empirical
  Methods in Natural Language Processing and the 9th International Joint
  Conference on Natural Language Processing (EMNLP-IJCNLP)}}.
  \bibinfo{publisher}{Association for Computational Linguistics},
  \bibinfo{address}{Hong Kong, China}, \bibinfo{pages}{552--562}.
\newblock
\urldef\tempurl%
\url{https://doi.org/10.18653/v1/D19-1052}
\showDOI{\tempurl}


\bibitem[\protect\citeauthoryear{Celikyilmaz, Bosselut, He, and
  Choi}{Celikyilmaz et~al\mbox{.}}{2018}]%
        {celikyilmaz2018deep}
\bibfield{author}{\bibinfo{person}{Asli Celikyilmaz}, \bibinfo{person}{Antoine
  Bosselut}, \bibinfo{person}{Xiaodong He}, {and} \bibinfo{person}{Yejin
  Choi}.} \bibinfo{year}{2018}\natexlab{}.
\newblock \showarticletitle{Deep Communicating Agents for Abstractive
  Summarization}. In \bibinfo{booktitle}{\emph{Proceedings of the 2018
  Conference of the North {A}merican Chapter of the Association for
  Computational Linguistics: Human Language Technologies, Volume 1 (Long
  Papers)}}. \bibinfo{publisher}{Association for Computational Linguistics},
  \bibinfo{address}{New Orleans, Louisiana}, \bibinfo{pages}{1662--1675}.
\newblock
\urldef\tempurl%
\url{https://doi.org/10.18653/v1/N18-1150}
\showDOI{\tempurl}


\bibitem[\protect\citeauthoryear{Celikyilmaz, Clark, and Gao}{Celikyilmaz
  et~al\mbox{.}}{2020}]%
        {celikyilmaz2020evaluation}
\bibfield{author}{\bibinfo{person}{Asli Celikyilmaz},
  \bibinfo{person}{Elizabeth Clark}, {and} \bibinfo{person}{Jianfeng Gao}.}
  \bibinfo{year}{2020}\natexlab{}.
\newblock \showarticletitle{Evaluation of Text Generation: {A} Survey}.
\newblock \bibinfo{journal}{\emph{CoRR}}  \bibinfo{volume}{abs/2006.14799}
  (\bibinfo{year}{2020}).
\newblock
\showeprint[arXiv]{2006.14799}
\urldef\tempurl%
\url{https://arxiv.org/abs/2006.14799}
\showURL{%
\tempurl}


\bibitem[\protect\citeauthoryear{Chan, Ong, Pung, Zhang, and Fu}{Chan
  et~al\mbox{.}}{2021}]%
        {CoCon}
\bibfield{author}{\bibinfo{person}{Alvin Chan}, \bibinfo{person}{Yew-Soon Ong},
  \bibinfo{person}{Bill Pung}, \bibinfo{person}{Aston Zhang}, {and}
  \bibinfo{person}{Jie Fu}.} \bibinfo{year}{2021}\natexlab{}.
\newblock \showarticletitle{CoCon: A Self-Supervised Approach for Controlled
  Text Generation}. In \bibinfo{booktitle}{\emph{International Conference on
  Learning Representations}}.
\newblock
\urldef\tempurl%
\url{https://openreview.net/forum?id=VD_ozqvBy4W}
\showURL{%
\tempurl}


\bibitem[\protect\citeauthoryear{Chandu, Prabhumoye, Salakhutdinov, and
  Black}{Chandu et~al\mbox{.}}{2019}]%
        {prabhumoye2019my}
\bibfield{author}{\bibinfo{person}{Khyathi Chandu}, \bibinfo{person}{Shrimai
  Prabhumoye}, \bibinfo{person}{Ruslan Salakhutdinov}, {and}
  \bibinfo{person}{Alan~W Black}.} \bibinfo{year}{2019}\natexlab{}.
\newblock \showarticletitle{{``}My Way of Telling a Story{''}: Persona based
  Grounded Story Generation}. In \bibinfo{booktitle}{\emph{Proceedings of the
  Second Workshop on Storytelling}}. \bibinfo{publisher}{Association for
  Computational Linguistics}, \bibinfo{address}{Florence, Italy},
  \bibinfo{pages}{11--21}.
\newblock
\urldef\tempurl%
\url{https://doi.org/10.18653/v1/W19-3402}
\showDOI{\tempurl}


\bibitem[\protect\citeauthoryear{Chang, Yuan, Iyyer, and McCallum}{Chang
  et~al\mbox{.}}{2021}]%
        {chang2021changing}
\bibfield{author}{\bibinfo{person}{Haw-Shiuan Chang}, \bibinfo{person}{Jiaming
  Yuan}, \bibinfo{person}{Mohit Iyyer}, {and} \bibinfo{person}{Andrew
  McCallum}.} \bibinfo{year}{2021}\natexlab{}.
\newblock \showarticletitle{Changing the Mind of Transformers for
  Topically-Controllable Language Generation}. In
  \bibinfo{booktitle}{\emph{Proceedings of the 16th Conference of the European
  Chapter of the Association for Computational Linguistics: Main Volume}}.
  \bibinfo{publisher}{Association for Computational Linguistics},
  \bibinfo{address}{Online}, \bibinfo{pages}{2601--2611}.
\newblock
\urldef\tempurl%
\url{https://doi.org/10.18653/v1/2021.eacl-main.223}
\showDOI{\tempurl}


\bibitem[\protect\citeauthoryear{Chen, Pan, Liu, and Sun}{Chen
  et~al\mbox{.}}{2019a}]%
        {chen2019unsupervised}
\bibfield{author}{\bibinfo{person}{Cheng-Kuan Chen}, \bibinfo{person}{Zhufeng
  Pan}, \bibinfo{person}{Ming-Yu Liu}, {and} \bibinfo{person}{Min Sun}.}
  \bibinfo{year}{2019}\natexlab{a}.
\newblock \showarticletitle{Unsupervised stylish image description generation
  via domain layer norm}. In \bibinfo{booktitle}{\emph{Proceedings of the AAAI
  Conference on Artificial Intelligence}}, Vol.~\bibinfo{volume}{33}.
  \bibinfo{pages}{8151--8158}.
\newblock


\bibitem[\protect\citeauthoryear{Chen, Yi, Sun, Li, Yang, and Guo}{Chen
  et~al\mbox{.}}{2019b}]%
        {chen2019sentiment}
\bibfield{author}{\bibinfo{person}{Huimin Chen}, \bibinfo{person}{Xiaoyuan Yi},
  \bibinfo{person}{Maosong Sun}, \bibinfo{person}{Wenhao Li},
  \bibinfo{person}{Cheng Yang}, {and} \bibinfo{person}{Zhipeng Guo}.}
  \bibinfo{year}{2019}\natexlab{b}.
\newblock \showarticletitle{Sentiment-Controllable Chinese Poetry Generation}.
  In \bibinfo{booktitle}{\emph{Proceedings of the Twenty-Eighth International
  Joint Conference on Artificial Intelligence, {IJCAI-19}}}.
  \bibinfo{publisher}{International Joint Conferences on Artificial
  Intelligence Organization}, \bibinfo{pages}{4925--4931}.
\newblock
\urldef\tempurl%
\url{https://doi.org/10.24963/ijcai.2019/684}
\showDOI{\tempurl}


\bibitem[\protect\citeauthoryear{Chowdhery, Narang, Devlin, Bosma, Mishra,
  Roberts, Barham, Chung, Sutton, Gehrmann, et~al\mbox{.}}{Chowdhery
  et~al\mbox{.}}{2022}]%
        {chowdhery2022palm}
\bibfield{author}{\bibinfo{person}{Aakanksha Chowdhery},
  \bibinfo{person}{Sharan Narang}, \bibinfo{person}{Jacob Devlin},
  \bibinfo{person}{Maarten Bosma}, \bibinfo{person}{Gaurav Mishra},
  \bibinfo{person}{Adam Roberts}, \bibinfo{person}{Paul Barham},
  \bibinfo{person}{Hyung~Won Chung}, \bibinfo{person}{Charles Sutton},
  \bibinfo{person}{Sebastian Gehrmann}, {et~al\mbox{.}}}
  \bibinfo{year}{2022}\natexlab{}.
\newblock \showarticletitle{Palm: Scaling language modeling with pathways}.
\newblock \bibinfo{journal}{\emph{arXiv preprint arXiv:2204.02311}}
  (\bibinfo{year}{2022}).
\newblock


\bibitem[\protect\citeauthoryear{Chung, Hou, Longpre, Zoph, Tay, Fedus, Li,
  Wang, Dehghani, Brahma, et~al\mbox{.}}{Chung et~al\mbox{.}}{2022}]%
        {chung2022scaling}
\bibfield{author}{\bibinfo{person}{Hyung~Won Chung}, \bibinfo{person}{Le Hou},
  \bibinfo{person}{Shayne Longpre}, \bibinfo{person}{Barret Zoph},
  \bibinfo{person}{Yi Tay}, \bibinfo{person}{William Fedus},
  \bibinfo{person}{Eric Li}, \bibinfo{person}{Xuezhi Wang},
  \bibinfo{person}{Mostafa Dehghani}, \bibinfo{person}{Siddhartha Brahma},
  {et~al\mbox{.}}} \bibinfo{year}{2022}\natexlab{}.
\newblock \showarticletitle{Scaling instruction-finetuned language models}.
\newblock \bibinfo{journal}{\emph{arXiv preprint arXiv:2210.11416}}
  (\bibinfo{year}{2022}).
\newblock


\bibitem[\protect\citeauthoryear{Conneau and Lample}{Conneau and
  Lample}{2019}]%
        {XLM}
\bibfield{author}{\bibinfo{person}{Alexis Conneau} {and}
  \bibinfo{person}{Guillaume Lample}.} \bibinfo{year}{2019}\natexlab{}.
\newblock \bibinfo{booktitle}{\emph{Cross-Lingual Language Model Pretraining}}.
\newblock \bibinfo{publisher}{Curran Associates Inc.}, \bibinfo{address}{Red
  Hook, NY, USA}.
\newblock
\urldef\tempurl%
\url{https://proceedings.neurips.cc/paper/2019/file/c04c19c2c2474dbf5f7ac4372c5b9af1-Paper.pdf}
\showURL{%
\tempurl}


\bibitem[\protect\citeauthoryear{Dahlmeier, Liu, and Ng}{Dahlmeier
  et~al\mbox{.}}{2011}]%
        {dahlmeier2011tesla}
\bibfield{author}{\bibinfo{person}{Daniel Dahlmeier}, \bibinfo{person}{Chang
  Liu}, {and} \bibinfo{person}{Hwee~Tou Ng}.} \bibinfo{year}{2011}\natexlab{}.
\newblock \showarticletitle{{TESLA} at {WMT} 2011: Translation Evaluation and
  Tunable Metric}. In \bibinfo{booktitle}{\emph{Proceedings of the Sixth
  Workshop on Statistical Machine Translation}}.
  \bibinfo{publisher}{Association for Computational Linguistics},
  \bibinfo{address}{Edinburgh, Scotland}, \bibinfo{pages}{78--84}.
\newblock
\urldef\tempurl%
\url{https://aclanthology.org/W11-2106}
\showURL{%
\tempurl}


\bibitem[\protect\citeauthoryear{Dai, Yang, Yang, Carbonell, Le, and
  Salakhutdinov}{Dai et~al\mbox{.}}{2019}]%
        {transformer_xl}
\bibfield{author}{\bibinfo{person}{Zihang Dai}, \bibinfo{person}{Zhilin Yang},
  \bibinfo{person}{Yiming Yang}, \bibinfo{person}{Jaime Carbonell},
  \bibinfo{person}{Quoc Le}, {and} \bibinfo{person}{Ruslan Salakhutdinov}.}
  \bibinfo{year}{2019}\natexlab{}.
\newblock \showarticletitle{Transformer-{XL}: Attentive Language Models beyond
  a Fixed-Length Context}. In \bibinfo{booktitle}{\emph{Proceedings of the 57th
  Annual Meeting of the Association for Computational Linguistics}}.
  \bibinfo{publisher}{Association for Computational Linguistics},
  \bibinfo{address}{Florence, Italy}, \bibinfo{pages}{2978--2988}.
\newblock
\urldef\tempurl%
\url{https://doi.org/10.18653/v1/P19-1285}
\showDOI{\tempurl}


\bibitem[\protect\citeauthoryear{Dathathri, Madotto, Lan, Hung, Frank, Molino,
  Yosinski, and Liu}{Dathathri et~al\mbox{.}}{2020}]%
        {pplm}
\bibfield{author}{\bibinfo{person}{Sumanth Dathathri}, \bibinfo{person}{Andrea
  Madotto}, \bibinfo{person}{Janice Lan}, \bibinfo{person}{Jane Hung},
  \bibinfo{person}{Eric Frank}, \bibinfo{person}{Piero Molino},
  \bibinfo{person}{Jason Yosinski}, {and} \bibinfo{person}{Rosanne Liu}.}
  \bibinfo{year}{2020}\natexlab{}.
\newblock \showarticletitle{Plug and Play Language Models: A Simple Approach to
  Controlled Text Generation}. In \bibinfo{booktitle}{\emph{International
  Conference on Learning Representations}}.
\newblock
\urldef\tempurl%
\url{https://openreview.net/forum?id=H1edEyBKDS}
\showURL{%
\tempurl}


\bibitem[\protect\citeauthoryear{Deng, Bakhtin, Ott, Szlam, and Ranzato}{Deng
  et~al\mbox{.}}{2020}]%
        {ebm_ctg}
\bibfield{author}{\bibinfo{person}{Yuntian Deng}, \bibinfo{person}{Anton
  Bakhtin}, \bibinfo{person}{Myle Ott}, \bibinfo{person}{Arthur Szlam}, {and}
  \bibinfo{person}{Marc'Aurelio Ranzato}.} \bibinfo{year}{2020}\natexlab{}.
\newblock \showarticletitle{Residual Energy-Based Models for Text Generation}.
  In \bibinfo{booktitle}{\emph{International Conference on Learning
  Representations}}.
\newblock
\urldef\tempurl%
\url{https://openreview.net/forum?id=B1l4SgHKDH}
\showURL{%
\tempurl}


\bibitem[\protect\citeauthoryear{Denkowski, Dyer, and Lavie}{Denkowski
  et~al\mbox{.}}{2014}]%
        {denkowski2014learning}
\bibfield{author}{\bibinfo{person}{Michael Denkowski}, \bibinfo{person}{Chris
  Dyer}, {and} \bibinfo{person}{Alon Lavie}.} \bibinfo{year}{2014}\natexlab{}.
\newblock \showarticletitle{Learning from Post-Editing: Online Model Adaptation
  for Statistical Machine Translation}. In
  \bibinfo{booktitle}{\emph{Proceedings of the 14th Conference of the
  {E}uropean Chapter of the Association for Computational Linguistics}}.
  \bibinfo{publisher}{Association for Computational Linguistics},
  \bibinfo{address}{Gothenburg, Sweden}, \bibinfo{pages}{395--404}.
\newblock
\urldef\tempurl%
\url{https://doi.org/10.3115/v1/E14-1042}
\showDOI{\tempurl}


\bibitem[\protect\citeauthoryear{Devlin, Chang, Lee, and Toutanova}{Devlin
  et~al\mbox{.}}{2019}]%
        {bert}
\bibfield{author}{\bibinfo{person}{Jacob Devlin}, \bibinfo{person}{Ming-Wei
  Chang}, \bibinfo{person}{Kenton Lee}, {and} \bibinfo{person}{Kristina
  Toutanova}.} \bibinfo{year}{2019}\natexlab{}.
\newblock \showarticletitle{{BERT}: Pre-training of Deep Bidirectional
  Transformers for Language Understanding}. In
  \bibinfo{booktitle}{\emph{Proceedings of the 2019 Conference of the North
  {A}merican Chapter of the Association for Computational Linguistics: Human
  Language Technologies, Volume 1 (Long and Short Papers)}}.
  \bibinfo{publisher}{Association for Computational Linguistics},
  \bibinfo{address}{Minneapolis, Minnesota}, \bibinfo{pages}{4171--4186}.
\newblock
\urldef\tempurl%
\url{https://doi.org/10.18653/v1/N19-1423}
\showDOI{\tempurl}


\bibitem[\protect\citeauthoryear{Dinan, Fan, Williams, Urbanek, Kiela, and
  Weston}{Dinan et~al\mbox{.}}{2020}]%
        {dinan2019queens}
\bibfield{author}{\bibinfo{person}{Emily Dinan}, \bibinfo{person}{Angela Fan},
  \bibinfo{person}{Adina Williams}, \bibinfo{person}{Jack Urbanek},
  \bibinfo{person}{Douwe Kiela}, {and} \bibinfo{person}{Jason Weston}.}
  \bibinfo{year}{2020}\natexlab{}.
\newblock \showarticletitle{Queens are Powerful too: Mitigating Gender Bias in
  Dialogue Generation}. In \bibinfo{booktitle}{\emph{Proceedings of the 2020
  Conference on Empirical Methods in Natural Language Processing (EMNLP)}}.
  \bibinfo{publisher}{Association for Computational Linguistics},
  \bibinfo{address}{Online}, \bibinfo{pages}{8173--8188}.
\newblock
\urldef\tempurl%
\url{https://doi.org/10.18653/v1/2020.emnlp-main.656}
\showDOI{\tempurl}


\bibitem[\protect\citeauthoryear{Dong, Yang, Wang, Wei, Liu, Wang, Gao, Zhou,
  and Hon}{Dong et~al\mbox{.}}{2019}]%
        {UniLM}
\bibfield{author}{\bibinfo{person}{Li Dong}, \bibinfo{person}{Nan Yang},
  \bibinfo{person}{Wenhui Wang}, \bibinfo{person}{Furu Wei},
  \bibinfo{person}{Xiaodong Liu}, \bibinfo{person}{Yu Wang},
  \bibinfo{person}{Jianfeng Gao}, \bibinfo{person}{Ming Zhou}, {and}
  \bibinfo{person}{Hsiao-Wuen Hon}.} \bibinfo{year}{2019}\natexlab{}.
\newblock \bibinfo{booktitle}{\emph{Unified Language Model Pre-Training for
  Natural Language Understanding and Generation}}.
\newblock \bibinfo{publisher}{Curran Associates Inc.}, \bibinfo{address}{Red
  Hook, NY, USA}.
\newblock
\urldef\tempurl%
\url{https://dl.acm.org/doi/pdf/10.5555/3454287.3455457}
\showURL{%
\tempurl}


\bibitem[\protect\citeauthoryear{Dong, Cordonnier, and Loukas}{Dong
  et~al\mbox{.}}{2021}]%
        {low_rank}
\bibfield{author}{\bibinfo{person}{Yihe Dong}, \bibinfo{person}{Jean{-}Baptiste
  Cordonnier}, {and} \bibinfo{person}{Andreas Loukas}.}
  \bibinfo{year}{2021}\natexlab{}.
\newblock \showarticletitle{Attention is Not All You Need: Pure Attention Loses
  Rank Doubly Exponentially with Depth}.
\newblock \bibinfo{journal}{\emph{CoRR}}  \bibinfo{volume}{abs/2103.03404}
  (\bibinfo{year}{2021}).
\newblock
\showeprint[arXiv]{2103.03404}
\urldef\tempurl%
\url{https://arxiv.org/abs/2103.03404}
\showURL{%
\tempurl}


\bibitem[\protect\citeauthoryear{Du, Shao, and Cardie}{Du
  et~al\mbox{.}}{2017}]%
        {du2017learning}
\bibfield{author}{\bibinfo{person}{Xinya Du}, \bibinfo{person}{Junru Shao},
  {and} \bibinfo{person}{Claire Cardie}.} \bibinfo{year}{2017}\natexlab{}.
\newblock \showarticletitle{Learning to Ask: Neural Question Generation for
  Reading Comprehension}. In \bibinfo{booktitle}{\emph{Proceedings of the 55th
  Annual Meeting of the Association for Computational Linguistics (Volume 1:
  Long Papers)}}. \bibinfo{publisher}{Association for Computational
  Linguistics}, \bibinfo{address}{Vancouver, Canada},
  \bibinfo{pages}{1342--1352}.
\newblock
\urldef\tempurl%
\url{https://doi.org/10.18653/v1/P17-1123}
\showDOI{\tempurl}


\bibitem[\protect\citeauthoryear{Fan, Lewis, and Dauphin}{Fan
  et~al\mbox{.}}{2018}]%
        {top_k}
\bibfield{author}{\bibinfo{person}{Angela Fan}, \bibinfo{person}{Mike Lewis},
  {and} \bibinfo{person}{Yann Dauphin}.} \bibinfo{year}{2018}\natexlab{}.
\newblock \showarticletitle{Hierarchical Neural Story Generation}. In
  \bibinfo{booktitle}{\emph{Proceedings of the 56th Annual Meeting of the
  Association for Computational Linguistics (Volume 1: Long Papers)}}.
  \bibinfo{publisher}{Association for Computational Linguistics},
  \bibinfo{address}{Melbourne, Australia}, \bibinfo{pages}{889--898}.
\newblock
\urldef\tempurl%
\url{https://doi.org/10.18653/v1/P18-1082}
\showDOI{\tempurl}


\bibitem[\protect\citeauthoryear{Fang, Zeng, Liu, Bo, Dong, and Chen}{Fang
  et~al\mbox{.}}{2021}]%
        {fang2021outline}
\bibfield{author}{\bibinfo{person}{Le Fang}, \bibinfo{person}{Tao Zeng},
  \bibinfo{person}{Chaochun Liu}, \bibinfo{person}{Liefeng Bo},
  \bibinfo{person}{Wen Dong}, {and} \bibinfo{person}{Changyou Chen}.}
  \bibinfo{year}{2021}\natexlab{}.
\newblock \showarticletitle{Outline to Story: Fine-grained Controllable Story
  Generation from Cascaded Events}.
\newblock \bibinfo{journal}{\emph{arXiv preprint arXiv:2101.00822}}
  (\bibinfo{year}{2021}).
\newblock
\urldef\tempurl%
\url{https://arxiv.org/pdf/2101.00822v1.pdf}
\showURL{%
\tempurl}


\bibitem[\protect\citeauthoryear{Ficler and Goldberg}{Ficler and
  Goldberg}{2017}]%
        {rnn_style}
\bibfield{author}{\bibinfo{person}{Jessica Ficler} {and} \bibinfo{person}{Yoav
  Goldberg}.} \bibinfo{year}{2017}\natexlab{}.
\newblock \showarticletitle{Controlling Linguistic Style Aspects in Neural
  Language Generation}. In \bibinfo{booktitle}{\emph{Proceedings of the
  Workshop on Stylistic Variation}}. \bibinfo{publisher}{Association for
  Computational Linguistics}, \bibinfo{address}{Copenhagen, Denmark},
  \bibinfo{pages}{94--104}.
\newblock
\urldef\tempurl%
\url{https://doi.org/10.18653/v1/W17-4912}
\showDOI{\tempurl}


\bibitem[\protect\citeauthoryear{Firdaus, Chauhan, Ekbal, and
  Bhattacharyya}{Firdaus et~al\mbox{.}}{2022a}]%
        {EmoSen_2020_tac}
\bibfield{author}{\bibinfo{person}{Mauajama Firdaus}, \bibinfo{person}{Hardik
  Chauhan}, \bibinfo{person}{Asif Ekbal}, {and} \bibinfo{person}{Pushpak
  Bhattacharyya}.} \bibinfo{year}{2022}\natexlab{a}.
\newblock \showarticletitle{EmoSen: Generating Sentiment and Emotion Controlled
  Responses in a Multimodal Dialogue System}.
\newblock \bibinfo{journal}{\emph{IEEE Transactions on Affective Computing}}
  \bibinfo{volume}{13}, \bibinfo{number}{3} (\bibinfo{year}{2022}),
  \bibinfo{pages}{1555--1566}.
\newblock
\urldef\tempurl%
\url{https://doi.org/10.1109/TAFFC.2020.3015491}
\showDOI{\tempurl}


\bibitem[\protect\citeauthoryear{Firdaus, Shandilya, Ekbal, and
  Bhattacharyya}{Firdaus et~al\mbox{.}}{2022b}]%
        {polite_trans_css}
\bibfield{author}{\bibinfo{person}{Mauajama Firdaus}, \bibinfo{person}{Arunav
  Shandilya}, \bibinfo{person}{Asif Ekbal}, {and} \bibinfo{person}{Pushpak
  Bhattacharyya}.} \bibinfo{year}{2022}\natexlab{b}.
\newblock \showarticletitle{Being Polite: Modeling Politeness Variation in a
  Personalized Dialog Agent}.
\newblock \bibinfo{journal}{\emph{IEEE Transactions on Computational Social
  Systems}} (\bibinfo{year}{2022}), \bibinfo{pages}{1--10}.
\newblock
\urldef\tempurl%
\url{https://doi.org/10.1109/TCSS.2022.3182986}
\showDOI{\tempurl}


\bibitem[\protect\citeauthoryear{Gao, Fisch, and Chen}{Gao
  et~al\mbox{.}}{2021}]%
        {gao-etal-2021-making}
\bibfield{author}{\bibinfo{person}{Tianyu Gao}, \bibinfo{person}{Adam Fisch},
  {and} \bibinfo{person}{Danqi Chen}.} \bibinfo{year}{2021}\natexlab{}.
\newblock \showarticletitle{Making Pre-trained Language Models Better Few-shot
  Learners}. In \bibinfo{booktitle}{\emph{Proceedings of the 59th Annual
  Meeting of the Association for Computational Linguistics and the 11th
  International Joint Conference on Natural Language Processing (Volume 1: Long
  Papers)}}. \bibinfo{publisher}{Association for Computational Linguistics},
  \bibinfo{address}{Online}, \bibinfo{pages}{3816--3830}.
\newblock
\urldef\tempurl%
\url{https://doi.org/10.18653/v1/2021.acl-long.295}
\showDOI{\tempurl}


\bibitem[\protect\citeauthoryear{Gao, Zhang, Lee, Galley, Brockett, Gao, and
  Dolan}{Gao et~al\mbox{.}}{2019}]%
        {gao2019structuring}
\bibfield{author}{\bibinfo{person}{Xiang Gao}, \bibinfo{person}{Yizhe Zhang},
  \bibinfo{person}{Sungjin Lee}, \bibinfo{person}{Michel Galley},
  \bibinfo{person}{Chris Brockett}, \bibinfo{person}{Jianfeng Gao}, {and}
  \bibinfo{person}{Bill Dolan}.} \bibinfo{year}{2019}\natexlab{}.
\newblock \showarticletitle{Structuring Latent Spaces for Stylized Response
  Generation}. In \bibinfo{booktitle}{\emph{Proceedings of the 2019 Conference
  on Empirical Methods in Natural Language Processing and the 9th International
  Joint Conference on Natural Language Processing (EMNLP-IJCNLP)}}.
  \bibinfo{publisher}{Association for Computational Linguistics},
  \bibinfo{address}{Hong Kong, China}, \bibinfo{pages}{1814--1823}.
\newblock
\urldef\tempurl%
\url{https://doi.org/10.18653/v1/D19-1190}
\showDOI{\tempurl}


\bibitem[\protect\citeauthoryear{Gkatzia and Mahamood}{Gkatzia and
  Mahamood}{2015}]%
        {gkatzia2015snapshot}
\bibfield{author}{\bibinfo{person}{Dimitra Gkatzia} {and} \bibinfo{person}{Saad
  Mahamood}.} \bibinfo{year}{2015}\natexlab{}.
\newblock \showarticletitle{A Snapshot of {NLG} Evaluation Practices 2005 -
  2014}. In \bibinfo{booktitle}{\emph{Proceedings of the 15th {E}uropean
  Workshop on Natural Language Generation ({ENLG})}}.
  \bibinfo{publisher}{Association for Computational Linguistics},
  \bibinfo{address}{Brighton, UK}, \bibinfo{pages}{57--60}.
\newblock
\urldef\tempurl%
\url{https://doi.org/10.18653/v1/W15-4708}
\showDOI{\tempurl}


\bibitem[\protect\citeauthoryear{Goldfarb-Tarrant, Chakrabarty, Weischedel, and
  Peng}{Goldfarb-Tarrant et~al\mbox{.}}{2020}]%
        {goldfarb2020content}
\bibfield{author}{\bibinfo{person}{Seraphina Goldfarb-Tarrant},
  \bibinfo{person}{Tuhin Chakrabarty}, \bibinfo{person}{Ralph Weischedel},
  {and} \bibinfo{person}{Nanyun Peng}.} \bibinfo{year}{2020}\natexlab{}.
\newblock \showarticletitle{Content Planning for Neural Story Generation with
  Aristotelian Rescoring}. In \bibinfo{booktitle}{\emph{Proceedings of the 2020
  Conference on Empirical Methods in Natural Language Processing (EMNLP)}}.
  \bibinfo{publisher}{Association for Computational Linguistics},
  \bibinfo{address}{Online}, \bibinfo{pages}{4319--4338}.
\newblock
\urldef\tempurl%
\url{https://doi.org/10.18653/v1/2020.emnlp-main.351}
\showDOI{\tempurl}


\bibitem[\protect\citeauthoryear{Goodfellow, Pouget-Abadie, Mirza, Xu,
  Warde-Farley, Ozair, Courville, and Bengio}{Goodfellow et~al\mbox{.}}{2020}]%
        {GANs}
\bibfield{author}{\bibinfo{person}{Ian Goodfellow}, \bibinfo{person}{Jean
  Pouget-Abadie}, \bibinfo{person}{Mehdi Mirza}, \bibinfo{person}{Bing Xu},
  \bibinfo{person}{David Warde-Farley}, \bibinfo{person}{Sherjil Ozair},
  \bibinfo{person}{Aaron Courville}, {and} \bibinfo{person}{Yoshua Bengio}.}
  \bibinfo{year}{2020}\natexlab{}.
\newblock \showarticletitle{Generative adversarial networks}.
\newblock \bibinfo{journal}{\emph{Commun. ACM}} \bibinfo{volume}{63},
  \bibinfo{number}{11} (\bibinfo{year}{2020}), \bibinfo{pages}{139--144}.
\newblock


\bibitem[\protect\citeauthoryear{Grover, Song, Agarwal, Tran, Kapoor, Horvitz,
  and Ermon}{Grover et~al\mbox{.}}{2019}]%
        {bias_correction}
\bibfield{author}{\bibinfo{person}{Aditya Grover}, \bibinfo{person}{Jiaming
  Song}, \bibinfo{person}{Alekh Agarwal}, \bibinfo{person}{Kenneth Tran},
  \bibinfo{person}{Ashish Kapoor}, \bibinfo{person}{Eric Horvitz}, {and}
  \bibinfo{person}{Stefano Ermon}.} \bibinfo{year}{2019}\natexlab{}.
\newblock \bibinfo{booktitle}{\emph{Bias Correction of Learned Generative
  Models Using Likelihood-Free Importance Weighting}}.
\newblock \bibinfo{publisher}{Curran Associates Inc.}, \bibinfo{address}{Red
  Hook, NY, USA}.
\newblock
\urldef\tempurl%
\url{https://papers.nips.cc/paper/2019/file/d76d8deea9c19cc9aaf2237d2bf2f785-Paper.pdf}
\showURL{%
\tempurl}


\bibitem[\protect\citeauthoryear{Hashimoto, Zhang, and Liang}{Hashimoto
  et~al\mbox{.}}{2019}]%
        {hashimoto2019unifying}
\bibfield{author}{\bibinfo{person}{Tatsunori~B. Hashimoto},
  \bibinfo{person}{Hugh Zhang}, {and} \bibinfo{person}{Percy Liang}.}
  \bibinfo{year}{2019}\natexlab{}.
\newblock \showarticletitle{Unifying Human and Statistical Evaluation for
  Natural Language Generation}. In \bibinfo{booktitle}{\emph{Proceedings of the
  2019 Conference of the North {A}merican Chapter of the Association for
  Computational Linguistics: Human Language Technologies, Volume 1 (Long and
  Short Papers)}}. \bibinfo{publisher}{Association for Computational
  Linguistics}, \bibinfo{address}{Minneapolis, Minnesota},
  \bibinfo{pages}{1689--1701}.
\newblock
\urldef\tempurl%
\url{https://doi.org/10.18653/v1/N19-1169}
\showDOI{\tempurl}


\bibitem[\protect\citeauthoryear{He, Wang, Neubig, and Berg-Kirkpatrick}{He
  et~al\mbox{.}}{2020}]%
        {He2020APF}
\bibfield{author}{\bibinfo{person}{Junxian He}, \bibinfo{person}{Xinyi Wang},
  \bibinfo{person}{Graham Neubig}, {and} \bibinfo{person}{Taylor
  Berg-Kirkpatrick}.} \bibinfo{year}{2020}\natexlab{}.
\newblock \showarticletitle{A Probabilistic Formulation of Unsupervised Text
  Style Transfer}.
\newblock \bibinfo{journal}{\emph{ArXiv}}  \bibinfo{volume}{abs/2002.03912}
  (\bibinfo{year}{2020}).
\newblock


\bibitem[\protect\citeauthoryear{He}{He}{2021}]%
        {CBART}
\bibfield{author}{\bibinfo{person}{Xingwei He}.}
  \bibinfo{year}{2021}\natexlab{}.
\newblock \showarticletitle{Parallel Refinements for Lexically Constrained Text
  Generation with {BART}}. In \bibinfo{booktitle}{\emph{Proceedings of the 2021
  Conference on Empirical Methods in Natural Language Processing}}.
  \bibinfo{publisher}{Association for Computational Linguistics},
  \bibinfo{address}{Online and Punta Cana, Dominican Republic},
  \bibinfo{pages}{8653--8666}.
\newblock
\urldef\tempurl%
\url{https://doi.org/10.18653/v1/2021.emnlp-main.681}
\showDOI{\tempurl}


\bibitem[\protect\citeauthoryear{Holtzman, Buys, Du, Forbes, and Choi}{Holtzman
  et~al\mbox{.}}{2020}]%
        {Nucleus}
\bibfield{author}{\bibinfo{person}{Ari Holtzman}, \bibinfo{person}{Jan Buys},
  \bibinfo{person}{Li Du}, \bibinfo{person}{Maxwell Forbes}, {and}
  \bibinfo{person}{Yejin Choi}.} \bibinfo{year}{2020}\natexlab{}.
\newblock \showarticletitle{The Curious Case of Neural Text Degeneration}. In
  \bibinfo{booktitle}{\emph{International Conference on Learning
  Representations}}.
\newblock
\urldef\tempurl%
\url{https://openreview.net/forum?id=rygGQyrFvH}
\showURL{%
\tempurl}


\bibitem[\protect\citeauthoryear{Horacek}{Horacek}{2001}]%
        {nlu_nlg}
\bibfield{author}{\bibinfo{person}{Helmut Horacek}.}
  \bibinfo{year}{2001}\natexlab{}.
\newblock \showarticletitle{Building Natural Language Generation Systems - Ehud
  Reiter and Robert Dale (Eds.), University of Aberdeen and Macquarie
  University, Cambridge University Press, 2000, {ISBN} 0-521-62036-8}.
\newblock \bibinfo{journal}{\emph{Artif. Intell. Medicine}}
  \bibinfo{volume}{22}, \bibinfo{number}{3} (\bibinfo{year}{2001}),
  \bibinfo{pages}{277--280}.
\newblock
\urldef\tempurl%
\url{https://doi.org/10.1016/S0933-3657(00)00114-7}
\showDOI{\tempurl}


\bibitem[\protect\citeauthoryear{Hu, Yang, Liang, Salakhutdinov, and Xing}{Hu
  et~al\mbox{.}}{2017}]%
        {icml_ctg_vae}
\bibfield{author}{\bibinfo{person}{Zhiting Hu}, \bibinfo{person}{Zichao Yang},
  \bibinfo{person}{Xiaodan Liang}, \bibinfo{person}{Ruslan Salakhutdinov},
  {and} \bibinfo{person}{Eric~P. Xing}.} \bibinfo{year}{2017}\natexlab{}.
\newblock \showarticletitle{Controllable Text Generation}.
\newblock \bibinfo{journal}{\emph{CoRR}}  \bibinfo{volume}{abs/1703.00955}
  (\bibinfo{year}{2017}).
\newblock
\showeprint[arxiv]{1703.00955}
\urldef\tempurl%
\url{http://arxiv.org/abs/1703.00955}
\showURL{%
\tempurl}


\bibitem[\protect\citeauthoryear{Hua and Wang}{Hua and Wang}{2020}]%
        {pair_emnlp2020}
\bibfield{author}{\bibinfo{person}{Xinyu Hua} {and} \bibinfo{person}{Lu Wang}.}
  \bibinfo{year}{2020}\natexlab{}.
\newblock \showarticletitle{{PAIR}: Planning and Iterative Refinement in
  Pre-trained Transformers for Long Text Generation}. In
  \bibinfo{booktitle}{\emph{Proceedings of the 2020 Conference on Empirical
  Methods in Natural Language Processing (EMNLP)}}.
  \bibinfo{publisher}{Association for Computational Linguistics},
  \bibinfo{address}{Online}, \bibinfo{pages}{781--793}.
\newblock
\urldef\tempurl%
\url{https://doi.org/10.18653/v1/2020.emnlp-main.57}
\showDOI{\tempurl}


\bibitem[\protect\citeauthoryear{Huang, Zhu, and Gao}{Huang
  et~al\mbox{.}}{2020b}]%
        {huang2020challenges}
\bibfield{author}{\bibinfo{person}{Minlie Huang}, \bibinfo{person}{Xiaoyan
  Zhu}, {and} \bibinfo{person}{Jianfeng Gao}.}
  \bibinfo{year}{2020}\natexlab{b}.
\newblock \showarticletitle{Challenges in building intelligent open-domain
  dialog systems}.
\newblock \bibinfo{journal}{\emph{ACM Transactions on Information Systems
  (TOIS)}} \bibinfo{volume}{38}, \bibinfo{number}{3} (\bibinfo{year}{2020}),
  \bibinfo{pages}{1--32}.
\newblock


\bibitem[\protect\citeauthoryear{Huang, Zhang, Jiang, Stanforth, Welbl, Rae,
  Maini, Yogatama, and Kohli}{Huang et~al\mbox{.}}{2020a}]%
        {huang2019reducing}
\bibfield{author}{\bibinfo{person}{Po-Sen Huang}, \bibinfo{person}{Huan Zhang},
  \bibinfo{person}{Ray Jiang}, \bibinfo{person}{Robert Stanforth},
  \bibinfo{person}{Johannes Welbl}, \bibinfo{person}{Jack Rae},
  \bibinfo{person}{Vishal Maini}, \bibinfo{person}{Dani Yogatama}, {and}
  \bibinfo{person}{Pushmeet Kohli}.} \bibinfo{year}{2020}\natexlab{a}.
\newblock \showarticletitle{Reducing Sentiment Bias in Language Models via
  Counterfactual Evaluation}. In \bibinfo{booktitle}{\emph{Findings of the
  Association for Computational Linguistics: EMNLP 2020}}.
  \bibinfo{publisher}{Association for Computational Linguistics},
  \bibinfo{address}{Online}, \bibinfo{pages}{65--83}.
\newblock
\urldef\tempurl%
\url{https://doi.org/10.18653/v1/2020.findings-emnlp.7}
\showDOI{\tempurl}


\bibitem[\protect\citeauthoryear{Ipeirotis, Provost, and Wang}{Ipeirotis
  et~al\mbox{.}}{2010}]%
        {ipeirotis2010quality}
\bibfield{author}{\bibinfo{person}{Panagiotis~G. Ipeirotis},
  \bibinfo{person}{Foster Provost}, {and} \bibinfo{person}{Jing Wang}.}
  \bibinfo{year}{2010}\natexlab{}.
\newblock \showarticletitle{Quality Management on Amazon Mechanical Turk}. In
  \bibinfo{booktitle}{\emph{Proceedings of the ACM SIGKDD Workshop on Human
  Computation}} (Washington DC) \emph{(\bibinfo{series}{HCOMP '10})}.
  \bibinfo{publisher}{Association for Computing Machinery},
  \bibinfo{address}{New York, NY, USA}, \bibinfo{pages}{64–67}.
\newblock
\showISBNx{9781450302227}
\urldef\tempurl%
\url{https://doi.org/10.1145/1837885.1837906}
\showDOI{\tempurl}


\bibitem[\protect\citeauthoryear{Ji, Lee, Frieske, Yu, Su, Xu, Ishii, Bang,
  Madotto, and Fung}{Ji et~al\mbox{.}}{2023}]%
        {Hallucination}
\bibfield{author}{\bibinfo{person}{Ziwei Ji}, \bibinfo{person}{Nayeon Lee},
  \bibinfo{person}{Rita Frieske}, \bibinfo{person}{Tiezheng Yu},
  \bibinfo{person}{Dan Su}, \bibinfo{person}{Yan Xu}, \bibinfo{person}{Etsuko
  Ishii}, \bibinfo{person}{Ye~Jin Bang}, \bibinfo{person}{Andrea Madotto},
  {and} \bibinfo{person}{Pascale Fung}.} \bibinfo{year}{2023}\natexlab{}.
\newblock \showarticletitle{Survey of Hallucination in Natural Language
  Generation}.
\newblock \bibinfo{journal}{\emph{ACM Comput. Surv.}} \bibinfo{volume}{55},
  \bibinfo{number}{12}, Article \bibinfo{articleno}{248} (\bibinfo{date}{mar}
  \bibinfo{year}{2023}), \bibinfo{numpages}{38}~pages.
\newblock
\showISSN{0360-0300}
\urldef\tempurl%
\url{https://doi.org/10.1145/3571730}
\showDOI{\tempurl}


\bibitem[\protect\citeauthoryear{Jiang, Xu, Araki, and Neubig}{Jiang
  et~al\mbox{.}}{2020}]%
        {TACL2020}
\bibfield{author}{\bibinfo{person}{Zhengbao Jiang}, \bibinfo{person}{Frank~F.
  Xu}, \bibinfo{person}{Jun Araki}, {and} \bibinfo{person}{Graham Neubig}.}
  \bibinfo{year}{2020}\natexlab{}.
\newblock \showarticletitle{How Can We Know What Language Models Know?}
\newblock \bibinfo{journal}{\emph{Transactions of the Association for
  Computational Linguistics}}  \bibinfo{volume}{8} (\bibinfo{year}{2020}),
  \bibinfo{pages}{423--438}.
\newblock
\urldef\tempurl%
\url{https://doi.org/10.1162/tacl_a_00324}
\showDOI{\tempurl}


\bibitem[\protect\citeauthoryear{Kale and Rastogi}{Kale and Rastogi}{2020a}]%
        {kale-rastogi-2020-template}
\bibfield{author}{\bibinfo{person}{Mihir Kale} {and} \bibinfo{person}{Abhinav
  Rastogi}.} \bibinfo{year}{2020}\natexlab{a}.
\newblock \showarticletitle{Template Guided Text Generation for Task-Oriented
  Dialogue}. In \bibinfo{booktitle}{\emph{Proceedings of the 2020 Conference on
  Empirical Methods in Natural Language Processing (EMNLP)}}.
  \bibinfo{publisher}{Association for Computational Linguistics},
  \bibinfo{address}{Online}, \bibinfo{pages}{6505--6520}.
\newblock
\urldef\tempurl%
\url{https://doi.org/10.18653/v1/2020.emnlp-main.527}
\showDOI{\tempurl}


\bibitem[\protect\citeauthoryear{Kale and Rastogi}{Kale and Rastogi}{2020b}]%
        {t2t_plm}
\bibfield{author}{\bibinfo{person}{Mihir Kale} {and} \bibinfo{person}{Abhinav
  Rastogi}.} \bibinfo{year}{2020}\natexlab{b}.
\newblock \showarticletitle{Text-to-Text Pre-Training for Data-to-Text Tasks}.
  In \bibinfo{booktitle}{\emph{Proceedings of the 13th International Conference
  on Natural Language Generation}}. \bibinfo{publisher}{Association for
  Computational Linguistics}, \bibinfo{address}{Dublin, Ireland},
  \bibinfo{pages}{97--102}.
\newblock
\urldef\tempurl%
\url{https://aclanthology.org/2020.inlg-1.14}
\showURL{%
\tempurl}


\bibitem[\protect\citeauthoryear{Ke, Zhou, Lin, Li, Zhou, Zhu, and Huang}{Ke
  et~al\mbox{.}}{2022}]%
        {ke-etal-2022-ctrleval}
\bibfield{author}{\bibinfo{person}{Pei Ke}, \bibinfo{person}{Hao Zhou},
  \bibinfo{person}{Yankai Lin}, \bibinfo{person}{Peng Li}, \bibinfo{person}{Jie
  Zhou}, \bibinfo{person}{Xiaoyan Zhu}, {and} \bibinfo{person}{Minlie Huang}.}
  \bibinfo{year}{2022}\natexlab{}.
\newblock \showarticletitle{{CTRLE}val: An Unsupervised Reference-Free Metric
  for Evaluating Controlled Text Generation}. In
  \bibinfo{booktitle}{\emph{Proceedings of the 60th Annual Meeting of the
  Association for Computational Linguistics (Volume 1: Long Papers)}}.
  \bibinfo{publisher}{Association for Computational Linguistics},
  \bibinfo{address}{Dublin, Ireland}, \bibinfo{pages}{2306--2319}.
\newblock
\urldef\tempurl%
\url{https://doi.org/10.18653/v1/2022.acl-long.164}
\showDOI{\tempurl}


\bibitem[\protect\citeauthoryear{Keskar, McCann, Varshney, Xiong, and
  Socher}{Keskar et~al\mbox{.}}{2019}]%
        {CTRL}
\bibfield{author}{\bibinfo{person}{Nitish~Shirish Keskar},
  \bibinfo{person}{Bryan McCann}, \bibinfo{person}{Lav~R. Varshney},
  \bibinfo{person}{Caiming Xiong}, {and} \bibinfo{person}{Richard Socher}.}
  \bibinfo{year}{2019}\natexlab{}.
\newblock \showarticletitle{{CTRL:} {A} Conditional Transformer Language Model
  for Controllable Generation}.
\newblock \bibinfo{journal}{\emph{CoRR}}  \bibinfo{volume}{abs/1909.05858}
  (\bibinfo{year}{2019}).
\newblock
\showeprint[arxiv]{1909.05858}
\urldef\tempurl%
\url{http://arxiv.org/abs/1909.05858}
\showURL{%
\tempurl}


\bibitem[\protect\citeauthoryear{Khalifa, Elsahar, and Dymetman}{Khalifa
  et~al\mbox{.}}{2021}]%
        {gdc}
\bibfield{author}{\bibinfo{person}{Muhammad Khalifa}, \bibinfo{person}{Hady
  Elsahar}, {and} \bibinfo{person}{Marc Dymetman}.}
  \bibinfo{year}{2021}\natexlab{}.
\newblock \showarticletitle{A Distributional Approach to Controlled Text
  Generation}. In \bibinfo{booktitle}{\emph{International Conference on
  Learning Representations}}.
\newblock
\urldef\tempurl%
\url{https://openreview.net/forum?id=jWkw45-9AbL}
\showURL{%
\tempurl}


\bibitem[\protect\citeauthoryear{Kingma and Welling}{Kingma and
  Welling}{2013}]%
        {VAEs}
\bibfield{author}{\bibinfo{person}{Diederik~P Kingma} {and}
  \bibinfo{person}{Max Welling}.} \bibinfo{year}{2013}\natexlab{}.
\newblock \showarticletitle{Auto-encoding variational bayes}.
\newblock \bibinfo{journal}{\emph{arXiv preprint arXiv:1312.6114}}
  (\bibinfo{year}{2013}).
\newblock


\bibitem[\protect\citeauthoryear{Kitaev, Kaiser, and Levskaya}{Kitaev
  et~al\mbox{.}}{2020}]%
        {reformer}
\bibfield{author}{\bibinfo{person}{Nikita Kitaev}, \bibinfo{person}{Lukasz
  Kaiser}, {and} \bibinfo{person}{Anselm Levskaya}.}
  \bibinfo{year}{2020}\natexlab{}.
\newblock \showarticletitle{Reformer: The Efficient Transformer}. In
  \bibinfo{booktitle}{\emph{International Conference on Learning
  Representations}}.
\newblock
\urldef\tempurl%
\url{https://openreview.net/forum?id=rkgNKkHtvB}
\showURL{%
\tempurl}


\bibitem[\protect\citeauthoryear{Krause, Gotmare, McCann, Keskar, Joty, Socher,
  and Rajani}{Krause et~al\mbox{.}}{2021}]%
        {GeDi}
\bibfield{author}{\bibinfo{person}{Ben Krause},
  \bibinfo{person}{Akhilesh~Deepak Gotmare}, \bibinfo{person}{Bryan McCann},
  \bibinfo{person}{Nitish~Shirish Keskar}, \bibinfo{person}{Shafiq Joty},
  \bibinfo{person}{Richard Socher}, {and} \bibinfo{person}{Nazneen~Fatema
  Rajani}.} \bibinfo{year}{2021}\natexlab{}.
\newblock \showarticletitle{{G}e{D}i: Generative Discriminator Guided Sequence
  Generation}. In \bibinfo{booktitle}{\emph{Findings of the Association for
  Computational Linguistics: EMNLP 2021}}. \bibinfo{publisher}{Association for
  Computational Linguistics}, \bibinfo{address}{Punta Cana, Dominican
  Republic}, \bibinfo{pages}{4929--4952}.
\newblock
\urldef\tempurl%
\url{https://doi.org/10.18653/v1/2021.findings-emnlp.424}
\showDOI{\tempurl}


\bibitem[\protect\citeauthoryear{Lester, Al-Rfou, and Constant}{Lester
  et~al\mbox{.}}{2021}]%
        {p_tuning}
\bibfield{author}{\bibinfo{person}{Brian Lester}, \bibinfo{person}{Rami
  Al-Rfou}, {and} \bibinfo{person}{Noah Constant}.}
  \bibinfo{year}{2021}\natexlab{}.
\newblock \showarticletitle{The Power of Scale for Parameter-Efficient Prompt
  Tuning}. In \bibinfo{booktitle}{\emph{Proceedings of the 2021 Conference on
  Empirical Methods in Natural Language Processing}}.
  \bibinfo{publisher}{Association for Computational Linguistics},
  \bibinfo{address}{Online and Punta Cana, Dominican Republic},
  \bibinfo{pages}{3045--3059}.
\newblock
\urldef\tempurl%
\url{https://doi.org/10.18653/v1/2021.emnlp-main.243}
\showDOI{\tempurl}


\bibitem[\protect\citeauthoryear{Li, Galley, Brockett, Gao, and Dolan}{Li
  et~al\mbox{.}}{2016a}]%
        {li2016diversity}
\bibfield{author}{\bibinfo{person}{Jiwei Li}, \bibinfo{person}{Michel Galley},
  \bibinfo{person}{Chris Brockett}, \bibinfo{person}{Jianfeng Gao}, {and}
  \bibinfo{person}{Bill Dolan}.} \bibinfo{year}{2016}\natexlab{a}.
\newblock \showarticletitle{A Diversity-Promoting Objective Function for Neural
  Conversation Models}. In \bibinfo{booktitle}{\emph{Proceedings of the 2016
  Conference of the North {A}merican Chapter of the Association for
  Computational Linguistics: Human Language Technologies}}.
  \bibinfo{publisher}{Association for Computational Linguistics},
  \bibinfo{address}{San Diego, California}, \bibinfo{pages}{110--119}.
\newblock
\urldef\tempurl%
\url{https://doi.org/10.18653/v1/N16-1014}
\showDOI{\tempurl}


\bibitem[\protect\citeauthoryear{Li, Galley, Brockett, Spithourakis, Gao, and
  Dolan}{Li et~al\mbox{.}}{2016b}]%
        {li2016persona}
\bibfield{author}{\bibinfo{person}{Jiwei Li}, \bibinfo{person}{Michel Galley},
  \bibinfo{person}{Chris Brockett}, \bibinfo{person}{Georgios Spithourakis},
  \bibinfo{person}{Jianfeng Gao}, {and} \bibinfo{person}{Bill Dolan}.}
  \bibinfo{year}{2016}\natexlab{b}.
\newblock \showarticletitle{A Persona-Based Neural Conversation Model}. In
  \bibinfo{booktitle}{\emph{Proceedings of the 54th Annual Meeting of the
  Association for Computational Linguistics (Volume 1: Long Papers)}}.
  \bibinfo{publisher}{Association for Computational Linguistics},
  \bibinfo{address}{Berlin, Germany}, \bibinfo{pages}{994--1003}.
\newblock
\urldef\tempurl%
\url{https://doi.org/10.18653/v1/P16-1094}
\showDOI{\tempurl}


\bibitem[\protect\citeauthoryear{Li, Zhang, Liu, and Shi}{Li
  et~al\mbox{.}}{2020b}]%
        {rigid_format_acl2020}
\bibfield{author}{\bibinfo{person}{Piji Li}, \bibinfo{person}{Haisong Zhang},
  \bibinfo{person}{Xiaojiang Liu}, {and} \bibinfo{person}{Shuming Shi}.}
  \bibinfo{year}{2020}\natexlab{b}.
\newblock \showarticletitle{Rigid Formats Controlled Text Generation}. In
  \bibinfo{booktitle}{\emph{Proceedings of the 58th Annual Meeting of the
  Association for Computational Linguistics, {ACL} 2020, Online, July 5-10,
  2020}}, \bibfield{editor}{\bibinfo{person}{Dan Jurafsky},
  \bibinfo{person}{Joyce Chai}, \bibinfo{person}{Natalie Schluter}, {and}
  \bibinfo{person}{Joel~R. Tetreault}} (Eds.). \bibinfo{publisher}{Association
  for Computational Linguistics}, \bibinfo{pages}{742--751}.
\newblock
\urldef\tempurl%
\url{https://doi.org/10.18653/v1/2020.acl-main.68}
\showDOI{\tempurl}


\bibitem[\protect\citeauthoryear{Li, Feng, Wang, Song, Zhang, and Wang}{Li
  et~al\mbox{.}}{2020a}]%
        {emotion_dialogue_generation}
\bibfield{author}{\bibinfo{person}{Shifeng Li}, \bibinfo{person}{Shi Feng},
  \bibinfo{person}{Daling Wang}, \bibinfo{person}{Kaisong Song},
  \bibinfo{person}{Yifei Zhang}, {and} \bibinfo{person}{Weichao Wang}.}
  \bibinfo{year}{2020}\natexlab{a}.
\newblock \showarticletitle{EmoElicitor: An Open Domain Response Generation
  Model with User Emotional Reaction Awareness}. In
  \bibinfo{booktitle}{\emph{Proceedings of the Twenty-Ninth International Joint
  Conference on Artificial Intelligence, {IJCAI-20}}},
  \bibfield{editor}{\bibinfo{person}{Christian Bessiere}} (Ed.).
  \bibinfo{publisher}{International Joint Conferences on Artificial
  Intelligence Organization}, \bibinfo{pages}{3637--3643}.
\newblock
\urldef\tempurl%
\url{https://doi.org/10.24963/ijcai.2020/503}
\showDOI{\tempurl}
\newblock
\shownote{Main track.}


\bibitem[\protect\citeauthoryear{Li and Liang}{Li and Liang}{2021}]%
        {prefix_tuning}
\bibfield{author}{\bibinfo{person}{Xiang~Lisa Li} {and} \bibinfo{person}{Percy
  Liang}.} \bibinfo{year}{2021}\natexlab{}.
\newblock \showarticletitle{Prefix-Tuning: Optimizing Continuous Prompts for
  Generation}. In \bibinfo{booktitle}{\emph{Proceedings of the 59th Annual
  Meeting of the Association for Computational Linguistics and the 11th
  International Joint Conference on Natural Language Processing (Volume 1: Long
  Papers)}}. \bibinfo{publisher}{Association for Computational Linguistics},
  \bibinfo{address}{Online}, \bibinfo{pages}{4582--4597}.
\newblock
\urldef\tempurl%
\url{https://doi.org/10.18653/v1/2021.acl-long.353}
\showDOI{\tempurl}


\bibitem[\protect\citeauthoryear{Li, Thickstun, Gulrajani, Liang, and
  Hashimoto}{Li et~al\mbox{.}}{2022}]%
        {Li-2022-DiffusionLM}
\bibfield{author}{\bibinfo{person}{Xiang~Lisa Li}, \bibinfo{person}{John
  Thickstun}, \bibinfo{person}{Ishaan Gulrajani}, \bibinfo{person}{Percy
  Liang}, {and} \bibinfo{person}{Tatsunori Hashimoto}.}
  \bibinfo{year}{2022}\natexlab{}.
\newblock \showarticletitle{Diffusion-LM Improves Controllable Text
  Generation}.
\newblock \bibinfo{journal}{\emph{ArXiv}}  \bibinfo{volume}{abs/2205.14217}
  (\bibinfo{year}{2022}).
\newblock
\urldef\tempurl%
\url{https://arxiv.org/abs/2205.14217}
\showURL{%
\tempurl}


\bibitem[\protect\citeauthoryear{Liao, Wang, Liu, and Jiang}{Liao
  et~al\mbox{.}}{2019}]%
        {liao2019gpt}
\bibfield{author}{\bibinfo{person}{Yi Liao}, \bibinfo{person}{Yasheng Wang},
  \bibinfo{person}{Qun Liu}, {and} \bibinfo{person}{Xin Jiang}.}
  \bibinfo{year}{2019}\natexlab{}.
\newblock \showarticletitle{Gpt-based generation for classical chinese poetry}.
\newblock \bibinfo{journal}{\emph{arXiv preprint arXiv:1907.00151}}
  (\bibinfo{year}{2019}).
\newblock
\urldef\tempurl%
\url{https://arxiv.org/abs/1907.00151}
\showURL{%
\tempurl}


\bibitem[\protect\citeauthoryear{Lin}{Lin}{2004}]%
        {lin2004rouge}
\bibfield{author}{\bibinfo{person}{Chin-Yew Lin}.}
  \bibinfo{year}{2004}\natexlab{}.
\newblock \showarticletitle{{ROUGE}: A Package for Automatic Evaluation of
  Summaries}. In \bibinfo{booktitle}{\emph{Text Summarization Branches Out}}.
  \bibinfo{publisher}{Association for Computational Linguistics},
  \bibinfo{address}{Barcelona, Spain}, \bibinfo{pages}{74--81}.
\newblock
\urldef\tempurl%
\url{https://aclanthology.org/W04-1013}
\showURL{%
\tempurl}


\bibitem[\protect\citeauthoryear{Lin, Madotto, Bang, and Fung}{Lin
  et~al\mbox{.}}{2021}]%
        {aaai2021_Adapter_Bot}
\bibfield{author}{\bibinfo{person}{Zhaojiang Lin}, \bibinfo{person}{Andrea
  Madotto}, \bibinfo{person}{Yejin Bang}, {and} \bibinfo{person}{Pascale
  Fung}.} \bibinfo{year}{2021}\natexlab{}.
\newblock \showarticletitle{The Adapter-Bot: All-In-One Controllable
  Conversational Model}. In \bibinfo{booktitle}{\emph{Thirty-Fifth {AAAI}
  Conference on Artificial Intelligence, {AAAI} 2021, Thirty-Third Conference
  on Innovative Applications of Artificial Intelligence, {IAAI} 2021, The
  Eleventh Symposium on Educational Advances in Artificial Intelligence, {EAAI}
  2021, Virtual Event, February 2-9, 2021}}. \bibinfo{publisher}{{AAAI} Press},
  \bibinfo{pages}{16081--16083}.
\newblock
\urldef\tempurl%
\url{https://ojs.aaai.org/index.php/AAAI/article/view/18018}
\showURL{%
\tempurl}


\bibitem[\protect\citeauthoryear{Lin and Riedl}{Lin and Riedl}{2021}]%
        {lin2021plug}
\bibfield{author}{\bibinfo{person}{Zhiyu Lin} {and} \bibinfo{person}{Mark~O
  Riedl}.} \bibinfo{year}{2021}\natexlab{}.
\newblock \showarticletitle{Plug-and-Blend: A Framework for Plug-and-Play
  Controllable Story Generation with Sketches}. In
  \bibinfo{booktitle}{\emph{Proceedings of the AAAI Conference on Artificial
  Intelligence and Interactive Digital Entertainment}},
  Vol.~\bibinfo{volume}{17}. \bibinfo{pages}{58--65}.
\newblock
\urldef\tempurl%
\url{https://arxiv.org/abs/2104.04039}
\showURL{%
\tempurl}


\bibitem[\protect\citeauthoryear{Liu, Sap, Lu, Swayamdipta, Bhagavatula, Smith,
  and Choi}{Liu et~al\mbox{.}}{2021c}]%
        {liu-etal-2021-dexperts}
\bibfield{author}{\bibinfo{person}{Alisa Liu}, \bibinfo{person}{Maarten Sap},
  \bibinfo{person}{Ximing Lu}, \bibinfo{person}{Swabha Swayamdipta},
  \bibinfo{person}{Chandra Bhagavatula}, \bibinfo{person}{Noah~A. Smith}, {and}
  \bibinfo{person}{Yejin Choi}.} \bibinfo{year}{2021}\natexlab{c}.
\newblock \showarticletitle{{DE}xperts: Decoding-Time Controlled Text
  Generation with Experts and Anti-Experts}. In
  \bibinfo{booktitle}{\emph{Proceedings of the 59th Annual Meeting of the
  Association for Computational Linguistics and the 11th International Joint
  Conference on Natural Language Processing (Volume 1: Long Papers)}}.
  \bibinfo{publisher}{Association for Computational Linguistics},
  \bibinfo{address}{Online}, \bibinfo{pages}{6691--6706}.
\newblock
\urldef\tempurl%
\url{https://doi.org/10.18653/v1/2021.acl-long.522}
\showDOI{\tempurl}


\bibitem[\protect\citeauthoryear{Liu, Dahlmeier, and Ng}{Liu
  et~al\mbox{.}}{2010}]%
        {liu2010tesla}
\bibfield{author}{\bibinfo{person}{Chang Liu}, \bibinfo{person}{Daniel
  Dahlmeier}, {and} \bibinfo{person}{Hwee~Tou Ng}.}
  \bibinfo{year}{2010}\natexlab{}.
\newblock \showarticletitle{{TESLA}: Translation Evaluation of Sentences with
  Linear-Programming-Based Analysis}. In \bibinfo{booktitle}{\emph{Proceedings
  of the Joint Fifth Workshop on Statistical Machine Translation and
  {M}etrics{MATR}}}. \bibinfo{publisher}{Association for Computational
  Linguistics}, \bibinfo{address}{Uppsala, Sweden}, \bibinfo{pages}{354--359}.
\newblock
\urldef\tempurl%
\url{https://aclanthology.org/W10-1754}
\showURL{%
\tempurl}


\bibitem[\protect\citeauthoryear{Liu, Li, Yu, Huang, Liu, Zhao, and Yan}{Liu
  et~al\mbox{.}}{2020b}]%
        {liu2020character}
\bibfield{author}{\bibinfo{person}{Danyang Liu}, \bibinfo{person}{Juntao Li},
  \bibinfo{person}{Meng-Hsuan Yu}, \bibinfo{person}{Ziming Huang},
  \bibinfo{person}{Gongshen Liu}, \bibinfo{person}{Dongyan Zhao}, {and}
  \bibinfo{person}{Rui Yan}.} \bibinfo{year}{2020}\natexlab{b}.
\newblock \showarticletitle{A character-centric neural model for automated
  story generation}. In \bibinfo{booktitle}{\emph{Proceedings of the AAAI
  Conference on Artificial Intelligence}}, Vol.~\bibinfo{volume}{34}.
  \bibinfo{pages}{1725--1732}.
\newblock
\urldef\tempurl%
\url{https://ojs.aaai.org//index.php/AAAI/article/view/5536}
\showURL{%
\tempurl}


\bibitem[\protect\citeauthoryear{Liu, Yuan, Fu, Jiang, Hayashi, and Neubig}{Liu
  et~al\mbox{.}}{2021d}]%
        {pretrain_prompt_survery}
\bibfield{author}{\bibinfo{person}{Pengfei Liu}, \bibinfo{person}{Weizhe Yuan},
  \bibinfo{person}{Jinlan Fu}, \bibinfo{person}{Zhengbao Jiang},
  \bibinfo{person}{Hiroaki Hayashi}, {and} \bibinfo{person}{Graham Neubig}.}
  \bibinfo{year}{2021}\natexlab{d}.
\newblock \showarticletitle{Pre-train, Prompt, and Predict: {A} Systematic
  Survey of Prompting Methods in Natural Language Processing}.
\newblock \bibinfo{journal}{\emph{CoRR}}  \bibinfo{volume}{abs/2107.13586}
  (\bibinfo{year}{2021}).
\newblock
\showeprint[arXiv]{2107.13586}
\urldef\tempurl%
\url{https://arxiv.org/abs/2107.13586}
\showURL{%
\tempurl}


\bibitem[\protect\citeauthoryear{Liu, Jia, Wei, Xu, Wang, and Vosoughi}{Liu
  et~al\mbox{.}}{2021a}]%
        {liu2021mitigating}
\bibfield{author}{\bibinfo{person}{Ruibo Liu}, \bibinfo{person}{Chenyan Jia},
  \bibinfo{person}{Jason Wei}, \bibinfo{person}{Guangxuan Xu},
  \bibinfo{person}{Lili Wang}, {and} \bibinfo{person}{Soroush Vosoughi}.}
  \bibinfo{year}{2021}\natexlab{a}.
\newblock \showarticletitle{Mitigating political bias in language models
  through reinforced calibration}. In \bibinfo{booktitle}{\emph{Proceedings of
  the AAAI Conference on Artificial Intelligence}}, Vol.~\bibinfo{volume}{35}.
  \bibinfo{pages}{14857--14866}.
\newblock
\urldef\tempurl%
\url{https://arxiv.org/abs/2104.14795}
\showURL{%
\tempurl}


\bibitem[\protect\citeauthoryear{Liu, Xu, Jia, Ma, Wang, and Vosoughi}{Liu
  et~al\mbox{.}}{2020c}]%
        {data_boost}
\bibfield{author}{\bibinfo{person}{Ruibo Liu}, \bibinfo{person}{Guangxuan Xu},
  \bibinfo{person}{Chenyan Jia}, \bibinfo{person}{Weicheng Ma},
  \bibinfo{person}{Lili Wang}, {and} \bibinfo{person}{Soroush Vosoughi}.}
  \bibinfo{year}{2020}\natexlab{c}.
\newblock \showarticletitle{Data Boost: Text Data Augmentation Through
  Reinforcement Learning Guided Conditional Generation}. In
  \bibinfo{booktitle}{\emph{Proceedings of the 2020 Conference on Empirical
  Methods in Natural Language Processing, {EMNLP} 2020, Online, November 16-20,
  2020}}, \bibfield{editor}{\bibinfo{person}{Bonnie Webber},
  \bibinfo{person}{Trevor Cohn}, \bibinfo{person}{Yulan He}, {and}
  \bibinfo{person}{Yang Liu}} (Eds.). \bibinfo{publisher}{Association for
  Computational Linguistics}, \bibinfo{pages}{9031--9041}.
\newblock
\urldef\tempurl%
\url{https://doi.org/10.18653/v1/2020.emnlp-main.726}
\showDOI{\tempurl}


\bibitem[\protect\citeauthoryear{Liu, Gu, Goyal, Li, Edunov, Ghazvininejad,
  Lewis, and Zettlemoyer}{Liu et~al\mbox{.}}{2020a}]%
        {mbart}
\bibfield{author}{\bibinfo{person}{Yinhan Liu}, \bibinfo{person}{Jiatao Gu},
  \bibinfo{person}{Naman Goyal}, \bibinfo{person}{Xian Li},
  \bibinfo{person}{Sergey Edunov}, \bibinfo{person}{Marjan Ghazvininejad},
  \bibinfo{person}{Mike Lewis}, {and} \bibinfo{person}{Luke Zettlemoyer}.}
  \bibinfo{year}{2020}\natexlab{a}.
\newblock \showarticletitle{Multilingual Denoising Pre-training for Neural
  Machine Translation}.
\newblock \bibinfo{journal}{\emph{Transactions of the Association for
  Computational Linguistics}}  \bibinfo{volume}{8} (\bibinfo{year}{2020}),
  \bibinfo{pages}{726--742}.
\newblock
\urldef\tempurl%
\url{https://doi.org/10.1162/tacl_a_00343}
\showDOI{\tempurl}


\bibitem[\protect\citeauthoryear{Liu, Maier, Minker, and Ultes}{Liu
  et~al\mbox{.}}{2021b}]%
        {liu2021naturalness}
\bibfield{author}{\bibinfo{person}{Ye Liu}, \bibinfo{person}{Wolfgang Maier},
  \bibinfo{person}{Wolfgang Minker}, {and} \bibinfo{person}{Stefan Ultes}.}
  \bibinfo{year}{2021}\natexlab{b}.
\newblock \showarticletitle{Naturalness Evaluation of Natural Language
  Generation in Task-oriented Dialogues using BERT}.
\newblock \bibinfo{journal}{\emph{arXiv preprint arXiv:2109.02938}}
  (\bibinfo{year}{2021}).
\newblock


\bibitem[\protect\citeauthoryear{Liu, Ott, Goyal, Du, Joshi, Chen, Levy, Lewis,
  Zettlemoyer, and Stoyanov}{Liu et~al\mbox{.}}{2019}]%
        {roberta}
\bibfield{author}{\bibinfo{person}{Yinhan Liu}, \bibinfo{person}{Myle Ott},
  \bibinfo{person}{Naman Goyal}, \bibinfo{person}{Jingfei Du},
  \bibinfo{person}{Mandar Joshi}, \bibinfo{person}{Danqi Chen},
  \bibinfo{person}{Omer Levy}, \bibinfo{person}{Mike Lewis},
  \bibinfo{person}{Luke Zettlemoyer}, {and} \bibinfo{person}{Veselin
  Stoyanov}.} \bibinfo{year}{2019}\natexlab{}.
\newblock \showarticletitle{RoBERTa: {A} Robustly Optimized {BERT} Pretraining
  Approach}.
\newblock \bibinfo{journal}{\emph{CoRR}}  \bibinfo{volume}{abs/1907.11692}
  (\bibinfo{year}{2019}).
\newblock
\showeprint[arxiv]{1907.11692}
\urldef\tempurl%
\url{http://arxiv.org/abs/1907.11692}
\showURL{%
\tempurl}


\bibitem[\protect\citeauthoryear{Logeswaran, Lee, and Bengio}{Logeswaran
  et~al\mbox{.}}{2018}]%
        {logeswaran2018content}
\bibfield{author}{\bibinfo{person}{Lajanugen Logeswaran},
  \bibinfo{person}{Honglak Lee}, {and} \bibinfo{person}{Samy Bengio}.}
  \bibinfo{year}{2018}\natexlab{}.
\newblock \showarticletitle{Content preserving text generation with attribute
  controls}. In \bibinfo{booktitle}{\emph{Advances in Neural Information
  Processing Systems}}, \bibfield{editor}{\bibinfo{person}{S.~Bengio},
  \bibinfo{person}{H.~Wallach}, \bibinfo{person}{H.~Larochelle},
  \bibinfo{person}{K.~Grauman}, \bibinfo{person}{N.~Cesa-Bianchi}, {and}
  \bibinfo{person}{R.~Garnett}} (Eds.), Vol.~\bibinfo{volume}{31}.
  \bibinfo{publisher}{Curran Associates, Inc.}
\newblock
\urldef\tempurl%
\url{https://proceedings.neurips.cc/paper/2018/file/7cf64379eb6f29a4d25c4b6a2df713e4-Paper.pdf}
\showURL{%
\tempurl}


\bibitem[\protect\citeauthoryear{Lowe, Noseworthy, Serban, Angelard-Gontier,
  Bengio, and Pineau}{Lowe et~al\mbox{.}}{2017}]%
        {lowe2017towards}
\bibfield{author}{\bibinfo{person}{Ryan Lowe}, \bibinfo{person}{Michael
  Noseworthy}, \bibinfo{person}{Iulian~Vlad Serban}, \bibinfo{person}{Nicolas
  Angelard-Gontier}, \bibinfo{person}{Yoshua Bengio}, {and}
  \bibinfo{person}{Joelle Pineau}.} \bibinfo{year}{2017}\natexlab{}.
\newblock \showarticletitle{Towards an Automatic {T}uring Test: Learning to
  Evaluate Dialogue Responses}. In \bibinfo{booktitle}{\emph{Proceedings of the
  55th Annual Meeting of the Association for Computational Linguistics (Volume
  1: Long Papers)}}. \bibinfo{publisher}{Association for Computational
  Linguistics}, \bibinfo{address}{Vancouver, Canada},
  \bibinfo{pages}{1116--1126}.
\newblock
\urldef\tempurl%
\url{https://doi.org/10.18653/v1/P17-1103}
\showDOI{\tempurl}


\bibitem[\protect\citeauthoryear{Luo, Dai, Yang, Liu, Chang, Sui, and Sun}{Luo
  et~al\mbox{.}}{2019}]%
        {luo2019learning}
\bibfield{author}{\bibinfo{person}{Fuli Luo}, \bibinfo{person}{Damai Dai},
  \bibinfo{person}{Pengcheng Yang}, \bibinfo{person}{Tianyu Liu},
  \bibinfo{person}{Baobao Chang}, \bibinfo{person}{Zhifang Sui}, {and}
  \bibinfo{person}{Xu Sun}.} \bibinfo{year}{2019}\natexlab{}.
\newblock \showarticletitle{Learning to Control the Fine-grained Sentiment for
  Story Ending Generation}. In \bibinfo{booktitle}{\emph{Proceedings of the
  57th Annual Meeting of the Association for Computational Linguistics}}.
  \bibinfo{publisher}{Association for Computational Linguistics},
  \bibinfo{address}{Florence, Italy}, \bibinfo{pages}{6020--6026}.
\newblock
\urldef\tempurl%
\url{https://doi.org/10.18653/v1/P19-1603}
\showDOI{\tempurl}


\bibitem[\protect\citeauthoryear{Malandrakis, Shen, Goyal, Gao, Sethi, and
  Metallinou}{Malandrakis et~al\mbox{.}}{2019}]%
        {malandrakis2019controlled}
\bibfield{author}{\bibinfo{person}{Nikolaos Malandrakis},
  \bibinfo{person}{Minmin Shen}, \bibinfo{person}{Anuj Goyal},
  \bibinfo{person}{Shuyang Gao}, \bibinfo{person}{Abhishek Sethi}, {and}
  \bibinfo{person}{Angeliki Metallinou}.} \bibinfo{year}{2019}\natexlab{}.
\newblock \showarticletitle{Controlled Text Generation for Data Augmentation in
  Intelligent Artificial Agents}. In \bibinfo{booktitle}{\emph{Proceedings of
  the 3rd Workshop on Neural Generation and Translation}}.
  \bibinfo{publisher}{Association for Computational Linguistics},
  \bibinfo{address}{Hong Kong}, \bibinfo{pages}{90--98}.
\newblock
\urldef\tempurl%
\url{https://doi.org/10.18653/v1/D19-5609}
\showDOI{\tempurl}


\bibitem[\protect\citeauthoryear{Mathur, Baldwin, and Cohn}{Mathur
  et~al\mbox{.}}{2019}]%
        {mathur-etal-2019-putting}
\bibfield{author}{\bibinfo{person}{Nitika Mathur}, \bibinfo{person}{Timothy
  Baldwin}, {and} \bibinfo{person}{Trevor Cohn}.}
  \bibinfo{year}{2019}\natexlab{}.
\newblock \showarticletitle{Putting Evaluation in Context: Contextual
  Embeddings Improve Machine Translation Evaluation}. In
  \bibinfo{booktitle}{\emph{Proceedings of the 57th Annual Meeting of the
  Association for Computational Linguistics}}. \bibinfo{publisher}{Association
  for Computational Linguistics}, \bibinfo{address}{Florence, Italy},
  \bibinfo{pages}{2799--2808}.
\newblock
\urldef\tempurl%
\url{https://doi.org/10.18653/v1/P19-1269}
\showDOI{\tempurl}


\bibitem[\protect\citeauthoryear{Mikolov, Chen, Corrado, and Dean}{Mikolov
  et~al\mbox{.}}{2013}]%
        {word2vector}
\bibfield{author}{\bibinfo{person}{Tomas Mikolov}, \bibinfo{person}{Kai Chen},
  \bibinfo{person}{Greg Corrado}, {and} \bibinfo{person}{Jeffrey Dean}.}
  \bibinfo{year}{2013}\natexlab{}.
\newblock \showarticletitle{Efficient estimation of word representations in
  vector space}.
\newblock \bibinfo{journal}{\emph{arXiv preprint arXiv:1301.3781}}
  (\bibinfo{year}{2013}).
\newblock


\bibitem[\protect\citeauthoryear{Mireshghallah, Goyal, and
  Berg-Kirkpatrick}{Mireshghallah et~al\mbox{.}}{2022}]%
        {mireshghallah-etal-2022-mix}
\bibfield{author}{\bibinfo{person}{Fatemehsadat Mireshghallah},
  \bibinfo{person}{Kartik Goyal}, {and} \bibinfo{person}{Taylor
  Berg-Kirkpatrick}.} \bibinfo{year}{2022}\natexlab{}.
\newblock \showarticletitle{Mix and Match: Learning-free Controllable Text
  Generationusing Energy Language Models}. In
  \bibinfo{booktitle}{\emph{Proceedings of the 60th Annual Meeting of the
  Association for Computational Linguistics (Volume 1: Long Papers)}}.
  \bibinfo{publisher}{Association for Computational Linguistics},
  \bibinfo{address}{Dublin, Ireland}, \bibinfo{pages}{401--415}.
\newblock
\urldef\tempurl%
\url{https://doi.org/10.18653/v1/2022.acl-long.31}
\showDOI{\tempurl}


\bibitem[\protect\citeauthoryear{Mitra, Hutto, and Gilbert}{Mitra
  et~al\mbox{.}}{2015}]%
        {mitra2015comparing}
\bibfield{author}{\bibinfo{person}{Tanushree Mitra}, \bibinfo{person}{C.J.
  Hutto}, {and} \bibinfo{person}{Eric Gilbert}.}
  \bibinfo{year}{2015}\natexlab{}.
\newblock \showarticletitle{Comparing Person- and Process-Centric Strategies
  for Obtaining Quality Data on Amazon Mechanical Turk}. In
  \bibinfo{booktitle}{\emph{Proceedings of the 33rd Annual ACM Conference on
  Human Factors in Computing Systems}} (Seoul, Republic of Korea)
  \emph{(\bibinfo{series}{CHI '15})}. \bibinfo{publisher}{Association for
  Computing Machinery}, \bibinfo{address}{New York, NY, USA},
  \bibinfo{pages}{1345–1354}.
\newblock
\showISBNx{9781450331456}
\urldef\tempurl%
\url{https://doi.org/10.1145/2702123.2702553}
\showDOI{\tempurl}


\bibitem[\protect\citeauthoryear{Nan, Radev, Zhang, Rau, Sivaprasad, Hsieh,
  Tang, Vyas, Verma, Krishna, Liu, Irwanto, Pan, Rahman, Zaidi, Mutuma,
  Tarabar, Gupta, Yu, Tan, Lin, Xiong, Socher, and Rajani}{Nan
  et~al\mbox{.}}{2021}]%
        {dart}
\bibfield{author}{\bibinfo{person}{Linyong Nan}, \bibinfo{person}{Dragomir
  Radev}, \bibinfo{person}{Rui Zhang}, \bibinfo{person}{Amrit Rau},
  \bibinfo{person}{Abhinand Sivaprasad}, \bibinfo{person}{Chiachun Hsieh},
  \bibinfo{person}{Xiangru Tang}, \bibinfo{person}{Aadit Vyas},
  \bibinfo{person}{Neha Verma}, \bibinfo{person}{Pranav Krishna},
  \bibinfo{person}{Yangxiaokang Liu}, \bibinfo{person}{Nadia Irwanto},
  \bibinfo{person}{Jessica Pan}, \bibinfo{person}{Faiaz Rahman},
  \bibinfo{person}{Ahmad Zaidi}, \bibinfo{person}{Mutethia Mutuma},
  \bibinfo{person}{Yasin Tarabar}, \bibinfo{person}{Ankit Gupta},
  \bibinfo{person}{Tao Yu}, \bibinfo{person}{Yi~Chern Tan},
  \bibinfo{person}{Xi~Victoria Lin}, \bibinfo{person}{Caiming Xiong},
  \bibinfo{person}{Richard Socher}, {and} \bibinfo{person}{Nazneen~Fatema
  Rajani}.} \bibinfo{year}{2021}\natexlab{}.
\newblock \showarticletitle{{DART}: Open-Domain Structured Data Record to Text
  Generation}. In \bibinfo{booktitle}{\emph{Proceedings of the 2021 Conference
  of the North American Chapter of the Association for Computational
  Linguistics: Human Language Technologies}}. \bibinfo{publisher}{Association
  for Computational Linguistics}, \bibinfo{address}{Online},
  \bibinfo{pages}{432--447}.
\newblock
\urldef\tempurl%
\url{https://doi.org/10.18653/v1/2021.naacl-main.37}
\showDOI{\tempurl}


\bibitem[\protect\citeauthoryear{Niu and Bansal}{Niu and Bansal}{2018}]%
        {niu2018polite}
\bibfield{author}{\bibinfo{person}{Tong Niu} {and} \bibinfo{person}{Mohit
  Bansal}.} \bibinfo{year}{2018}\natexlab{}.
\newblock \showarticletitle{Polite Dialogue Generation Without Parallel Data}.
\newblock \bibinfo{journal}{\emph{Transactions of the Association for
  Computational Linguistics}}  \bibinfo{volume}{6} (\bibinfo{year}{2018}),
  \bibinfo{pages}{373--389}.
\newblock
\showISSN{2307-387X}
\urldef\tempurl%
\url{https://transacl.org/ojs/index.php/tacl/article/view/1424}
\showURL{%
\tempurl}


\bibitem[\protect\citeauthoryear{Novikova, Du{\v{s}}ek, and Rieser}{Novikova
  et~al\mbox{.}}{2018}]%
        {novikova2018rankme}
\bibfield{author}{\bibinfo{person}{Jekaterina Novikova},
  \bibinfo{person}{Ond{\v{r}}ej Du{\v{s}}ek}, {and} \bibinfo{person}{Verena
  Rieser}.} \bibinfo{year}{2018}\natexlab{}.
\newblock \showarticletitle{{R}ank{ME}: Reliable Human Ratings for Natural
  Language Generation}. In \bibinfo{booktitle}{\emph{Proceedings of the 2018
  Conference of the North {A}merican Chapter of the Association for
  Computational Linguistics: Human Language Technologies, Volume 2 (Short
  Papers)}}. \bibinfo{publisher}{Association for Computational Linguistics},
  \bibinfo{address}{New Orleans, Louisiana}, \bibinfo{pages}{72--78}.
\newblock
\urldef\tempurl%
\url{https://doi.org/10.18653/v1/N18-2012}
\showDOI{\tempurl}


\bibitem[\protect\citeauthoryear{Ouyang, Wu, Jiang, Almeida, Wainwright,
  Mishkin, Zhang, Agarwal, Slama, Ray, Schulman, Hilton, Kelton, Miller,
  Simens, Askell, Welinder, Christiano, Leike, and Lowe}{Ouyang
  et~al\mbox{.}}{2022}]%
        {instructgpt}
\bibfield{author}{\bibinfo{person}{Long Ouyang}, \bibinfo{person}{Jeffrey Wu},
  \bibinfo{person}{Xu Jiang}, \bibinfo{person}{Diogo Almeida},
  \bibinfo{person}{Carroll Wainwright}, \bibinfo{person}{Pamela Mishkin},
  \bibinfo{person}{Chong Zhang}, \bibinfo{person}{Sandhini Agarwal},
  \bibinfo{person}{Katarina Slama}, \bibinfo{person}{Alex Ray},
  \bibinfo{person}{John Schulman}, \bibinfo{person}{Jacob Hilton},
  \bibinfo{person}{Fraser Kelton}, \bibinfo{person}{Luke Miller},
  \bibinfo{person}{Maddie Simens}, \bibinfo{person}{Amanda Askell},
  \bibinfo{person}{Peter Welinder}, \bibinfo{person}{Paul~F Christiano},
  \bibinfo{person}{Jan Leike}, {and} \bibinfo{person}{Ryan Lowe}.}
  \bibinfo{year}{2022}\natexlab{}.
\newblock \showarticletitle{Training language models to follow instructions
  with human feedback}. In \bibinfo{booktitle}{\emph{Advances in Neural
  Information Processing Systems}},
  \bibfield{editor}{\bibinfo{person}{S.~Koyejo}, \bibinfo{person}{S.~Mohamed},
  \bibinfo{person}{A.~Agarwal}, \bibinfo{person}{D.~Belgrave},
  \bibinfo{person}{K.~Cho}, {and} \bibinfo{person}{A.~Oh}} (Eds.),
  Vol.~\bibinfo{volume}{35}. \bibinfo{publisher}{Curran Associates, Inc.},
  \bibinfo{pages}{27730--27744}.
\newblock
\urldef\tempurl%
\url{https://proceedings.neurips.cc/paper_files/paper/2022/file/b1efde53be364a73914f58805a001731-Paper-Conference.pdf}
\showURL{%
\tempurl}


\bibitem[\protect\citeauthoryear{Papineni, Roukos, Ward, and Zhu}{Papineni
  et~al\mbox{.}}{2002}]%
        {papineni2002bleu}
\bibfield{author}{\bibinfo{person}{Kishore Papineni}, \bibinfo{person}{Salim
  Roukos}, \bibinfo{person}{Todd Ward}, {and} \bibinfo{person}{Wei-Jing Zhu}.}
  \bibinfo{year}{2002}\natexlab{}.
\newblock \showarticletitle{{B}leu: a Method for Automatic Evaluation of
  Machine Translation}. In \bibinfo{booktitle}{\emph{Proceedings of the 40th
  Annual Meeting of the Association for Computational Linguistics}}.
  \bibinfo{publisher}{Association for Computational Linguistics},
  \bibinfo{address}{Philadelphia, Pennsylvania, USA},
  \bibinfo{pages}{311--318}.
\newblock
\urldef\tempurl%
\url{https://doi.org/10.3115/1073083.1073135}
\showDOI{\tempurl}


\bibitem[\protect\citeauthoryear{Pascual, Egressy, Meister, Cotterell, and
  Wattenhofer}{Pascual et~al\mbox{.}}{2021}]%
        {word_simility_emnlp2021_findings}
\bibfield{author}{\bibinfo{person}{Damian Pascual}, \bibinfo{person}{Beni
  Egressy}, \bibinfo{person}{Clara Meister}, \bibinfo{person}{Ryan Cotterell},
  {and} \bibinfo{person}{Roger Wattenhofer}.} \bibinfo{year}{2021}\natexlab{}.
\newblock \showarticletitle{A Plug-and-Play Method for Controlled Text
  Generation}. In \bibinfo{booktitle}{\emph{Findings of the Association for
  Computational Linguistics: {EMNLP} 2021, Virtual Event / Punta Cana,
  Dominican Republic, 16-20 November, 2021}},
  \bibfield{editor}{\bibinfo{person}{Marie{-}Francine Moens},
  \bibinfo{person}{Xuanjing Huang}, \bibinfo{person}{Lucia Specia}, {and}
  \bibinfo{person}{Scott~Wen{-}tau Yih}} (Eds.).
  \bibinfo{publisher}{Association for Computational Linguistics},
  \bibinfo{pages}{3973--3997}.
\newblock
\urldef\tempurl%
\url{https://aclanthology.org/2021.findings-emnlp.334}
\showURL{%
\tempurl}


\bibitem[\protect\citeauthoryear{Peng, Zhu, Li, Li, Li, Zeng, and Gao}{Peng
  et~al\mbox{.}}{2020}]%
        {peng-etal-2020-shot}
\bibfield{author}{\bibinfo{person}{Baolin Peng}, \bibinfo{person}{Chenguang
  Zhu}, \bibinfo{person}{Chunyuan Li}, \bibinfo{person}{Xiujun Li},
  \bibinfo{person}{Jinchao Li}, \bibinfo{person}{Michael Zeng}, {and}
  \bibinfo{person}{Jianfeng Gao}.} \bibinfo{year}{2020}\natexlab{}.
\newblock \showarticletitle{Few-shot Natural Language Generation for
  Task-Oriented Dialog}. In \bibinfo{booktitle}{\emph{Findings of the
  Association for Computational Linguistics: EMNLP 2020}}.
  \bibinfo{publisher}{Association for Computational Linguistics},
  \bibinfo{address}{Online}, \bibinfo{pages}{172--182}.
\newblock
\urldef\tempurl%
\url{https://doi.org/10.18653/v1/2020.findings-emnlp.17}
\showDOI{\tempurl}


\bibitem[\protect\citeauthoryear{Peters, Neumann, Iyyer, Gardner, Clark, Lee,
  and Zettlemoyer}{Peters et~al\mbox{.}}{2018}]%
        {ELMo}
\bibfield{author}{\bibinfo{person}{Matthew~E. Peters}, \bibinfo{person}{Mark
  Neumann}, \bibinfo{person}{Mohit Iyyer}, \bibinfo{person}{Matt Gardner},
  \bibinfo{person}{Christopher Clark}, \bibinfo{person}{Kenton Lee}, {and}
  \bibinfo{person}{Luke Zettlemoyer}.} \bibinfo{year}{2018}\natexlab{}.
\newblock \showarticletitle{Deep Contextualized Word Representations}. In
  \bibinfo{booktitle}{\emph{Proceedings of the 2018 Conference of the North
  {A}merican Chapter of the Association for Computational Linguistics: Human
  Language Technologies, Volume 1 (Long Papers)}}.
  \bibinfo{publisher}{Association for Computational Linguistics},
  \bibinfo{address}{New Orleans, Louisiana}, \bibinfo{pages}{2227--2237}.
\newblock
\urldef\tempurl%
\url{https://doi.org/10.18653/v1/N18-1202}
\showDOI{\tempurl}


\bibitem[\protect\citeauthoryear{Pillutla, Swayamdipta, Zellers, Thickstun,
  Choi, and Harchaoui}{Pillutla et~al\mbox{.}}{2021}]%
        {MAUVE_human_machine}
\bibfield{author}{\bibinfo{person}{Krishna Pillutla}, \bibinfo{person}{Swabha
  Swayamdipta}, \bibinfo{person}{Rowan Zellers}, \bibinfo{person}{John
  Thickstun}, \bibinfo{person}{Yejin Choi}, {and} \bibinfo{person}{Za{\"{\i}}d
  Harchaoui}.} \bibinfo{year}{2021}\natexlab{}.
\newblock \showarticletitle{{MAUVE:} Human-Machine Divergence Curves for
  Evaluating Open-Ended Text Generation}.
\newblock \bibinfo{journal}{\emph{CoRR}}  \bibinfo{volume}{abs/2102.01454}
  (\bibinfo{year}{2021}).
\newblock
\showeprint[arXiv]{2102.01454}
\urldef\tempurl%
\url{https://arxiv.org/abs/2102.01454}
\showURL{%
\tempurl}


\bibitem[\protect\citeauthoryear{Prabhumoye, Black, and
  Salakhutdinov}{Prabhumoye et~al\mbox{.}}{2020}]%
        {survey_ctg}
\bibfield{author}{\bibinfo{person}{Shrimai Prabhumoye}, \bibinfo{person}{Alan~W
  Black}, {and} \bibinfo{person}{Ruslan Salakhutdinov}.}
  \bibinfo{year}{2020}\natexlab{}.
\newblock \showarticletitle{Exploring Controllable Text Generation Techniques}.
  In \bibinfo{booktitle}{\emph{Proceedings of the 28th International Conference
  on Computational Linguistics}}. \bibinfo{publisher}{International Committee
  on Computational Linguistics}, \bibinfo{address}{Barcelona, Spain (Online)},
  \bibinfo{pages}{1--14}.
\newblock
\urldef\tempurl%
\url{https://doi.org/10.18653/v1/2020.coling-main.1}
\showDOI{\tempurl}


\bibitem[\protect\citeauthoryear{Pryzant, Martinez, Dass, Kurohashi, Jurafsky,
  and Yang}{Pryzant et~al\mbox{.}}{2020}]%
        {pryzant2020automatically}
\bibfield{author}{\bibinfo{person}{Reid Pryzant},
  \bibinfo{person}{Richard~Diehl Martinez}, \bibinfo{person}{Nathan Dass},
  \bibinfo{person}{Sadao Kurohashi}, \bibinfo{person}{Dan Jurafsky}, {and}
  \bibinfo{person}{Diyi Yang}.} \bibinfo{year}{2020}\natexlab{}.
\newblock \showarticletitle{Automatically neutralizing subjective bias in
  text}. In \bibinfo{booktitle}{\emph{Proceedings of the AAAI conference on
  artificial intelligence}}, Vol.~\bibinfo{volume}{34}.
  \bibinfo{pages}{480--489}.
\newblock


\bibitem[\protect\citeauthoryear{Puduppully, Dong, and Lapata}{Puduppully
  et~al\mbox{.}}{2019}]%
        {puduppully2019data}
\bibfield{author}{\bibinfo{person}{Ratish Puduppully}, \bibinfo{person}{Li
  Dong}, {and} \bibinfo{person}{Mirella Lapata}.}
  \bibinfo{year}{2019}\natexlab{}.
\newblock \showarticletitle{Data-to-text generation with content selection and
  planning}. In \bibinfo{booktitle}{\emph{Proceedings of the AAAI conference on
  artificial intelligence}}, Vol.~\bibinfo{volume}{33}.
  \bibinfo{pages}{6908--6915}.
\newblock


\bibitem[\protect\citeauthoryear{Qian, Dong, Shen, Wei, and Chen}{Qian
  et~al\mbox{.}}{2022}]%
        {qian-etal-2022-controllable}
\bibfield{author}{\bibinfo{person}{Jing Qian}, \bibinfo{person}{Li Dong},
  \bibinfo{person}{Yelong Shen}, \bibinfo{person}{Furu Wei}, {and}
  \bibinfo{person}{Weizhu Chen}.} \bibinfo{year}{2022}\natexlab{}.
\newblock \showarticletitle{Controllable Natural Language Generation with
  Contrastive Prefixes}. In \bibinfo{booktitle}{\emph{Findings of the
  Association for Computational Linguistics: ACL 2022}}.
  \bibinfo{publisher}{Association for Computational Linguistics},
  \bibinfo{address}{Dublin, Ireland}, \bibinfo{pages}{2912--2924}.
\newblock
\urldef\tempurl%
\url{https://doi.org/10.18653/v1/2022.findings-acl.229}
\showDOI{\tempurl}


\bibitem[\protect\citeauthoryear{Qian, Muaz, Zhang, and Hyun}{Qian
  et~al\mbox{.}}{2019}]%
        {qian2019reducing}
\bibfield{author}{\bibinfo{person}{Yusu Qian}, \bibinfo{person}{Urwa Muaz},
  \bibinfo{person}{Ben Zhang}, {and} \bibinfo{person}{Jae~Won Hyun}.}
  \bibinfo{year}{2019}\natexlab{}.
\newblock \showarticletitle{Reducing gender bias in word-level language models
  with a gender-equalizing loss function}.
\newblock \bibinfo{journal}{\emph{arXiv preprint arXiv:1905.12801}}
  (\bibinfo{year}{2019}).
\newblock


\bibitem[\protect\citeauthoryear{Qin, Galley, Brockett, Liu, Gao, Dolan, Choi,
  and Gao}{Qin et~al\mbox{.}}{2019}]%
        {qin2019conversing}
\bibfield{author}{\bibinfo{person}{Lianhui Qin}, \bibinfo{person}{Michel
  Galley}, \bibinfo{person}{Chris Brockett}, \bibinfo{person}{Xiaodong Liu},
  \bibinfo{person}{Xiang Gao}, \bibinfo{person}{Bill Dolan},
  \bibinfo{person}{Yejin Choi}, {and} \bibinfo{person}{Jianfeng Gao}.}
  \bibinfo{year}{2019}\natexlab{}.
\newblock \showarticletitle{Conversing by Reading: Contentful Neural
  Conversation with On-demand Machine Reading}. In
  \bibinfo{booktitle}{\emph{Proceedings of the 57th Conference of the
  Association for Computational Linguistics, ACL 2019, Florence, Italy, July
  28- August 2, 2019, Volume 1: Long Papers}},
  \bibfield{editor}{\bibinfo{person}{Anna Korhonen}, \bibinfo{person}{David~R.
  Traum}, {and} \bibinfo{person}{Lluís Màrquez}} (Eds.).
  \bibinfo{publisher}{Association for Computational Linguistics},
  \bibinfo{pages}{5427--5436}.
\newblock
\showISBNx{978-1-950737-48-2}
\urldef\tempurl%
\url{https://www.aclweb.org/anthology/P19-1539/}
\showURL{%
\tempurl}


\bibitem[\protect\citeauthoryear{Qin, Welleck, Khashabi, and Choi}{Qin
  et~al\mbox{.}}{2022}]%
        {COLD_Decoding}
\bibfield{author}{\bibinfo{person}{Lianhui Qin}, \bibinfo{person}{Sean
  Welleck}, \bibinfo{person}{Daniel Khashabi}, {and} \bibinfo{person}{Yejin
  Choi}.} \bibinfo{year}{2022}\natexlab{}.
\newblock \bibinfo{title}{COLD Decoding: Energy-based Constrained Text
  Generation with Langevin Dynamics}.
\newblock
\newblock
\urldef\tempurl%
\url{https://doi.org/10.48550/ARXIV.2202.11705}
\showDOI{\tempurl}


\bibitem[\protect\citeauthoryear{Radford, Wu, Child, Luan, Amodei, Sutskever,
  et~al\mbox{.}}{Radford et~al\mbox{.}}{2019}]%
        {gpt2}
\bibfield{author}{\bibinfo{person}{Alec Radford}, \bibinfo{person}{Jeffrey Wu},
  \bibinfo{person}{Rewon Child}, \bibinfo{person}{David Luan},
  \bibinfo{person}{Dario Amodei}, \bibinfo{person}{Ilya Sutskever},
  {et~al\mbox{.}}} \bibinfo{year}{2019}\natexlab{}.
\newblock \showarticletitle{Language models are unsupervised multitask
  learners}.
\newblock \bibinfo{journal}{\emph{OpenAI blog}} \bibinfo{volume}{1},
  \bibinfo{number}{8} (\bibinfo{year}{2019}), \bibinfo{pages}{9}.
\newblock


\bibitem[\protect\citeauthoryear{Raffel, Shazeer, Roberts, Lee, Narang, Matena,
  Zhou, Li, and Liu}{Raffel et~al\mbox{.}}{2020}]%
        {t5}
\bibfield{author}{\bibinfo{person}{Colin Raffel}, \bibinfo{person}{Noam
  Shazeer}, \bibinfo{person}{Adam Roberts}, \bibinfo{person}{Katherine Lee},
  \bibinfo{person}{Sharan Narang}, \bibinfo{person}{Michael Matena},
  \bibinfo{person}{Yanqi Zhou}, \bibinfo{person}{Wei Li}, {and}
  \bibinfo{person}{Peter~J. Liu}.} \bibinfo{year}{2020}\natexlab{}.
\newblock \showarticletitle{Exploring the Limits of Transfer Learning with a
  Unified Text-to-Text Transformer}.
\newblock \bibinfo{journal}{\emph{Journal of Machine Learning Research}}
  \bibinfo{volume}{21}, \bibinfo{number}{140} (\bibinfo{year}{2020}),
  \bibinfo{pages}{1--67}.
\newblock
\urldef\tempurl%
\url{http://jmlr.org/papers/v21/20-074.html}
\showURL{%
\tempurl}


\bibitem[\protect\citeauthoryear{Rashkin, Reitter, Tomar, and Das}{Rashkin
  et~al\mbox{.}}{2021}]%
        {rashkin-etal-2021-increasing}
\bibfield{author}{\bibinfo{person}{Hannah Rashkin}, \bibinfo{person}{David
  Reitter}, \bibinfo{person}{Gaurav~Singh Tomar}, {and}
  \bibinfo{person}{Dipanjan Das}.} \bibinfo{year}{2021}\natexlab{}.
\newblock \showarticletitle{Increasing Faithfulness in Knowledge-Grounded
  Dialogue with Controllable Features}. In
  \bibinfo{booktitle}{\emph{Proceedings of the 59th Annual Meeting of the
  Association for Computational Linguistics and the 11th International Joint
  Conference on Natural Language Processing (Volume 1: Long Papers)}}.
  \bibinfo{publisher}{Association for Computational Linguistics},
  \bibinfo{address}{Online}, \bibinfo{pages}{704--718}.
\newblock
\urldef\tempurl%
\url{https://doi.org/10.18653/v1/2021.acl-long.58}
\showDOI{\tempurl}


\bibitem[\protect\citeauthoryear{Reimers and Gurevych}{Reimers and
  Gurevych}{2019}]%
        {reimers2019sentence}
\bibfield{author}{\bibinfo{person}{Nils Reimers} {and} \bibinfo{person}{Iryna
  Gurevych}.} \bibinfo{year}{2019}\natexlab{}.
\newblock \showarticletitle{Sentence-{BERT}: Sentence Embeddings using
  {S}iamese {BERT}-Networks}. In \bibinfo{booktitle}{\emph{Proceedings of the
  2019 Conference on Empirical Methods in Natural Language Processing and the
  9th International Joint Conference on Natural Language Processing
  (EMNLP-IJCNLP)}}. \bibinfo{publisher}{Association for Computational
  Linguistics}, \bibinfo{address}{Hong Kong, China},
  \bibinfo{pages}{3982--3992}.
\newblock
\urldef\tempurl%
\url{https://doi.org/10.18653/v1/D19-1410}
\showDOI{\tempurl}


\bibitem[\protect\citeauthoryear{Ribeiro, Schmitt, Sch{\"u}tze, and
  Gurevych}{Ribeiro et~al\mbox{.}}{2021a}]%
        {plm_amr}
\bibfield{author}{\bibinfo{person}{Leonardo F.~R. Ribeiro},
  \bibinfo{person}{Martin Schmitt}, \bibinfo{person}{Hinrich Sch{\"u}tze},
  {and} \bibinfo{person}{Iryna Gurevych}.} \bibinfo{year}{2021}\natexlab{a}.
\newblock \showarticletitle{Investigating Pretrained Language Models for
  Graph-to-Text Generation}. In \bibinfo{booktitle}{\emph{Proceedings of the
  3rd Workshop on Natural Language Processing for Conversational AI}}.
  \bibinfo{publisher}{Association for Computational Linguistics},
  \bibinfo{address}{Online}, \bibinfo{pages}{211--227}.
\newblock
\urldef\tempurl%
\url{https://doi.org/10.18653/v1/2021.nlp4convai-1.20}
\showDOI{\tempurl}


\bibitem[\protect\citeauthoryear{Ribeiro, Zhang, and Gurevych}{Ribeiro
  et~al\mbox{.}}{2021b}]%
        {structAdapter}
\bibfield{author}{\bibinfo{person}{Leonardo F.~R. Ribeiro},
  \bibinfo{person}{Yue Zhang}, {and} \bibinfo{person}{Iryna Gurevych}.}
  \bibinfo{year}{2021}\natexlab{b}.
\newblock \showarticletitle{Structural Adapters in Pretrained Language Models
  for AMR-to-Text Generation}. In \bibinfo{booktitle}{\emph{Proceedings of the
  2021 Conference on Empirical Methods in Natural Language Processing, {EMNLP}
  2021, Virtual Event / Punta Cana, Dominican Republic, 7-11 November, 2021}},
  \bibfield{editor}{\bibinfo{person}{Marie{-}Francine Moens},
  \bibinfo{person}{Xuanjing Huang}, \bibinfo{person}{Lucia Specia}, {and}
  \bibinfo{person}{Scott~Wen{-}tau Yih}} (Eds.).
  \bibinfo{publisher}{Association for Computational Linguistics},
  \bibinfo{pages}{4269--4282}.
\newblock
\urldef\tempurl%
\url{https://aclanthology.org/2021.emnlp-main.351}
\showURL{%
\tempurl}


\bibitem[\protect\citeauthoryear{Ruan and Ling}{Ruan and Ling}{2021}]%
        {emo_vae}
\bibfield{author}{\bibinfo{person}{Yu-Ping Ruan} {and} \bibinfo{person}{Zhenhua
  Ling}.} \bibinfo{year}{2021}\natexlab{}.
\newblock \showarticletitle{Emotion-Regularized Conditional Variational
  Autoencoder for Emotional Response Generation}.
\newblock \bibinfo{journal}{\emph{IEEE Transactions on Affective Computing}}
  (\bibinfo{year}{2021}).
\newblock
\urldef\tempurl%
\url{https://arxiv.org/abs/2104.08857}
\showURL{%
\tempurl}


\bibitem[\protect\citeauthoryear{Saha, Singh, Kumar, Mathew, and
  Mukherjee}{Saha et~al\mbox{.}}{2022}]%
        {ijcai2022p716}
\bibfield{author}{\bibinfo{person}{Punyajoy Saha}, \bibinfo{person}{Kanishk
  Singh}, \bibinfo{person}{Adarsh Kumar}, \bibinfo{person}{Binny Mathew}, {and}
  \bibinfo{person}{Animesh Mukherjee}.} \bibinfo{year}{2022}\natexlab{}.
\newblock \showarticletitle{CounterGeDi: A Controllable Approach to Generate
  Polite, Detoxified and Emotional Counterspeech}. In
  \bibinfo{booktitle}{\emph{Proceedings of the Thirty-First International Joint
  Conference on Artificial Intelligence, {IJCAI-22}}},
  \bibfield{editor}{\bibinfo{person}{Lud~De Raedt}} (Ed.).
  \bibinfo{publisher}{International Joint Conferences on Artificial
  Intelligence Organization}, \bibinfo{pages}{5157--5163}.
\newblock
\urldef\tempurl%
\url{https://doi.org/10.24963/ijcai.2022/716}
\showDOI{\tempurl}
\newblock
\shownote{AI for Good.}


\bibitem[\protect\citeauthoryear{Samanta, Agarwal, and Ganguly}{Samanta
  et~al\mbox{.}}{2020}]%
        {samanta2020fine}
\bibfield{author}{\bibinfo{person}{Bidisha Samanta}, \bibinfo{person}{Mohit
  Agarwal}, {and} \bibinfo{person}{Niloy Ganguly}.}
  \bibinfo{year}{2020}\natexlab{}.
\newblock \showarticletitle{Fine-grained Sentiment Controlled Text Generation}.
\newblock \bibinfo{journal}{\emph{arXiv preprint arXiv:2006.09891}}
  (\bibinfo{year}{2020}).
\newblock
\urldef\tempurl%
\url{https://arxiv.org/abs/2006.09891}
\showURL{%
\tempurl}


\bibitem[\protect\citeauthoryear{Scao, Fan, Akiki, Pavlick, Ili{\'c}, Hesslow,
  Castagn{\'e}, Luccioni, Yvon, Gall{\'e}, et~al\mbox{.}}{Scao
  et~al\mbox{.}}{2022}]%
        {scao2022bloom}
\bibfield{author}{\bibinfo{person}{Teven~Le Scao}, \bibinfo{person}{Angela
  Fan}, \bibinfo{person}{Christopher Akiki}, \bibinfo{person}{Ellie Pavlick},
  \bibinfo{person}{Suzana Ili{\'c}}, \bibinfo{person}{Daniel Hesslow},
  \bibinfo{person}{Roman Castagn{\'e}}, \bibinfo{person}{Alexandra~Sasha
  Luccioni}, \bibinfo{person}{Fran{\c{c}}ois Yvon}, \bibinfo{person}{Matthias
  Gall{\'e}}, {et~al\mbox{.}}} \bibinfo{year}{2022}\natexlab{}.
\newblock \showarticletitle{Bloom: A 176b-parameter open-access multilingual
  language model}.
\newblock \bibinfo{journal}{\emph{arXiv preprint arXiv:2211.05100}}
  (\bibinfo{year}{2022}).
\newblock


\bibitem[\protect\citeauthoryear{Scialom, Dray, Lamprier, Piwowarski, and
  Staiano}{Scialom et~al\mbox{.}}{2020}]%
        {DAS}
\bibfield{author}{\bibinfo{person}{Thomas Scialom},
  \bibinfo{person}{Paul-Alexis Dray}, \bibinfo{person}{Sylvain Lamprier},
  \bibinfo{person}{Benjamin Piwowarski}, {and} \bibinfo{person}{Jacopo
  Staiano}.} \bibinfo{year}{2020}\natexlab{}.
\newblock \showarticletitle{Discriminative Adversarial Search for Abstractive
  Summarization}. In \bibinfo{booktitle}{\emph{Proceedings of the 37th
  International Conference on Machine Learning}}
  \emph{(\bibinfo{series}{ICML'20})}. \bibinfo{publisher}{JMLR.org}, Article
  \bibinfo{articleno}{793}, \bibinfo{numpages}{10}~pages.
\newblock
\urldef\tempurl%
\url{https://arxiv.org/abs/2002.10375}
\showURL{%
\tempurl}


\bibitem[\protect\citeauthoryear{Sellam, Das, and Parikh}{Sellam
  et~al\mbox{.}}{2020}]%
        {sellam2020bleurt}
\bibfield{author}{\bibinfo{person}{Thibault Sellam}, \bibinfo{person}{Dipanjan
  Das}, {and} \bibinfo{person}{Ankur Parikh}.} \bibinfo{year}{2020}\natexlab{}.
\newblock \showarticletitle{{BLEURT}: Learning Robust Metrics for Text
  Generation}. In \bibinfo{booktitle}{\emph{Proceedings of the 58th Annual
  Meeting of the Association for Computational Linguistics}}.
  \bibinfo{publisher}{Association for Computational Linguistics},
  \bibinfo{address}{Online}, \bibinfo{pages}{7881--7892}.
\newblock
\urldef\tempurl%
\url{https://doi.org/10.18653/v1/2020.acl-main.704}
\showDOI{\tempurl}


\bibitem[\protect\citeauthoryear{Sennrich, Haddow, and Birch}{Sennrich
  et~al\mbox{.}}{2016}]%
        {sennrich2016controlling}
\bibfield{author}{\bibinfo{person}{Rico Sennrich}, \bibinfo{person}{Barry
  Haddow}, {and} \bibinfo{person}{Alexandra Birch}.}
  \bibinfo{year}{2016}\natexlab{}.
\newblock \showarticletitle{Controlling Politeness in Neural Machine
  Translation via Side Constraints}. In \bibinfo{booktitle}{\emph{Proceedings
  of the 2016 Conference of the North {A}merican Chapter of the Association for
  Computational Linguistics: Human Language Technologies}}.
  \bibinfo{publisher}{Association for Computational Linguistics},
  \bibinfo{address}{San Diego, California}, \bibinfo{pages}{35--40}.
\newblock
\urldef\tempurl%
\url{https://doi.org/10.18653/v1/N16-1005}
\showDOI{\tempurl}


\bibitem[\protect\citeauthoryear{Shao, Shao, Wang, Wang, and Gao}{Shao
  et~al\mbox{.}}{2021}]%
        {sent_style_cikm_2021}
\bibfield{author}{\bibinfo{person}{Yizhan Shao}, \bibinfo{person}{Tong Shao},
  \bibinfo{person}{Minghao Wang}, \bibinfo{person}{Peng Wang}, {and}
  \bibinfo{person}{Jie Gao}.} \bibinfo{year}{2021}\natexlab{}.
\newblock \showarticletitle{A Sentiment and Style Controllable Approach for
  Chinese Poetry Generation}. In \bibinfo{booktitle}{\emph{{CIKM} '21: The 30th
  {ACM} International Conference on Information and Knowledge Management,
  Virtual Event, Queensland, Australia, November 1 - 5, 2021}},
  \bibfield{editor}{\bibinfo{person}{Gianluca Demartini},
  \bibinfo{person}{Guido Zuccon}, \bibinfo{person}{J.~Shane Culpepper},
  \bibinfo{person}{Zi~Huang}, {and} \bibinfo{person}{Hanghang Tong}} (Eds.).
  \bibinfo{publisher}{{ACM}}, \bibinfo{pages}{4784--4788}.
\newblock
\urldef\tempurl%
\url{https://doi.org/10.1145/3459637.3481964}
\showDOI{\tempurl}


\bibitem[\protect\citeauthoryear{Sheng, Chang, Natarajan, and Peng}{Sheng
  et~al\mbox{.}}{2020}]%
        {sheng2020towards}
\bibfield{author}{\bibinfo{person}{Emily Sheng}, \bibinfo{person}{Kai-Wei
  Chang}, \bibinfo{person}{Premkumar Natarajan}, {and} \bibinfo{person}{Nanyun
  Peng}.} \bibinfo{year}{2020}\natexlab{}.
\newblock \showarticletitle{Towards controllable biases in language
  generation}.
\newblock \bibinfo{journal}{\emph{arXiv preprint arXiv:2005.00268}}
  (\bibinfo{year}{2020}).
\newblock


\bibitem[\protect\citeauthoryear{Sheng, Song, Tan, Ren, Ye, Zhang, and
  Qin}{Sheng et~al\mbox{.}}{2021}]%
        {aaai2021_songmass}
\bibfield{author}{\bibinfo{person}{Zhonghao Sheng}, \bibinfo{person}{Kaitao
  Song}, \bibinfo{person}{Xu Tan}, \bibinfo{person}{Yi Ren},
  \bibinfo{person}{Wei Ye}, \bibinfo{person}{Shikun Zhang}, {and}
  \bibinfo{person}{Tao Qin}.} \bibinfo{year}{2021}\natexlab{}.
\newblock \showarticletitle{SongMASS: Automatic Song Writing with Pre-training
  and Alignment Constraint}. In \bibinfo{booktitle}{\emph{Thirty-Fifth {AAAI}
  Conference on Artificial Intelligence, {AAAI} 2021, Thirty-Third Conference
  on Innovative Applications of Artificial Intelligence, {IAAI} 2021, The
  Eleventh Symposium on Educational Advances in Artificial Intelligence, {EAAI}
  2021, Virtual Event, February 2-9, 2021}}. \bibinfo{publisher}{{AAAI} Press},
  \bibinfo{pages}{13798--13805}.
\newblock
\urldef\tempurl%
\url{https://ojs.aaai.org/index.php/AAAI/article/view/17626}
\showURL{%
\tempurl}


\bibitem[\protect\citeauthoryear{Shi, Zhou, Miao, and Li}{Shi
  et~al\mbox{.}}{2020}]%
        {shi2020dispersed}
\bibfield{author}{\bibinfo{person}{Wenxian Shi}, \bibinfo{person}{Hao Zhou},
  \bibinfo{person}{Ning Miao}, {and} \bibinfo{person}{Lei Li}.}
  \bibinfo{year}{2020}\natexlab{}.
\newblock \showarticletitle{Dispersed Exponential Family Mixture VAEs for
  Interpretable Text Generation}. In \bibinfo{booktitle}{\emph{International
  Conference on Machine Learning}}. PMLR, \bibinfo{pages}{8840--8851}.
\newblock
\urldef\tempurl%
\url{http://proceedings.mlr.press/v119/shi20f/shi20f.pdf}
\showURL{%
\tempurl}


\bibitem[\protect\citeauthoryear{Shin, Razeghi, Logan~IV, Wallace, and
  Singh}{Shin et~al\mbox{.}}{2020}]%
        {AutoPrompt}
\bibfield{author}{\bibinfo{person}{Taylor Shin}, \bibinfo{person}{Yasaman
  Razeghi}, \bibinfo{person}{Robert~L. Logan~IV}, \bibinfo{person}{Eric
  Wallace}, {and} \bibinfo{person}{Sameer Singh}.}
  \bibinfo{year}{2020}\natexlab{}.
\newblock \showarticletitle{{A}uto{P}rompt: {E}liciting {K}nowledge from
  {L}anguage {M}odels with {A}utomatically {G}enerated {P}rompts}. In
  \bibinfo{booktitle}{\emph{Proceedings of the 2020 Conference on Empirical
  Methods in Natural Language Processing (EMNLP)}}.
  \bibinfo{publisher}{Association for Computational Linguistics},
  \bibinfo{address}{Online}, \bibinfo{pages}{4222--4235}.
\newblock
\urldef\tempurl%
\url{https://doi.org/10.18653/v1/2020.emnlp-main.346}
\showDOI{\tempurl}


\bibitem[\protect\citeauthoryear{Sohn, Yan, and Lee}{Sohn
  et~al\mbox{.}}{2015}]%
        {LSOR_DCGM}
\bibfield{author}{\bibinfo{person}{Kihyuk Sohn}, \bibinfo{person}{Xinchen Yan},
  {and} \bibinfo{person}{Honglak Lee}.} \bibinfo{year}{2015}\natexlab{}.
\newblock \showarticletitle{Learning Structured Output Representation Using
  Deep Conditional Generative Models}. In \bibinfo{booktitle}{\emph{Proceedings
  of the 28th International Conference on Neural Information Processing Systems
  - Volume 2}} (Montreal, Canada) \emph{(\bibinfo{series}{NIPS'15})}.
  \bibinfo{publisher}{MIT Press}, \bibinfo{address}{Cambridge, MA, USA},
  \bibinfo{pages}{3483–3491}.
\newblock
\urldef\tempurl%
\url{https://dl.acm.org/doi/10.5555/2969442.2969628}
\showURL{%
\tempurl}


\bibitem[\protect\citeauthoryear{Song, Wang, Zhang, Zhang, and Liu}{Song
  et~al\mbox{.}}{2021}]%
        {acl2021_persona}
\bibfield{author}{\bibinfo{person}{Haoyu Song}, \bibinfo{person}{Yan Wang},
  \bibinfo{person}{Kaiyan Zhang}, \bibinfo{person}{Wei{-}Nan Zhang}, {and}
  \bibinfo{person}{Ting Liu}.} \bibinfo{year}{2021}\natexlab{}.
\newblock \showarticletitle{BoB: {BERT} Over {BERT} for Training Persona-based
  Dialogue Models from Limited Personalized Data}. In
  \bibinfo{booktitle}{\emph{Proceedings of the 59th Annual Meeting of the
  Association for Computational Linguistics and the 11th International Joint
  Conference on Natural Language Processing, {ACL/IJCNLP} 2021, (Volume 1: Long
  Papers), Virtual Event, August 1-6, 2021}},
  \bibfield{editor}{\bibinfo{person}{Chengqing Zong}, \bibinfo{person}{Fei
  Xia}, \bibinfo{person}{Wenjie Li}, {and} \bibinfo{person}{Roberto Navigli}}
  (Eds.). \bibinfo{publisher}{Association for Computational Linguistics},
  \bibinfo{pages}{167--177}.
\newblock
\urldef\tempurl%
\url{https://doi.org/10.18653/v1/2021.acl-long.14}
\showDOI{\tempurl}


\bibitem[\protect\citeauthoryear{Song, Wang, Zhang, Liu, and Liu}{Song
  et~al\mbox{.}}{2020b}]%
        {acl2020_persona}
\bibfield{author}{\bibinfo{person}{Haoyu Song}, \bibinfo{person}{Yan Wang},
  \bibinfo{person}{Weinan Zhang}, \bibinfo{person}{Xiaojiang Liu}, {and}
  \bibinfo{person}{Ting Liu}.} \bibinfo{year}{2020}\natexlab{b}.
\newblock \showarticletitle{Generate, Delete and Rewrite: {A} Three-Stage
  Framework for Improving Persona Consistency of Dialogue Generation}. In
  \bibinfo{booktitle}{\emph{Proceedings of the 58th Annual Meeting of the
  Association for Computational Linguistics, {ACL} 2020, Online, July 5-10,
  2020}}, \bibfield{editor}{\bibinfo{person}{Dan Jurafsky},
  \bibinfo{person}{Joyce Chai}, \bibinfo{person}{Natalie Schluter}, {and}
  \bibinfo{person}{Joel~R. Tetreault}} (Eds.). \bibinfo{publisher}{Association
  for Computational Linguistics}, \bibinfo{pages}{5821--5831}.
\newblock
\urldef\tempurl%
\url{https://doi.org/10.18653/v1/2020.acl-main.516}
\showDOI{\tempurl}


\bibitem[\protect\citeauthoryear{Song, Wang, Su, Zhang, Xu, Ge, and Yu}{Song
  et~al\mbox{.}}{2020a}]%
        {acl2020_graph2text}
\bibfield{author}{\bibinfo{person}{Linfeng Song}, \bibinfo{person}{Ante Wang},
  \bibinfo{person}{Jinsong Su}, \bibinfo{person}{Yue Zhang},
  \bibinfo{person}{Kun Xu}, \bibinfo{person}{Yubin Ge}, {and}
  \bibinfo{person}{Dong Yu}.} \bibinfo{year}{2020}\natexlab{a}.
\newblock \showarticletitle{Structural Information Preserving for Graph-to-Text
  Generation}. In \bibinfo{booktitle}{\emph{Proceedings of the 58th Annual
  Meeting of the Association for Computational Linguistics, {ACL} 2020, Online,
  July 5-10, 2020}}, \bibfield{editor}{\bibinfo{person}{Dan Jurafsky},
  \bibinfo{person}{Joyce Chai}, \bibinfo{person}{Natalie Schluter}, {and}
  \bibinfo{person}{Joel~R. Tetreault}} (Eds.). \bibinfo{publisher}{Association
  for Computational Linguistics}, \bibinfo{pages}{7987--7998}.
\newblock
\urldef\tempurl%
\url{https://doi.org/10.18653/v1/2020.acl-main.712}
\showDOI{\tempurl}


\bibitem[\protect\citeauthoryear{Song, Zheng, Liu, Xu, and Huang}{Song
  et~al\mbox{.}}{2019}]%
        {acl_2019_emo_ctg}
\bibfield{author}{\bibinfo{person}{Zhenqiao Song}, \bibinfo{person}{Xiaoqing
  Zheng}, \bibinfo{person}{Lu Liu}, \bibinfo{person}{Mu Xu}, {and}
  \bibinfo{person}{Xuanjing Huang}.} \bibinfo{year}{2019}\natexlab{}.
\newblock \showarticletitle{Generating Responses with a Specific Emotion in
  Dialog}. In \bibinfo{booktitle}{\emph{Proceedings of the 57th Conference of
  the Association for Computational Linguistics, {ACL} 2019, Florence, Italy,
  July 28- August 2, 2019, Volume 1: Long Papers}},
  \bibfield{editor}{\bibinfo{person}{Anna Korhonen}, \bibinfo{person}{David~R.
  Traum}, {and} \bibinfo{person}{Llu{\'{\i}}s M{\`{a}}rquez}} (Eds.).
  \bibinfo{publisher}{Association for Computational Linguistics},
  \bibinfo{pages}{3685--3695}.
\newblock
\urldef\tempurl%
\url{https://doi.org/10.18653/v1/p19-1359}
\showDOI{\tempurl}


\bibitem[\protect\citeauthoryear{Stiennon, Ouyang, Wu, Ziegler, Lowe, Voss,
  Radford, Amodei, and Christiano}{Stiennon et~al\mbox{.}}{2020}]%
        {lshf_nips2020}
\bibfield{author}{\bibinfo{person}{Nisan Stiennon}, \bibinfo{person}{Long
  Ouyang}, \bibinfo{person}{Jeff Wu}, \bibinfo{person}{Daniel~M. Ziegler},
  \bibinfo{person}{Ryan Lowe}, \bibinfo{person}{Chelsea Voss},
  \bibinfo{person}{Alec Radford}, \bibinfo{person}{Dario Amodei}, {and}
  \bibinfo{person}{Paul Christiano}.} \bibinfo{year}{2020}\natexlab{}.
\newblock \showarticletitle{Learning to Summarize from Human Feedback}. In
  \bibinfo{booktitle}{\emph{Proceedings of the 34th International Conference on
  Neural Information Processing Systems}} (Vancouver, BC, Canada)
  \emph{(\bibinfo{series}{NIPS'20})}. \bibinfo{publisher}{Curran Associates
  Inc.}, \bibinfo{address}{Red Hook, NY, USA}, Article
  \bibinfo{articleno}{253}, \bibinfo{numpages}{14}~pages.
\newblock
\showISBNx{9781713829546}
\urldef\tempurl%
\url{https://proceedings.neurips.cc/paper/2020/file/1f89885d556929e98d3ef9b86448f951-Paper.pdf}
\showURL{%
\tempurl}


\bibitem[\protect\citeauthoryear{Su, Vandyke, Wang, Fang, and Collier}{Su
  et~al\mbox{.}}{2021}]%
        {findins_acl2021_PlanthenGenerate}
\bibfield{author}{\bibinfo{person}{Yixuan Su}, \bibinfo{person}{David Vandyke},
  \bibinfo{person}{Sihui Wang}, \bibinfo{person}{Yimai Fang}, {and}
  \bibinfo{person}{Nigel Collier}.} \bibinfo{year}{2021}\natexlab{}.
\newblock \showarticletitle{Plan-then-Generate: Controlled Data-to-Text
  Generation via Planning}. In \bibinfo{booktitle}{\emph{Findings of the
  Association for Computational Linguistics: {EMNLP} 2021, Virtual Event /
  Punta Cana, Dominican Republic, 16-20 November, 2021}},
  \bibfield{editor}{\bibinfo{person}{Marie{-}Francine Moens},
  \bibinfo{person}{Xuanjing Huang}, \bibinfo{person}{Lucia Specia}, {and}
  \bibinfo{person}{Scott~Wen{-}tau Yih}} (Eds.).
  \bibinfo{publisher}{Association for Computational Linguistics},
  \bibinfo{pages}{895--909}.
\newblock
\urldef\tempurl%
\url{https://aclanthology.org/2021.findings-emnlp.76}
\showURL{%
\tempurl}


\bibitem[\protect\citeauthoryear{Tambwekar, Dhuliawala, Martin, Mehta,
  Harrison, and Riedl}{Tambwekar et~al\mbox{.}}{2019}]%
        {tambwekar2018controllable}
\bibfield{author}{\bibinfo{person}{Pradyumna Tambwekar},
  \bibinfo{person}{Murtaza Dhuliawala}, \bibinfo{person}{Lara~J. Martin},
  \bibinfo{person}{Animesh Mehta}, \bibinfo{person}{Brent Harrison}, {and}
  \bibinfo{person}{Mark~O. Riedl}.} \bibinfo{year}{2019}\natexlab{}.
\newblock \showarticletitle{Controllable Neural Story Plot Generation via
  Reward Shaping}. In \bibinfo{booktitle}{\emph{Proceedings of the
  Twenty-Eighth International Joint Conference on Artificial Intelligence,
  {IJCAI-19}}}. \bibinfo{publisher}{International Joint Conferences on
  Artificial Intelligence Organization}, \bibinfo{pages}{5982--5988}.
\newblock
\urldef\tempurl%
\url{https://doi.org/10.24963/ijcai.2019/829}
\showDOI{\tempurl}


\bibitem[\protect\citeauthoryear{Tang, Li, and Jin}{Tang et~al\mbox{.}}{2019}]%
        {tang2019topic}
\bibfield{author}{\bibinfo{person}{Hongyin Tang}, \bibinfo{person}{Miao Li},
  {and} \bibinfo{person}{Beihong Jin}.} \bibinfo{year}{2019}\natexlab{}.
\newblock \showarticletitle{A Topic Augmented Text Generation Model: Joint
  Learning of Semantics and Structural Features}. In
  \bibinfo{booktitle}{\emph{Proceedings of the 2019 Conference on Empirical
  Methods in Natural Language Processing and the 9th International Joint
  Conference on Natural Language Processing (EMNLP-IJCNLP)}}.
  \bibinfo{publisher}{Association for Computational Linguistics},
  \bibinfo{address}{Hong Kong, China}, \bibinfo{pages}{5090--5099}.
\newblock
\urldef\tempurl%
\url{https://doi.org/10.18653/v1/D19-1513}
\showDOI{\tempurl}


\bibitem[\protect\citeauthoryear{Touvron, Lavril, Izacard, Martinet, Lachaux,
  Lacroix, Rozi{\`e}re, Goyal, Hambro, Azhar, et~al\mbox{.}}{Touvron
  et~al\mbox{.}}{2023}]%
        {touvron2023llama}
\bibfield{author}{\bibinfo{person}{Hugo Touvron}, \bibinfo{person}{Thibaut
  Lavril}, \bibinfo{person}{Gautier Izacard}, \bibinfo{person}{Xavier
  Martinet}, \bibinfo{person}{Marie-Anne Lachaux},
  \bibinfo{person}{Timoth{\'e}e Lacroix}, \bibinfo{person}{Baptiste
  Rozi{\`e}re}, \bibinfo{person}{Naman Goyal}, \bibinfo{person}{Eric Hambro},
  \bibinfo{person}{Faisal Azhar}, {et~al\mbox{.}}}
  \bibinfo{year}{2023}\natexlab{}.
\newblock \showarticletitle{Llama: Open and efficient foundation language
  models}.
\newblock \bibinfo{journal}{\emph{arXiv preprint arXiv:2302.13971}}
  (\bibinfo{year}{2023}).
\newblock


\bibitem[\protect\citeauthoryear{Tu, Pang, Wiseman, and Gimpel}{Tu
  et~al\mbox{.}}{2020}]%
        {engine_ebm_acl2020}
\bibfield{author}{\bibinfo{person}{Lifu Tu}, \bibinfo{person}{Richard~Yuanzhe
  Pang}, \bibinfo{person}{Sam Wiseman}, {and} \bibinfo{person}{Kevin Gimpel}.}
  \bibinfo{year}{2020}\natexlab{}.
\newblock \showarticletitle{{ENGINE:} Energy-Based Inference Networks for
  Non-Autoregressive Machine Translation}. In
  \bibinfo{booktitle}{\emph{Proceedings of the 58th Annual Meeting of the
  Association for Computational Linguistics, {ACL} 2020, Online, July 5-10,
  2020}}, \bibfield{editor}{\bibinfo{person}{Dan Jurafsky},
  \bibinfo{person}{Joyce Chai}, \bibinfo{person}{Natalie Schluter}, {and}
  \bibinfo{person}{Joel~R. Tetreault}} (Eds.). \bibinfo{publisher}{Association
  for Computational Linguistics}, \bibinfo{pages}{2819--2826}.
\newblock
\urldef\tempurl%
\url{https://doi.org/10.18653/v1/2020.acl-main.251}
\showDOI{\tempurl}


\bibitem[\protect\citeauthoryear{{van der Lee}, Gatt, {van Miltenburg}, and
  Krahmer}{{van der Lee} et~al\mbox{.}}{2021}]%
        {van2020human}
\bibfield{author}{\bibinfo{person}{Chris {van der Lee}},
  \bibinfo{person}{Albert Gatt}, \bibinfo{person}{Emiel {van Miltenburg}},
  {and} \bibinfo{person}{Emiel Krahmer}.} \bibinfo{year}{2021}\natexlab{}.
\newblock \showarticletitle{Human evaluation of automatically generated text:
  Current trends and best practice guidelines}.
\newblock \bibinfo{journal}{\emph{Computer Speech \& Language}}
  \bibinfo{volume}{67} (\bibinfo{year}{2021}), \bibinfo{pages}{101151}.
\newblock
\showISSN{0885-2308}
\urldef\tempurl%
\url{https://doi.org/10.1016/j.csl.2020.101151}
\showDOI{\tempurl}


\bibitem[\protect\citeauthoryear{Vaswani, Shazeer, Parmar, Uszkoreit, Jones,
  Gomez, Kaiser, and Polosukhin}{Vaswani et~al\mbox{.}}{2017}]%
        {attention_is_all_your_need}
\bibfield{author}{\bibinfo{person}{Ashish Vaswani}, \bibinfo{person}{Noam
  Shazeer}, \bibinfo{person}{Niki Parmar}, \bibinfo{person}{Jakob Uszkoreit},
  \bibinfo{person}{Llion Jones}, \bibinfo{person}{Aidan~N Gomez},
  \bibinfo{person}{{\L}ukasz Kaiser}, {and} \bibinfo{person}{Illia
  Polosukhin}.} \bibinfo{year}{2017}\natexlab{}.
\newblock \showarticletitle{Attention is all you need}. In
  \bibinfo{booktitle}{\emph{Advances in neural information processing
  systems}}. \bibinfo{pages}{5998--6008}.
\newblock
\urldef\tempurl%
\url{https://papers.nips.cc/paper/2017/file/3f5ee243547dee91fbd053c1c4a845aa-Paper.pdf}
\showURL{%
\tempurl}


\bibitem[\protect\citeauthoryear{Vechtomova, Bahuleyan, Ghabussi, and
  John}{Vechtomova et~al\mbox{.}}{2018}]%
        {chinses_lyric}
\bibfield{author}{\bibinfo{person}{Olga Vechtomova}, \bibinfo{person}{Hareesh
  Bahuleyan}, \bibinfo{person}{Amirpasha Ghabussi}, {and}
  \bibinfo{person}{Vineet John}.} \bibinfo{year}{2018}\natexlab{}.
\newblock \showarticletitle{Generating lyrics with variational autoencoder and
  multi-modal artist embeddings}.
\newblock \bibinfo{journal}{\emph{CoRR}}  \bibinfo{volume}{abs/1812.08318}
  (\bibinfo{year}{2018}).
\newblock
\showeprint[arxiv]{1812.08318}
\urldef\tempurl%
\url{http://arxiv.org/abs/1812.08318}
\showURL{%
\tempurl}


\bibitem[\protect\citeauthoryear{Wang, Chen, He, Zhong, Tao, and Yang}{Wang
  et~al\mbox{.}}{2021a}]%
        {wang-etal-2021-template}
\bibfield{author}{\bibinfo{person}{Dingmin Wang}, \bibinfo{person}{Ziyao Chen},
  \bibinfo{person}{Wanwei He}, \bibinfo{person}{Li Zhong},
  \bibinfo{person}{Yunzhe Tao}, {and} \bibinfo{person}{Min Yang}.}
  \bibinfo{year}{2021}\natexlab{a}.
\newblock \showarticletitle{A Template-guided Hybrid Pointer Network for
  Knowledge-based Task-oriented Dialogue Systems}. In
  \bibinfo{booktitle}{\emph{Proceedings of the 1st Workshop on
  Document-grounded Dialogue and Conversational Question Answering (DialDoc
  2021)}}. \bibinfo{publisher}{Association for Computational Linguistics},
  \bibinfo{address}{Online}, \bibinfo{pages}{18--28}.
\newblock
\urldef\tempurl%
\url{https://doi.org/10.18653/v1/2021.dialdoc-1.3}
\showDOI{\tempurl}


\bibitem[\protect\citeauthoryear{Wang and Wan}{Wang and Wan}{2018}]%
        {SentiGAN}
\bibfield{author}{\bibinfo{person}{Ke Wang} {and} \bibinfo{person}{Xiaojun
  Wan}.} \bibinfo{year}{2018}\natexlab{}.
\newblock \showarticletitle{SentiGAN: Generating Sentimental Texts via Mixture
  Adversarial Networks}. In \bibinfo{booktitle}{\emph{Proceedings of the
  Twenty-Seventh International Joint Conference on Artificial Intelligence,
  {IJCAI-18}}}. \bibinfo{publisher}{International Joint Conferences on
  Artificial Intelligence Organization}, \bibinfo{pages}{4446--4452}.
\newblock
\urldef\tempurl%
\url{https://doi.org/10.24963/ijcai.2018/618}
\showDOI{\tempurl}


\bibitem[\protect\citeauthoryear{Wang, Wei, Cheng, Li, Shan, Zhang, Zhang, and
  Huang}{Wang et~al\mbox{.}}{2020}]%
        {wang2019keep}
\bibfield{author}{\bibinfo{person}{Ruize Wang}, \bibinfo{person}{Zhongyu Wei},
  \bibinfo{person}{Ying Cheng}, \bibinfo{person}{Piji Li},
  \bibinfo{person}{Haijun Shan}, \bibinfo{person}{Ji Zhang},
  \bibinfo{person}{Qi Zhang}, {and} \bibinfo{person}{Xuanjing Huang}.}
  \bibinfo{year}{2020}\natexlab{}.
\newblock \showarticletitle{Keep it Consistent: Topic-Aware Storytelling from
  an Image Stream via Iterative Multi-agent Communication}. In
  \bibinfo{booktitle}{\emph{Proceedings of the 28th International Conference on
  Computational Linguistics}}. \bibinfo{publisher}{International Committee on
  Computational Linguistics}, \bibinfo{address}{Barcelona, Spain (Online)},
  \bibinfo{pages}{2250--2260}.
\newblock
\urldef\tempurl%
\url{https://doi.org/10.18653/v1/2020.coling-main.204}
\showDOI{\tempurl}


\bibitem[\protect\citeauthoryear{Wang, Gan, Xu, Zhang, Wang, Shen, Chen, and
  Carin}{Wang et~al\mbox{.}}{2019}]%
        {wang2019topic}
\bibfield{author}{\bibinfo{person}{Wenlin Wang}, \bibinfo{person}{Zhe Gan},
  \bibinfo{person}{Hongteng Xu}, \bibinfo{person}{Ruiyi Zhang},
  \bibinfo{person}{Guoyin Wang}, \bibinfo{person}{Dinghan Shen},
  \bibinfo{person}{Changyou Chen}, {and} \bibinfo{person}{Lawrence Carin}.}
  \bibinfo{year}{2019}\natexlab{}.
\newblock \showarticletitle{Topic-Guided Variational Auto-Encoder for Text
  Generation}. In \bibinfo{booktitle}{\emph{Proceedings of the 2019 Conference
  of the North {A}merican Chapter of the Association for Computational
  Linguistics: Human Language Technologies, Volume 1 (Long and Short Papers)}}.
  \bibinfo{publisher}{Association for Computational Linguistics},
  \bibinfo{address}{Minneapolis, Minnesota}, \bibinfo{pages}{166--177}.
\newblock
\urldef\tempurl%
\url{https://doi.org/10.18653/v1/N19-1015}
\showDOI{\tempurl}


\bibitem[\protect\citeauthoryear{Wang, Wood, Wan, Dras, and Johnson}{Wang
  et~al\mbox{.}}{2021b}]%
        {metion_flag}
\bibfield{author}{\bibinfo{person}{Yufei Wang}, \bibinfo{person}{Ian Wood},
  \bibinfo{person}{Stephen Wan}, \bibinfo{person}{Mark Dras}, {and}
  \bibinfo{person}{Mark Johnson}.} \bibinfo{year}{2021}\natexlab{b}.
\newblock \showarticletitle{Mention Flags ({MF}): Constraining
  Transformer-based Text Generators}. In \bibinfo{booktitle}{\emph{Proceedings
  of the 59th Annual Meeting of the Association for Computational Linguistics
  and the 11th International Joint Conference on Natural Language Processing
  (Volume 1: Long Papers)}}. \bibinfo{publisher}{Association for Computational
  Linguistics}, \bibinfo{address}{Online}, \bibinfo{pages}{103--113}.
\newblock
\urldef\tempurl%
\url{https://doi.org/10.18653/v1/2021.acl-long.9}
\showDOI{\tempurl}


\bibitem[\protect\citeauthoryear{Wei, Bosma, Zhao, Guu, Yu, Lester, Du, Dai,
  and Le}{Wei et~al\mbox{.}}{2022}]%
        {flan}
\bibfield{author}{\bibinfo{person}{Jason Wei}, \bibinfo{person}{Maarten Bosma},
  \bibinfo{person}{Vincent Zhao}, \bibinfo{person}{Kelvin Guu},
  \bibinfo{person}{Adams~Wei Yu}, \bibinfo{person}{Brian Lester},
  \bibinfo{person}{Nan Du}, \bibinfo{person}{Andrew~M. Dai}, {and}
  \bibinfo{person}{Quoc~V Le}.} \bibinfo{year}{2022}\natexlab{}.
\newblock \showarticletitle{Finetuned Language Models are Zero-Shot Learners}.
  In \bibinfo{booktitle}{\emph{International Conference on Learning
  Representations}}.
\newblock
\urldef\tempurl%
\url{https://openreview.net/forum?id=gEZrGCozdqR}
\showURL{%
\tempurl}


\bibitem[\protect\citeauthoryear{Wei, Liu, Mao, Guo, Zhu, Zhou, and Hu}{Wei
  et~al\mbox{.}}{2019}]%
        {cikm2019_emo_ctg}
\bibfield{author}{\bibinfo{person}{Wei Wei}, \bibinfo{person}{Jiayi Liu},
  \bibinfo{person}{Xianling Mao}, \bibinfo{person}{Guibing Guo},
  \bibinfo{person}{Feida Zhu}, \bibinfo{person}{Pan Zhou}, {and}
  \bibinfo{person}{Yuchong Hu}.} \bibinfo{year}{2019}\natexlab{}.
\newblock \showarticletitle{Emotion-aware Chat Machine: Automatic Emotional
  Response Generation for Human-like Emotional Interaction}. In
  \bibinfo{booktitle}{\emph{Proceedings of the 28th {ACM} International
  Conference on Information and Knowledge Management, {CIKM} 2019, Beijing,
  China, November 3-7, 2019}}, \bibfield{editor}{\bibinfo{person}{Wenwu Zhu},
  \bibinfo{person}{Dacheng Tao}, \bibinfo{person}{Xueqi Cheng},
  \bibinfo{person}{Peng Cui}, \bibinfo{person}{Elke~A. Rundensteiner},
  \bibinfo{person}{David Carmel}, \bibinfo{person}{Qi~He}, {and}
  \bibinfo{person}{Jeffrey~Xu Yu}} (Eds.). \bibinfo{publisher}{{ACM}},
  \bibinfo{pages}{1401--1410}.
\newblock
\urldef\tempurl%
\url{https://doi.org/10.1145/3357384.3357937}
\showDOI{\tempurl}


\bibitem[\protect\citeauthoryear{Welleck, Kulikov, Roller, Dinan, Cho, and
  Weston}{Welleck et~al\mbox{.}}{2019}]%
        {welleck2019neural}
\bibfield{author}{\bibinfo{person}{Sean Welleck}, \bibinfo{person}{Ilia
  Kulikov}, \bibinfo{person}{Stephen Roller}, \bibinfo{person}{Emily Dinan},
  \bibinfo{person}{Kyunghyun Cho}, {and} \bibinfo{person}{Jason Weston}.}
  \bibinfo{year}{2019}\natexlab{}.
\newblock \showarticletitle{Neural Text Generation With Unlikelihood Training}.
  In \bibinfo{booktitle}{\emph{International Conference on Learning
  Representations}}.
\newblock
\urldef\tempurl%
\url{https://arxiv.org/abs/1908.04319}
\showURL{%
\tempurl}


\bibitem[\protect\citeauthoryear{Wu, Galley, Brockett, Zhang, Gao, Quirk,
  Koncel{-}Kedziorski, Gao, Hajishirzi, Ostendorf, and Dolan}{Wu
  et~al\mbox{.}}{2021}]%
        {wu2020controllable}
\bibfield{author}{\bibinfo{person}{Zeqiu Wu}, \bibinfo{person}{Michel Galley},
  \bibinfo{person}{Chris Brockett}, \bibinfo{person}{Yizhe Zhang},
  \bibinfo{person}{Xiang Gao}, \bibinfo{person}{Chris Quirk},
  \bibinfo{person}{Rik Koncel{-}Kedziorski}, \bibinfo{person}{Jianfeng Gao},
  \bibinfo{person}{Hannaneh Hajishirzi}, \bibinfo{person}{Mari Ostendorf},
  {and} \bibinfo{person}{Bill Dolan}.} \bibinfo{year}{2021}\natexlab{}.
\newblock \showarticletitle{A Controllable Model of Grounded Response
  Generation}. In \bibinfo{booktitle}{\emph{Thirty-Fifth {AAAI} Conference on
  Artificial Intelligence, {AAAI} 2021, Thirty-Third Conference on Innovative
  Applications of Artificial Intelligence, {IAAI} 2021, The Eleventh Symposium
  on Educational Advances in Artificial Intelligence, {EAAI} 2021, Virtual
  Event, February 2-9, 2021}}. \bibinfo{publisher}{{AAAI} Press},
  \bibinfo{pages}{14085--14093}.
\newblock
\urldef\tempurl%
\url{https://ojs.aaai.org/index.php/AAAI/article/view/17658}
\showURL{%
\tempurl}


\bibitem[\protect\citeauthoryear{Xie, Raghunathan, Liang, and Ma}{Xie
  et~al\mbox{.}}{2022}]%
        {xie2022an}
\bibfield{author}{\bibinfo{person}{Sang~Michael Xie}, \bibinfo{person}{Aditi
  Raghunathan}, \bibinfo{person}{Percy Liang}, {and} \bibinfo{person}{Tengyu
  Ma}.} \bibinfo{year}{2022}\natexlab{}.
\newblock \showarticletitle{An Explanation of In-context Learning as Implicit
  Bayesian Inference}. In \bibinfo{booktitle}{\emph{International Conference on
  Learning Representations}}.
\newblock
\urldef\tempurl%
\url{https://openreview.net/forum?id=RdJVFCHjUMI}
\showURL{%
\tempurl}


\bibitem[\protect\citeauthoryear{Xu, Cao, and Cheung}{Xu et~al\mbox{.}}{2019}]%
        {unsupervised_GVD}
\bibfield{author}{\bibinfo{person}{Peng Xu}, \bibinfo{person}{Yanshuai Cao},
  {and} \bibinfo{person}{Jackie Chi~Kit Cheung}.}
  \bibinfo{year}{2019}\natexlab{}.
\newblock \showarticletitle{Unsupervised Controllable Text Generation with
  Global Variation Discovery and Disentanglement}.
\newblock \bibinfo{journal}{\emph{CoRR}}  \bibinfo{volume}{abs/1905.11975}
  (\bibinfo{year}{2019}).
\newblock
\showeprint[arxiv]{1905.11975}
\urldef\tempurl%
\url{http://arxiv.org/abs/1905.11975}
\showURL{%
\tempurl}


\bibitem[\protect\citeauthoryear{Xu, Patwary, Shoeybi, Puri, Fung, Anandkumar,
  and Catanzaro}{Xu et~al\mbox{.}}{2020}]%
        {MEGATRON-CNTRL}
\bibfield{author}{\bibinfo{person}{Peng Xu}, \bibinfo{person}{Mostofa Patwary},
  \bibinfo{person}{Mohammad Shoeybi}, \bibinfo{person}{Raul Puri},
  \bibinfo{person}{Pascale Fung}, \bibinfo{person}{Anima Anandkumar}, {and}
  \bibinfo{person}{Bryan Catanzaro}.} \bibinfo{year}{2020}\natexlab{}.
\newblock \showarticletitle{{MEGATRON}-{CNTRL}: Controllable Story Generation
  with External Knowledge Using Large-Scale Language Models}. In
  \bibinfo{booktitle}{\emph{Proceedings of the 2020 Conference on Empirical
  Methods in Natural Language Processing (EMNLP)}}.
  \bibinfo{publisher}{Association for Computational Linguistics},
  \bibinfo{address}{Online}, \bibinfo{pages}{2831--2845}.
\newblock
\urldef\tempurl%
\url{https://doi.org/10.18653/v1/2020.emnlp-main.226}
\showDOI{\tempurl}


\bibitem[\protect\citeauthoryear{Yang and Klein}{Yang and Klein}{2021}]%
        {FUDGE}
\bibfield{author}{\bibinfo{person}{Kevin Yang} {and} \bibinfo{person}{Dan
  Klein}.} \bibinfo{year}{2021}\natexlab{}.
\newblock \showarticletitle{{FUDGE}: Controlled Text Generation With Future
  Discriminators}. In \bibinfo{booktitle}{\emph{Proceedings of the 2021
  Conference of the North American Chapter of the Association for Computational
  Linguistics: Human Language Technologies}}. \bibinfo{publisher}{Association
  for Computational Linguistics}, \bibinfo{address}{Online},
  \bibinfo{pages}{3511--3535}.
\newblock
\urldef\tempurl%
\url{https://doi.org/10.18653/v1/2021.naacl-main.276}
\showDOI{\tempurl}


\bibitem[\protect\citeauthoryear{Yang, Liu, Lei, Yang, Xue, Chen, and Xie}{Yang
  et~al\mbox{.}}{2022}]%
        {tail_prompt}
\bibfield{author}{\bibinfo{person}{Kexin Yang}, \bibinfo{person}{Dayiheng Liu},
  \bibinfo{person}{Wenqiang Lei}, \bibinfo{person}{Baosong Yang},
  \bibinfo{person}{Mingfeng Xue}, \bibinfo{person}{Boxing Chen}, {and}
  \bibinfo{person}{Jun Xie}.} \bibinfo{year}{2022}\natexlab{}.
\newblock \bibinfo{title}{Tailor: A Prompt-Based Approach to Attribute-Based
  Controlled Text Generation}.
\newblock
\newblock
\urldef\tempurl%
\url{https://doi.org/10.48550/ARXIV.2204.13362}
\showDOI{\tempurl}


\bibitem[\protect\citeauthoryear{Yang, Li, Luo, Liu, and Sun}{Yang
  et~al\mbox{.}}{2019b}]%
        {yang2019enhancing}
\bibfield{author}{\bibinfo{person}{Pengcheng Yang}, \bibinfo{person}{Lei Li},
  \bibinfo{person}{Fuli Luo}, \bibinfo{person}{Tianyu Liu}, {and}
  \bibinfo{person}{Xu Sun}.} \bibinfo{year}{2019}\natexlab{b}.
\newblock \showarticletitle{Enhancing Topic-to-Essay Generation with External
  Commonsense Knowledge}. In \bibinfo{booktitle}{\emph{Proceedings of the 57th
  Annual Meeting of the Association for Computational Linguistics}}.
  \bibinfo{publisher}{Association for Computational Linguistics},
  \bibinfo{address}{Florence, Italy}, \bibinfo{pages}{2002--2012}.
\newblock
\urldef\tempurl%
\url{https://doi.org/10.18653/v1/P19-1193}
\showDOI{\tempurl}


\bibitem[\protect\citeauthoryear{Yang, Lin, Suo, and Li}{Yang
  et~al\mbox{.}}{2018}]%
        {thematic_poetry}
\bibfield{author}{\bibinfo{person}{Xiaopeng Yang}, \bibinfo{person}{Xiaowen
  Lin}, \bibinfo{person}{Shunda Suo}, {and} \bibinfo{person}{Ming Li}.}
  \bibinfo{year}{2018}\natexlab{}.
\newblock \showarticletitle{Generating Thematic Chinese Poetry Using
  Conditional Variational Autoencoders with Hybrid Decoders}. In
  \bibinfo{booktitle}{\emph{Proceedings of the 27th International Joint
  Conference on Artificial Intelligence}} (Stockholm, Sweden)
  \emph{(\bibinfo{series}{IJCAI'18})}. \bibinfo{publisher}{AAAI Press},
  \bibinfo{pages}{4539–4545}.
\newblock
\showISBNx{9780999241127}
\urldef\tempurl%
\url{https://arxiv.org/abs/1711.07632}
\showURL{%
\tempurl}


\bibitem[\protect\citeauthoryear{Yang, Dai, Yang, Carbonell, Salakhutdinov, and
  Le}{Yang et~al\mbox{.}}{2019a}]%
        {Xlnet}
\bibfield{author}{\bibinfo{person}{Zhilin Yang}, \bibinfo{person}{Zihang Dai},
  \bibinfo{person}{Yiming Yang}, \bibinfo{person}{Jaime Carbonell},
  \bibinfo{person}{Russ~R Salakhutdinov}, {and} \bibinfo{person}{Quoc~V Le}.}
  \bibinfo{year}{2019}\natexlab{a}.
\newblock \showarticletitle{XLNet: Generalized Autoregressive Pretraining for
  Language Understanding}. In \bibinfo{booktitle}{\emph{Advances in Neural
  Information Processing Systems}},
  \bibfield{editor}{\bibinfo{person}{H.~Wallach},
  \bibinfo{person}{H.~Larochelle}, \bibinfo{person}{A.~Beygelzimer},
  \bibinfo{person}{F.~d\textquotesingle Alch\'{e}-Buc},
  \bibinfo{person}{E.~Fox}, {and} \bibinfo{person}{R.~Garnett}} (Eds.),
  Vol.~\bibinfo{volume}{32}. \bibinfo{publisher}{Curran Associates, Inc.}
\newblock
\urldef\tempurl%
\url{https://proceedings.neurips.cc/paper/2019/file/dc6a7e655d7e5840e66733e9ee67cc69-Paper.pdf}
\showURL{%
\tempurl}


\bibitem[\protect\citeauthoryear{Zeldes, Padnos, Sharir, and Peleg}{Zeldes
  et~al\mbox{.}}{2020}]%
        {auxiliary_tuning}
\bibfield{author}{\bibinfo{person}{Y. Zeldes}, \bibinfo{person}{D. Padnos},
  \bibinfo{person}{O. Sharir}, {and} \bibinfo{person}{B. Peleg}.}
  \bibinfo{year}{2020}\natexlab{}.
\newblock \showarticletitle{Technical Report: Auxiliary Tuning and its
  Application to Conditional Text Generation}.
\newblock  (\bibinfo{year}{2020}).
\newblock
\urldef\tempurl%
\url{https://arxiv.org/abs/2006.16823}
\showURL{%
\tempurl}


\bibitem[\protect\citeauthoryear{Zeng and Nie}{Zeng and Nie}{2021}]%
        {nanal2021_multi-task}
\bibfield{author}{\bibinfo{person}{Yan Zeng} {and} \bibinfo{person}{Jian{-}Yun
  Nie}.} \bibinfo{year}{2021}\natexlab{}.
\newblock \showarticletitle{A Simple and Efficient Multi-Task Learning Approach
  for Conditioned Dialogue Generation}. In
  \bibinfo{booktitle}{\emph{Proceedings of the 2021 Conference of the North
  American Chapter of the Association for Computational Linguistics: Human
  Language Technologies, {NAACL-HLT} 2021, Online, June 6-11, 2021}},
  \bibfield{editor}{\bibinfo{person}{Kristina Toutanova}, \bibinfo{person}{Anna
  Rumshisky}, \bibinfo{person}{Luke Zettlemoyer}, \bibinfo{person}{Dilek
  Hakkani{-}T{\"{u}}r}, \bibinfo{person}{Iz~Beltagy}, \bibinfo{person}{Steven
  Bethard}, \bibinfo{person}{Ryan Cotterell}, \bibinfo{person}{Tanmoy
  Chakraborty}, {and} \bibinfo{person}{Yichao Zhou}} (Eds.).
  \bibinfo{publisher}{Association for Computational Linguistics},
  \bibinfo{pages}{4927--4939}.
\newblock
\urldef\tempurl%
\url{https://doi.org/10.18653/v1/2021.naacl-main.392}
\showDOI{\tempurl}


\bibitem[\protect\citeauthoryear{Zhang and Song}{Zhang and Song}{2022}]%
        {DisCup-2022}
\bibfield{author}{\bibinfo{person}{Hanqing Zhang} {and} \bibinfo{person}{Dawei
  Song}.} \bibinfo{year}{2022}\natexlab{}.
\newblock \showarticletitle{{D}is{C}up: Discriminator Cooperative Unlikelihood
  Prompt-tuning for Controllable Text Generation}. In
  \bibinfo{booktitle}{\emph{Proceedings of the 2022 Conference on Empirical
  Methods in Natural Language Processing}}. \bibinfo{publisher}{Association for
  Computational Linguistics}, \bibinfo{address}{Abu Dhabi, United Arab
  Emirates}, \bibinfo{pages}{3392--3406}.
\newblock
\urldef\tempurl%
\url{https://aclanthology.org/2022.emnlp-main.223}
\showURL{%
\tempurl}


\bibitem[\protect\citeauthoryear{Zhang, Wang, Yin, and Huang}{Zhang
  et~al\mbox{.}}{2019b}]%
        {zhang2019emotional}
\bibfield{author}{\bibinfo{person}{Rui Zhang}, \bibinfo{person}{Zhenyu Wang},
  \bibinfo{person}{Kai Yin}, {and} \bibinfo{person}{Zhenhua Huang}.}
  \bibinfo{year}{2019}\natexlab{b}.
\newblock \showarticletitle{Emotional text generation based on cross-domain
  sentiment transfer}.
\newblock \bibinfo{journal}{\emph{IEEE Access}}  \bibinfo{volume}{7}
  (\bibinfo{year}{2019}), \bibinfo{pages}{100081--100089}.
\newblock
\urldef\tempurl%
\url{https://ieeexplore.ieee.org/document/8772090}
\showURL{%
\tempurl}


\bibitem[\protect\citeauthoryear{Zhang, Dinan, Urbanek, Szlam, Kiela, and
  Weston}{Zhang et~al\mbox{.}}{2018}]%
        {zhang2018personalizing}
\bibfield{author}{\bibinfo{person}{Saizheng Zhang}, \bibinfo{person}{Emily
  Dinan}, \bibinfo{person}{Jack Urbanek}, \bibinfo{person}{Arthur Szlam},
  \bibinfo{person}{Douwe Kiela}, {and} \bibinfo{person}{Jason Weston}.}
  \bibinfo{year}{2018}\natexlab{}.
\newblock \showarticletitle{Personalizing Dialogue Agents: {I} have a dog, do
  you have pets too?}. In \bibinfo{booktitle}{\emph{Proceedings of the 56th
  Annual Meeting of the Association for Computational Linguistics (Volume 1:
  Long Papers)}}. \bibinfo{publisher}{Association for Computational
  Linguistics}, \bibinfo{address}{Melbourne, Australia},
  \bibinfo{pages}{2204--2213}.
\newblock
\urldef\tempurl%
\url{https://doi.org/10.18653/v1/P18-1205}
\showDOI{\tempurl}


\bibitem[\protect\citeauthoryear{Zhang, Kishore, Wu, Weinberger, and
  Artzi}{Zhang et~al\mbox{.}}{2020a}]%
        {BERTScore}
\bibfield{author}{\bibinfo{person}{Tianyi Zhang}, \bibinfo{person}{Varsha
  Kishore}, \bibinfo{person}{Felix Wu}, \bibinfo{person}{Kilian~Q. Weinberger},
  {and} \bibinfo{person}{Yoav Artzi}.} \bibinfo{year}{2020}\natexlab{a}.
\newblock \showarticletitle{BERTScore: Evaluating Text Generation with {BERT}}.
  In \bibinfo{booktitle}{\emph{8th International Conference on Learning
  Representations, {ICLR} 2020, Addis Ababa, Ethiopia, April 26-30, 2020}}.
  \bibinfo{publisher}{OpenReview.net}.
\newblock
\urldef\tempurl%
\url{https://openreview.net/forum?id=SkeHuCVFDr}
\showURL{%
\tempurl}


\bibitem[\protect\citeauthoryear{Zhang, Sun, Galley, Chen, Brockett, Gao, Gao,
  Liu, and Dolan}{Zhang et~al\mbox{.}}{2020b}]%
        {zhang2019dialogpt}
\bibfield{author}{\bibinfo{person}{Yizhe Zhang}, \bibinfo{person}{Siqi Sun},
  \bibinfo{person}{Michel Galley}, \bibinfo{person}{Yen-Chun Chen},
  \bibinfo{person}{Chris Brockett}, \bibinfo{person}{Xiang Gao},
  \bibinfo{person}{Jianfeng Gao}, \bibinfo{person}{Jingjing Liu}, {and}
  \bibinfo{person}{Bill Dolan}.} \bibinfo{year}{2020}\natexlab{b}.
\newblock \showarticletitle{{DIALOGPT} : Large-Scale Generative Pre-training
  for Conversational Response Generation}. In
  \bibinfo{booktitle}{\emph{Proceedings of the 58th Annual Meeting of the
  Association for Computational Linguistics: System Demonstrations}}.
  \bibinfo{publisher}{Association for Computational Linguistics},
  \bibinfo{address}{Online}, \bibinfo{pages}{270--278}.
\newblock
\urldef\tempurl%
\url{https://doi.org/10.18653/v1/2020.acl-demos.30}
\showDOI{\tempurl}


\bibitem[\protect\citeauthoryear{Zhang, Wang, Li, Gan, Brockett, and
  Dolan}{Zhang et~al\mbox{.}}{2020c}]%
        {point_insertion}
\bibfield{author}{\bibinfo{person}{Yizhe Zhang}, \bibinfo{person}{Guoyin Wang},
  \bibinfo{person}{Chunyuan Li}, \bibinfo{person}{Zhe Gan},
  \bibinfo{person}{Chris Brockett}, {and} \bibinfo{person}{Bill Dolan}.}
  \bibinfo{year}{2020}\natexlab{c}.
\newblock \showarticletitle{{POINTER}: Constrained Progressive Text Generation
  via Insertion-based Generative Pre-training}. In
  \bibinfo{booktitle}{\emph{Proceedings of the 2020 Conference on Empirical
  Methods in Natural Language Processing (EMNLP)}}.
  \bibinfo{publisher}{Association for Computational Linguistics},
  \bibinfo{address}{Online}, \bibinfo{pages}{8649--8670}.
\newblock
\urldef\tempurl%
\url{https://doi.org/10.18653/v1/2020.emnlp-main.698}
\showDOI{\tempurl}


\bibitem[\protect\citeauthoryear{Zhang, Han, Liu, Jiang, Sun, and Liu}{Zhang
  et~al\mbox{.}}{2019a}]%
        {ERNIE}
\bibfield{author}{\bibinfo{person}{Zhengyan Zhang}, \bibinfo{person}{Xu Han},
  \bibinfo{person}{Zhiyuan Liu}, \bibinfo{person}{Xin Jiang},
  \bibinfo{person}{Maosong Sun}, {and} \bibinfo{person}{Qun Liu}.}
  \bibinfo{year}{2019}\natexlab{a}.
\newblock \showarticletitle{{ERNIE}: Enhanced Language Representation with
  Informative Entities}. In \bibinfo{booktitle}{\emph{Proceedings of the 57th
  Annual Meeting of the Association for Computational Linguistics}}.
  \bibinfo{publisher}{Association for Computational Linguistics},
  \bibinfo{address}{Florence, Italy}, \bibinfo{pages}{1441--1451}.
\newblock
\urldef\tempurl%
\url{https://doi.org/10.18653/v1/P19-1139}
\showDOI{\tempurl}


\bibitem[\protect\citeauthoryear{Zhao, Walker, and Chaturvedi}{Zhao
  et~al\mbox{.}}{2020}]%
        {acl2020_data2text}
\bibfield{author}{\bibinfo{person}{Chao Zhao}, \bibinfo{person}{Marilyn~A.
  Walker}, {and} \bibinfo{person}{Snigdha Chaturvedi}.}
  \bibinfo{year}{2020}\natexlab{}.
\newblock \showarticletitle{Bridging the Structural Gap Between Encoding and
  Decoding for Data-To-Text Generation}. In
  \bibinfo{booktitle}{\emph{Proceedings of the 58th Annual Meeting of the
  Association for Computational Linguistics, {ACL} 2020, Online, July 5-10,
  2020}}, \bibfield{editor}{\bibinfo{person}{Dan Jurafsky},
  \bibinfo{person}{Joyce Chai}, \bibinfo{person}{Natalie Schluter}, {and}
  \bibinfo{person}{Joel~R. Tetreault}} (Eds.). \bibinfo{publisher}{Association
  for Computational Linguistics}, \bibinfo{pages}{2481--2491}.
\newblock
\urldef\tempurl%
\url{https://doi.org/10.18653/v1/2020.acl-main.224}
\showDOI{\tempurl}


\bibitem[\protect\citeauthoryear{Zhao, Mathieu, and LeCun}{Zhao
  et~al\mbox{.}}{2017}]%
        {energy_based_generative_adversarial_network}
\bibfield{author}{\bibinfo{person}{Junbo~Jake Zhao},
  \bibinfo{person}{Micha{\"{e}}l Mathieu}, {and} \bibinfo{person}{Yann LeCun}.}
  \bibinfo{year}{2017}\natexlab{}.
\newblock \showarticletitle{Energy-based Generative Adversarial Networks}. In
  \bibinfo{booktitle}{\emph{5th International Conference on Learning
  Representations, {ICLR} 2017, Toulon, France, April 24-26, 2017, Conference
  Track Proceedings}}. \bibinfo{publisher}{OpenReview.net}.
\newblock
\urldef\tempurl%
\url{https://openreview.net/forum?id=ryh9pmcee}
\showURL{%
\tempurl}


\bibitem[\protect\citeauthoryear{Zheng, Zhang, Huang, and Mao}{Zheng
  et~al\mbox{.}}{2020}]%
        {aaai2020_presonalized_dialogue_generation}
\bibfield{author}{\bibinfo{person}{Yinhe Zheng}, \bibinfo{person}{Rongsheng
  Zhang}, \bibinfo{person}{Minlie Huang}, {and} \bibinfo{person}{Xiaoxi Mao}.}
  \bibinfo{year}{2020}\natexlab{}.
\newblock \showarticletitle{A Pre-Training Based Personalized Dialogue
  Generation Model with Persona-Sparse Data}. In \bibinfo{booktitle}{\emph{The
  Thirty-Fourth {AAAI} Conference on Artificial Intelligence, {AAAI} 2020, The
  Thirty-Second Innovative Applications of Artificial Intelligence Conference,
  {IAAI} 2020, The Tenth {AAAI} Symposium on Educational Advances in Artificial
  Intelligence, {EAAI} 2020, New York, NY, USA, February 7-12, 2020}}.
  \bibinfo{publisher}{{AAAI} Press}, \bibinfo{pages}{9693--9700}.
\newblock
\urldef\tempurl%
\url{https://aaai.org/ojs/index.php/AAAI/article/view/6518}
\showURL{%
\tempurl}


\bibitem[\protect\citeauthoryear{Zhong, Zhang, Wang, Liu, and Miao}{Zhong
  et~al\mbox{.}}{2020}]%
        {zhong2020towards}
\bibfield{author}{\bibinfo{person}{Peixiang Zhong}, \bibinfo{person}{Chen
  Zhang}, \bibinfo{person}{Hao Wang}, \bibinfo{person}{Yong Liu}, {and}
  \bibinfo{person}{Chunyan Miao}.} \bibinfo{year}{2020}\natexlab{}.
\newblock \showarticletitle{Towards Persona-Based Empathetic Conversational
  Models}. In \bibinfo{booktitle}{\emph{Proceedings of the 2020 Conference on
  Empirical Methods in Natural Language Processing (EMNLP)}}.
  \bibinfo{publisher}{Association for Computational Linguistics},
  \bibinfo{address}{Online}, \bibinfo{pages}{6556--6566}.
\newblock
\urldef\tempurl%
\url{https://doi.org/10.18653/v1/2020.emnlp-main.531}
\showDOI{\tempurl}


\bibitem[\protect\citeauthoryear{Zhou and Xu}{Zhou and Xu}{2020}]%
        {zhou2020learning}
\bibfield{author}{\bibinfo{person}{Wangchunshu Zhou} {and} \bibinfo{person}{Ke
  Xu}.} \bibinfo{year}{2020}\natexlab{}.
\newblock \showarticletitle{Learning to compare for better training and
  evaluation of open domain natural language generation models}. In
  \bibinfo{booktitle}{\emph{Proceedings of the AAAI Conference on Artificial
  Intelligence}}, Vol.~\bibinfo{volume}{34}. \bibinfo{pages}{9717--9724}.
\newblock
\urldef\tempurl%
\url{https://arxiv.org/abs/2002.05058}
\showURL{%
\tempurl}


\bibitem[\protect\citeauthoryear{Zhu, Lu, Zheng, Guo, Zhang, Wang, and Yu}{Zhu
  et~al\mbox{.}}{2018}]%
        {zhu2018texygen}
\bibfield{author}{\bibinfo{person}{Yaoming Zhu}, \bibinfo{person}{Sidi Lu},
  \bibinfo{person}{Lei Zheng}, \bibinfo{person}{Jiaxian Guo},
  \bibinfo{person}{Weinan Zhang}, \bibinfo{person}{Jun Wang}, {and}
  \bibinfo{person}{Yong Yu}.} \bibinfo{year}{2018}\natexlab{}.
\newblock \showarticletitle{Texygen: A Benchmarking Platform for Text
  Generation Models}. In \bibinfo{booktitle}{\emph{The 41st International ACM
  SIGIR Conference on Research \&amp; Development in Information Retrieval}}
  (Ann Arbor, MI, USA) \emph{(\bibinfo{series}{SIGIR '18})}.
  \bibinfo{publisher}{Association for Computing Machinery},
  \bibinfo{address}{New York, NY, USA}, \bibinfo{pages}{1097–1100}.
\newblock
\showISBNx{9781450356572}
\urldef\tempurl%
\url{https://doi.org/10.1145/3209978.3210080}
\showDOI{\tempurl}


\bibitem[\protect\citeauthoryear{Ziegler, Stiennon, Wu, Brown, Radford, Amodei,
  Christiano, and Irving}{Ziegler et~al\mbox{.}}{2019}]%
        {ft_human_preference}
\bibfield{author}{\bibinfo{person}{Daniel~M. Ziegler}, \bibinfo{person}{Nisan
  Stiennon}, \bibinfo{person}{Jeffrey Wu}, \bibinfo{person}{Tom~B. Brown},
  \bibinfo{person}{Alec Radford}, \bibinfo{person}{Dario Amodei},
  \bibinfo{person}{Paul~F. Christiano}, {and} \bibinfo{person}{Geoffrey
  Irving}.} \bibinfo{year}{2019}\natexlab{}.
\newblock \showarticletitle{Fine-Tuning Language Models from Human
  Preferences}.
\newblock \bibinfo{journal}{\emph{CoRR}}  \bibinfo{volume}{abs/1909.08593}
  (\bibinfo{year}{2019}).
\newblock
\showeprint[arxiv]{1909.08593}
\urldef\tempurl%
\url{http://arxiv.org/abs/1909.08593}
\showURL{%
\tempurl}


\bibitem[\protect\citeauthoryear{Zou, Yin, Zhong, Yang, Yang, and Tang}{Zou
  et~al\mbox{.}}{2021}]%
        {inverse_prompt}
\bibfield{author}{\bibinfo{person}{Xu Zou}, \bibinfo{person}{Da Yin},
  \bibinfo{person}{Qingyang Zhong}, \bibinfo{person}{Hongxia Yang},
  \bibinfo{person}{Zhilin Yang}, {and} \bibinfo{person}{Jie Tang}.}
  \bibinfo{year}{2021}\natexlab{}.
\newblock \showarticletitle{Controllable Generation from Pre-trained Language
  Models via Inverse Prompting}.
\newblock \bibinfo{journal}{\emph{CoRR}}  \bibinfo{volume}{abs/2103.10685}
  (\bibinfo{year}{2021}).
\newblock
\showeprint[arxiv]{2103.10685}
\urldef\tempurl%
\url{https://arxiv.org/abs/2103.10685}
\showURL{%
\tempurl}


\end{thebibliography}

\appendix

\end{document}